\documentclass{article}

\PassOptionsToPackage{numbers, compress}{natbib}

\usepackage[preprint]{style}

\usepackage[utf8]{inputenc}
\usepackage{fontawesome}
\usepackage[T1]{fontenc}

\usepackage{amsfonts}

\usepackage{nicefrac}

\usepackage{booktabs}

\usepackage{longtable}

\usepackage{array}

\usepackage{xltabular}

\usepackage{ragged2e}

\usepackage[table]{xcolor}

\usepackage{graphicx}

\usepackage{float}

\usepackage{pdflscape}

\usepackage{titletoc}

\usepackage[normalem]{ulem}

\usepackage{enumitem}

\usepackage{microtype}

\usepackage{url}

\usepackage{hyperref}

\title{CITYREP: A Unified Benchmark for Urban Representations Across Cities, Tasks, and Modalities}

\author{%
\begin{tabular}{c}
\textbf{Junyuan Liu}$^{1}$ \quad
\textbf{Xinglei Wang}$^{1}$ \quad
\textbf{Zichao Zeng}$^{1,2}$ \quad
\textbf{Jiazhuang Feng}$^{1}$ \\
\textbf{Quan Qin}$^{1,3}$ \quad
\textbf{Ilya Ilyankou}$^{1}$ \quad
\textbf{Guangsheng Dong}$^{1,4}$ \quad
\textbf{Tao Cheng}$^{1,\dagger}$ \\[0.6em]
{\normalfont\footnotesize $^{1}$SpaceTimeLab, University College London, UK} \\
{\normalfont\footnotesize $^{2}$3DIMPact, University College London, UK} \\
{\normalfont\footnotesize $^{3}$School of Resource and Environmental Sciences, Wuhan University, China} \\
{\normalfont\footnotesize $^{4}$State Key Laboratory of Information Engineering in Surveying, Mapping and Remote Sensing,} \\
{\normalfont\footnotesize Wuhan University, China} \\
{\normalfont\footnotesize $^{\dagger}$Corresponding author: \texttt{tao.cheng@ucl.ac.uk}}
\end{tabular}
}

\begin{document}
\maketitle

\begin{abstract}
Urban representation learning encodes complex urban environments into general-purpose embeddings for diverse downstream tasks and emerging urban foundation models. However, current evaluations are limited, typically focusing on one or two cities and tasks and relying on random splits that introduce spatial leakage, leading to inflated performance and weak support for cross-location generalization and fair comparison.
To address this, we propose \textbf{CityRep}, a unified benchmark that evaluates urban representations across data modalities, cities, and tasks using spatially structured splits.
CityRep consists of three key components: (1) a spatial unit-agnostic evaluation framework that supports heterogeneous urban representations through a standardized alignment module; (2) a unified evaluation protocol using block-based spatial splits to mitigate spatial leakage and enable rigorous model comparison; and (3) an extensible multi-city, multi-task benchmark suite spanning 8 cities and 8 tasks across regression, classification, and distribution prediction.
We evaluate 11 representative urban representation models. Results show that performance is highly sensitive to the split protocol, with random splits inflating scores and altering model rankings. We also observe substantial variability across cities and tasks, underscoring the need for generalization-aware evaluation.
CityRep is released as a reproducible benchmark with datasets, evaluation pipelines, and diagnostic tools to facilitate fair comparison and support future research in urban representation learning towards urban foundation models.\\
\faCubes~Code:~\url{https://github.com/inwind0212/CityRep}.

\end{abstract}

\section{Introduction}
\label{sec:introduction}

\begin{figure}
    \centering
    \includegraphics[width=1\linewidth]{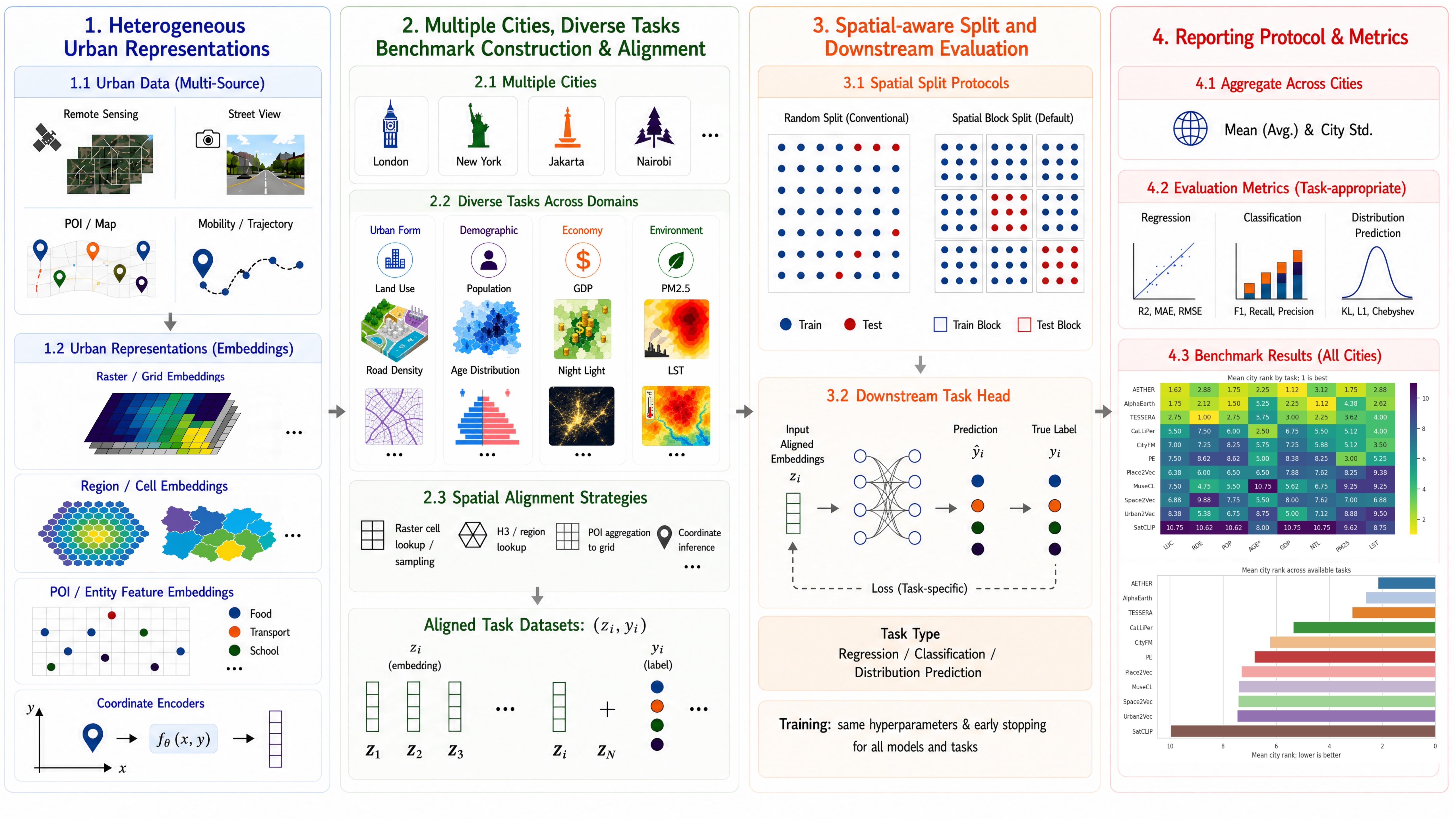}
    \vspace{-1em}
    \caption{\textbf{Framework of CityRep Benchmark.} CityRep standardizes the evaluation of heterogeneous urban representations by aligning different spatial supports to common downstream task units, evaluating them across eight cities and eight tasks, and using spatial block splits to mitigate leakage.}
    \label{fig:framework}
\end{figure}

Urban representation learning seeks to turn heterogeneous observations of cities into reusable spatial embeddings. Recent models draw on remote sensing~\cite{klemmer2025satclip, feng2025tessera, brown2025alphaearth}, street-view imagery~\cite{wang2020urban2vec, yong2024musecl}, and points of interest~\cite{yan2017itdl, huang2022sppe, wang2025multi, balsebre2024city, liu2025enriching} to encode geographic entities, regions, or locations. This motivation parallels the broader shift toward foundation models: representations learned from broad urban data are expected to transfer across tasks, locations, and domains. Yet the evaluation of these representations remains much less unified. Reported results are often tied to a particular model interface or a particular format of downstream task, making it difficult to assess whether an embedding is broadly useful or only effective under a narrow evaluation setup.

Existing evaluations are also too narrow to support claims about general-purpose urban representations. 
Many studies evaluate on one or two cities, a small number of tasks, or a single label type. Such experiments are valuable for demonstrating a specific application, but they do not reveal whether a representation transfers across urban contexts or across qualitatively different prediction problems.
This limitation is especially important for the emerging urban foundation models, whose value depends on broad reuse.
A benchmark should therefore cover multiple cities, multiple urban domains, and multiple task types, while remaining extensible to incorporate new cities, tasks, and models.

Finally, evaluation must explicitly account for spatial dependence. In urban representation learning, we expect models to leverage information from observed regions to make predictions in unseen areas, rather than merely interpolate among nearby samples. While this challenge has been well recognized in spatial validation studies~\cite{roberts2017crossvalidation,valavi2019blockcv,meyer2019spatial}, most existing urban representation models are still evaluated using random splits that ignore spatial structure. 
We address this gap by establishing a unified spatially structured evaluation protocol. 
Our benchmark provides empirical evidence that random splits can substantially inflate performance and lead to over-optimistic conclusions about model generalization.

We introduce \textbf{CityRep}, a unified benchmark for urban representations across modalities, tasks, and cities, as shown in Figure~\ref{fig:framework}.
The central goal of CityRep is to move urban representation evaluation beyond narrow, single-setting comparisons.
Rather than assessing an embedding on one city, one task, or one type of urban label, CityRep asks whether a representation remains useful across a sufficiently broad set of urban phenomena and geographic contexts.
To operationalize this goal, we construct downstream evaluation data from four key dimensions of urban systems: morphology, demographics, economy, and environment.
These dimensions are instantiated as eight tasks across eight cities, covering classification, regression, and distribution prediction.
CityRep pairs this broad task suite with a common evaluation protocol for heterogeneous representations.
Models based on rasters, regions, entities, or coordinates can be evaluated under the same downstream interface, while spatially structured splits are used to reduce leakage between nearby training and test samples.
CityRep therefore makes it possible to examine not only average performance, but also how representation quality changes across urban contexts, task domains, label types, and split protocols.
This provides a basis for assessing whether urban representations are genuinely general-purpose, rather than effective only under narrow evaluation settings. In summary, our contributions are: 
\begin{itemize}

    \item We introduce \textbf{CityRep}, a unified and extensible benchmark for urban representation learning that supports heterogeneous representation types across data modalities, cities, and downstream urban tasks.

    \item We design a spatially structured evaluation methodology, including spatial-unit alignment and block-based spatial splits, to enable fair comparison across heterogeneous urban representations while mitigating spatial leakage.

    \item We conduct a large-scale empirical study of eleven representative urban and geospatial representation models across eight cities and eight tasks, showing that benchmark conclusions are highly sensitive to the evaluation protocol, task domain, and urban context.

    \item We publicly release datasets, evaluation pipelines, processed benchmarks, model manifests, and diagnostic tools to support reproducible research on urban representation learning and urban foundation models.

\end{itemize}

\section{Related Work}
\label{sec:related_work}

\paragraph{Urban Representation Learning}
\label{sec:related_urban_representation}
Existing urban representation learning methods are highly heterogeneous, differing in the data they utilise, the spatial units they operate upon, and the urban signals they encode. Following geographic information systems (GIS) taxonomy, this heterogeneity is largely shaped by whether models ingest vector data, such as points, polylines, and polygons, or raster data, such as satellite and street-view imagery. Vector-based methods often rely on POI data but produce different outputs: category embedding methods, including Place2Vec~\citep{yan2017itdl}, POI2Vec~\citep{feng2017poi2vec}, and SPPE~\citep{huang2022sppe}, capture spatial co-occurrence patterns of POI types and require aggregation to represent urban spaces, whereas entity embedding models such as Urban2Vec~\citep{wang2020urban2vec}, HGI~\citep{huang2023hgi}, and CityFM~\citep{balsebre2024city} directly encode regions, buildings, or roads. Raster-based methods, by contrast, naturally produce grid-cell embeddings, with earth-observation foundation models such as AlphaEarth Foundation~\citep{brown2025alphaearth} and TESSERA~\citep{feng2025tessera} enabling dense large-scale representations, and AETHER further incorporating POI semantics into raster foundations~\citep{liu2025beyond}. A related line of coordinate-based encoders, including Space2Vec~\citep{mai2020iclr}, SatCLIP~\citep{klemmer2025satclip}, and CaLLiPer~\citep{wang2025multi}, learns representations for continuous locations from POIs, imagery, or language supervision. Consequently, the resulting embeddings operate over disparate spatial supports (e.g., regions, H3 cells~\citep{uberh3}, raster grids, and coordinates) and capture varying urban information. This creates a central evaluation challenge: heterogeneous methods cannot be fairly compared without spatial alignment to standard task units, and evaluation on only a few downstream tasks is insufficient for representations capturing diverse urban signals.

\paragraph{Geospatial Benchmarks and Spatial Evaluation}
\label{sec:related_geospatial_benchmarks}

Urban representation learning can be viewed as a fine-grained, city-focused branch of geospatial representation learning.
Existing geospatial benchmarks have largely started from image- or raster-centered settings.
TorchGeo provides reusable infrastructure for geospatial data loading, sampling, and model development~\cite{stewart2025torchgeo}, while GEO-Bench, SatlasPretrain, and PANGAEA standardize Earth-observation evaluation and pretraining across tasks, sensors, resolutions, regions, and temporal settings~\cite{lacoste2023geobench,bastani2023satlaspretrain,marsocci2026pangaea}.
Recent benchmarks move closer to spatial representation learning.
TorchSpatial evaluates general-purpose location encoders~\cite{wu2024torchspatial}, OBSR evaluates geospatial embedders on regional and trajectory tasks~\cite{moska2025obsr}, and MoRA introduces human-centric social and economic prediction tasks based on mobility-centered representations~\cite{wen2026mora}.
However, they still provide limited evidence on whether representations capture fine-grained intra-urban structure and functions.
Spatial evaluation is also critical. Prior work shows that random splits can overestimate performance under spatial dependence and recommends spatially structured validation for assessing transfer to unseen areas~\cite{roberts2017crossvalidation,valavi2019blockcv,meyer2019spatial,ploton2020spatial}.
These gaps motivate a unified multi-city benchmark for heterogeneous urban representations under spatially robust evaluation protocols.

\section{CityRep Framework and Benchmark}
\label{sec:cityrep}
CityRep aims to make urban representations comparable across spatial units, tasks, and cities.
Figure~\ref{fig:framework} provides an overview of the CityRep benchmark. 
Urban representations arise in diverse forms, including raster-based embeddings, region-level features, POI or entity representations, and coordinate-based encoders, while downstream labels are defined over heterogeneous spatial units. 
To enable comparison across such settings, CityRep first aligns each representation to common task units across multiple cities and domains through standardized spatial alignment strategies, constructing unified task datasets. 
These aligned features are then evaluated under spatially structured split protocols, ensuring that performance reflects generalization to unseen areas rather than interpolation among nearby samples. 
Finally, results are aggregated across cities using task-appropriate metrics, enabling consistent and robust comparison of different urban representation models.

\subsection{Problem Definition}
\label{sec:problem_definition}

Urban representation learning aims to compress heterogeneous observations of a city into reusable spatial embeddings. Let $\mathcal{D}_{m,c}$ denote the input data used by representation model $m$ in city $c$, such as satellite imagery, street-view imagery, points of interest, road networks, or geographic coordinates. The model transforms these observations into a city representation
\begin{equation}
    E_{m,c}=f_m(\mathcal{D}_{m,c}),
\end{equation}
where $E_{m,c}$ may be defined on a raster grid, a set of regions or cells, a collection of POIs or spatial entities, or a continuous coordinate domain. CityRep does not prescribe how $f_m$ is trained. Instead, it evaluates whether the resulting representation captures transferable urban information that is useful for downstream prediction across multiple task domains.

For each city $c$ and downstream task $t$, CityRep defines a set of task units $\mathcal{U}_{c,t}=\{u_i\}_{i=1}^{n_{c,t}}$ and labels $\mathbf{y}_{c,t}$. Because the native unit of $E_{m,c}$ generally differs from the unit of $\mathcal{U}_{c,t}$, the central benchmark operation is spatial alignment:
\begin{equation}
    \mathbf{X}_{m,c,t}=A(E_{m,c},\mathcal{U}_{c,t}), \qquad \mathbf{X}_{m,c,t}\in\mathbb{R}^{n_{c,t}\times d_m},
    \label{eq:alignment}
\end{equation}
where $A(\cdot)$ maps the native representation to the task units and $d_m$ is the embedding dimension. Each row of $\mathbf{X}_{m,c,t}$ is the feature vector assigned to one task unit and is paired with the corresponding label in $\mathbf{y}_{c,t}$.

Given the aligned features, CityRep evaluates each representation with a fixed downstream predictor
\begin{equation}
    \hat{\mathbf{y}}_{c,t}=g_{\theta,t}(\mathbf{X}_{m,c,t}),
\end{equation}
where the prediction head is chosen according to the task type: regression, classification, or distribution prediction. This formulation separates representation learning from benchmark evaluation. Models may differ in input data modality, pretraining objective, and native spatial support, but they are compared by the same question: after alignment to the downstream task units, how much task-relevant urban information does the representation provide?

\subsection{Spatial Alignment}
\label{sec:alignment}

Spatial alignment is the mechanism that makes heterogeneous urban representations comparable.
Representation models and downstream tasks are often defined on different spatial supports, such as raster cells, regions, entities, or coordinates.
CityRep therefore treats alignment as a spatial matching problem: for each downstream task unit $u_i$, the benchmark assigns a representation vector that corresponds to the same location or spatial area.
The goal is not to force all models and tasks onto a single universal grid, but to preserve each task's native evaluation unit while mapping every representation to that unit in a consistent way.

CityRep implements this principle according to the spatial relationship between the representation unit and the downstream task unit. For raster or region-level embeddings, if the representation units are finer than the task unit, CityRep aggregates the embeddings within the task unit.
If the representation unit is coarser than the task unit, all task units covered by the same representation unit share its embedding.
When the spatial supports are directly compatible, alignment reduces to raster sampling, cell lookup, or spatial join. For entity-level embeddings, such as POI or map-entity representations, CityRep first aggregates entities to an intermediate support, such as H3 cells, and then applies the same region-matching rules.
This avoids directly aggregating sparse and unevenly distributed entities to every downstream task unit, which can otherwise produce many missing features and degrade downstream evaluation; an ablation supporting this H3-first design is provided in Appendix~\ref{app:h3-first-aggregation}.
For coordinate encoders, no stored spatial support is required: CityRep queries the encoder at a representative coordinate of each task unit, such as a point location, raster-cell center, or polygon representative point, and creates an embedding from these sample points.

\subsection{Spatial Split}
\label{sec:spatial_splits}

CityRep uses spatial splitting to define the generalization target of the benchmark. Let $\mathcal{U}_{c,t}$ be the task units for city $c$ and task $t$. Instead of drawing train and test samples independently from $\mathcal{U}_{c,t}$, we first partition the spatial extent of the task into a set of non-overlapping blocks $\mathcal{B}_{c,t}=\{B_j\}_{j=1}^{J_{c,t}}$. Each task unit is assigned to one block by spatial containment or by the location of its representative point:
\begin{equation}
    b(u_i) \in \mathcal{B}_{c,t}.
\end{equation}
The train, validation, and test sets are then formed by assigning blocks, not individual task units, to disjoint subsets:
\begin{equation}
    \mathcal{B}_{c,t}
    =
    \mathcal{B}^{\mathrm{train}}_{c,t,k}
    \cup
    \mathcal{B}^{\mathrm{val}}_{c,t,k}
    \cup
    \mathcal{B}^{\mathrm{test}}_{c,t,k},
\end{equation}
where $k$ indexes the random seed and the three block sets are mutually disjoint. The corresponding task-unit split is induced by block membership:
\begin{equation}
    \mathcal{U}^{s}_{c,t,k}
    =
    \{u_i \in \mathcal{U}_{c,t}: b(u_i)\in \mathcal{B}^{s}_{c,t,k}\},
    \qquad
    s\in\{\mathrm{train},\mathrm{val},\mathrm{test}\}.
\end{equation}

This formulation makes spatial separation part of the evaluation protocol. Test samples are held out together with spatially proximate samples within the same block, reducing the chance that performance is driven primarily by local interpolation from adjacent training points. The split is task-specific because different tasks may have different spatial extents, valid masks, and label supports, but it is model-invariant: all models evaluated on the same city--task pair use the same block partition and the same seed-specific block assignment. In the current benchmark instantiation, we use a $10 \times 10$ spatial block partition for the main results and report a block-granularity sensitivity analysis in Appendix~\ref{app:split20_results}. Detailed split configurations, visualization examples, and cross-seed test-block statistics are provided in Appendix~\ref{app:spatial-split}.

\subsection{Tasks and Dataset}
\label{sec:tasks_datasets}
CityRep is built around eight downstream tasks that reflect different dimensions of urban systems. The goal is to test whether an urban representation captures information that transfers beyond a single visual pattern or geographic prior. The tasks span regression, classification, and distribution prediction.

\paragraph{Downstream Tasks.}
CityRep includes eight downstream tasks organized into four urban domains: Morphology ($\spadesuit$), Demographics ($\heartsuit$), Economy ($\diamondsuit$), and Environment ($\clubsuit$). Details of the downstream task datasets and raw data sources are provided in Appendix~\ref{app:raw-data}.
\begin{itemize}[leftmargin=*, label=-, nosep]
    \item \textbf{Land-use classification (LUC)$^\spadesuit$.}
    This task uses city-specific zoning or land-use datasets from official or public planning sources~\cite{nyc_dcp_pluto_25v4,ura_master_plan_2019_landuse_2025,nsw_dphi_epi_land_zoning,city_of_cape_town_zoning_2019,dki_jakarta_pergub_31_2022_rdtr,mcgm_mumbai_elu_2012,worldbank_nairobi_landuse_2010}. It evaluates whether representations capture semantic urban functions such as residential, commercial, industrial, transportation, green space, institutional, utilities, water bodies, and mixed-use areas.

    \item \textbf{Road-density regression (RDE)$^\spadesuit$.}
    This task uses OpenStreetMap road-network data~\cite{osm_road_network_extract_cityrep_2026}. It evaluates whether representations capture physical street structure and connectivity.

    \item \textbf{Population regression (POP)$^\heartsuit$.}
    This task uses WorldPop gridded population datasets~\cite{worldpop_population_counts_r2024b}. It evaluates whether representations capture spatial variation in population intensity.

    \item \textbf{Age-distribution prediction (AGE)$^\heartsuit$.}
    This task uses WorldPop age--sex datasets~\cite{worldpop_agesex_r2024b}. It evaluates whether representations capture demographic composition across age groups.

    \item \textbf{Gross Domestic Product regression (GDP)$^\diamondsuit$.}
    This task uses gridded GDP datasets from Kummu et al.~\citep{kummu_gridded_gdp_2025}. It evaluates whether representations capture spatial variation in economic output.

    \item \textbf{Nighttime lights regression (NTL)$^\diamondsuit$.}
    This task uses the VIIRS Nighttime Lights Annual V2.2 product~\citep{eog_viirs_vnl_v22,elvidge2021viirs_vnl}. It evaluates whether representations capture spatial patterns of human activity, electrification, commercial intensity, and infrastructure use visible through nighttime illumination.

    \item \textbf{PM$_{2.5}$ regression$^\clubsuit$.}
    This task uses SEDAC/CIESIN annual PM$_{2.5}$ concentration datasets~\cite{van_donkelaar_pm25_v5gl04_2024}. It evaluates whether representations capture fine particulate pollution exposure.

    \item \textbf{Land-surface-temperature regression (LST)$^\clubsuit$.}
    This task uses MODIS/Terra MOD11A2 daytime land-surface-temperature datasets~\cite{wan_mod11a2_v061_2021}. It evaluates whether representations capture surface thermal conditions related to land cover, density, vegetation, and impervious surface.
\end{itemize}

\paragraph{Cities and extensibility.}
Most sources used in CityRep are global or near-global, including WorldPop, gridded GDP, nighttime lights, PM$_{2.5}$, MODIS LST, and OpenStreetMap. As a result, adding a new city mainly requires defining the boundary, extracting the same source layers, and running the standard task construction and alignment pipeline. We instantiate the benchmark on London, New York, Singapore, Sydney, Mumbai, Nairobi, Jakarta, and Cape Town, covering developed and developing urban contexts across Europe, North America, Asia, Africa, and Australia. To further demonstrate extensibility, Appendix~\ref{app:extended-26-cities} extends the evaluation of global remote-sensing representations to 26 cities using the same benchmark pipeline.

Table~\ref{tab:city_task_counts} reports the number of downstream task units for each city and task. The counts vary because cities differ in spatial extent, valid masks, source resolution, and task support. Dense raster-derived tasks such as population, road density, and land-surface temperature contain many units, while coarser grids such as GDP, NTL, and PM$_{2.5}$ contain fewer.

\begin{table}[!tb]
\centering
\caption{Number of prediction units for each downstream task across eight cities.}
\label{tab:city_task_counts}
\scriptsize
\resizebox{\textwidth}{!}{
\begin{tabular}{lrrrrrrrr}
\toprule
Task & London & New York & Singapore & Sydney & Mumbai & Nairobi & Jakarta & Cape Town \\
\midrule
\rowcolor{gray!15}\multicolumn{9}{l}{$\spadesuit$~\textit{Morphology}} \\
Land use & 100,000 & 100,000 & 100,000 & 100,000 & 100,000 & 100,000 & 100,000 & 100,000 \\
Road density & 297,314 & 187,028 & 83,393 & 612,027 & 58,778 & 81,389 & 75,724 & 343,593 \\
\rowcolor{gray!15}\multicolumn{9}{l}{$\heartsuit$~\textit{Demographics}} \\
Population & 266,183 & 114,276 & 64,172 & 416,254 & 43,522 & 61,677 & 74,885 & 167,771 \\
Age distribution & 234,428 & 87,736 & 32,990 & 265,815 & 43,455 & 55,521 & 74,096 & 136,093 \\
\rowcolor{gray!15}\multicolumn{9}{l}{$\diamondsuit$~\textit{Economy}} \\
Gross Domestic Product & 2,977 & 1,812 & 806 & 6,121 & 575 & 817 & 756 & 3,420 \\
Nighttime lights & 2,977 & 1,867 & 839 & 6,119 & 583 & 818 & 756 & 3,434 \\
\rowcolor{gray!15}\multicolumn{9}{l}{$\clubsuit$~\textit{Environment}} \\
PM$_{2.5}$ & 2,072 & 1,038 & 551 & 4,231 & 407 & 569 & 527 & 2,380 \\
Land surface temperature & 295,364 & 112,433 & 70,009 & 592,033 & 55,394 & 81,389 & 75,117 & 339,759 \\
\bottomrule
\end{tabular}
}
\end{table}
\subsection{Selected Baselines} 
\label{sec:baselines}

We evaluate eleven representative urban and geospatial representation models. The reproduction pipeline uses Foursquare POIs~\cite{foursquare_os_places_2026}, 
Mapillary street-view imagery~\cite{mapillary_open_streetlevel_imagery_2026}, 
OpenStreetMap entities~\cite{osm_road_network_extract_cityrep_2026}, 
and public remote-sensing embedding products~\cite{feng2025tessera,brown2025alphaearth} 
as the main raw data sources.

\begin{itemize}[leftmargin=*, label=-, nosep]
    \item \textbf{PE}~\cite{mai2022review} Position encoding (PE) functions encode multi-scale location signals, serving as simple urban representations based entirely on spatial information. We select SphereC~\cite{mai2023sphere2vec} as a representative example to illustrate the performance of these methods.
    \item \textbf{Place2Vec}~\cite{yan2017itdl} learns place representations from POI context. We reproduce it with Foursquare POIs and aggregate the learned embeddings to H3 cells.

    \item \textbf{Space2Vec}~\cite{mai2020iclr} represents locations through a coordinate encoder. We train it with POI category supervision and query the encoder directly at raster-cell centers or land-use point coordinates.

    \item \textbf{CaLLiPer}~\cite{wang2025multi} is a coordinate-based urban representation pretrained via language supervision from POI textual descriptions. Its embeddings are exported through the same interface as Space2Vec.

    \item \textbf{CityFM}~\cite{balsebre2024city} learns urban representations from map entity information. We reproduce it using OpenStreetMap entities and export embeddings to H3 cells.

    \item \textbf{Urban2Vec}~\cite{wang2020urban2vec} combines street-view and POIs for learning region representations. We construct its inputs from Mapillary imagery and Foursquare POIs, and root the embeddings in H3 cells.

    \item \textbf{MuseCL}~\cite{yong2024musecl} is a multimodal urban representation model. Since consistent mobility data are unavailable across all benchmark cities, we implement a CityRep-compatible variant that replaces the mobility branch with Foursquare POI semantics while retaining the street-view, remote-sensing, and semantic fusion components.

    \item \textbf{SatCLIP}~\cite{klemmer2025satclip} is a pretrained geographic coordinate encoder. We use the model checkpoint and query it directly at downstream task locations. Although not intended for city-level representation learning, it is included for being a representative imagery-based coordinate embedding method.

    \item \textbf{TESSERA}~\cite{feng2025tessera} provides pretrained remote-sensing embedding rasters. We use the released embeddings as fixed raster representations and align them to task units by raster sampling.

    \item \textbf{AlphaEarth}~\cite{brown2025alphaearth} is a large-scale pretrained geospatial embedding product. We crop or sample its released raster embeddings for each city and task.

    \item \textbf{AETHER}~\cite{liu2025beyond} is a POI-guided alignment framework for pretrained imagery embeddings. We reproduce it by aligning AlphaEarth embedding inputs with Foursquare POI semantics.
\end{itemize}

\subsection{Evaluation}
\label{sec:evaluation}

CityRep evaluates each representation after spatial alignment to the downstream task units.
For a model $m$, city $c$, and task $t$, the aligned feature matrix $\mathbf{X}_{m,c,t}$ is paired with the task labels $\mathbf{y}_{c,t}$ and used to train a lightweight task head.
To make model comparison depend primarily on the representation rather than on downstream model engineering, CityRep uses the same predictor family and training protocol for all representation models.
After alignment, each representation is evaluated by a task-specific prediction head:
\begin{equation}
    \hat{\mathbf{y}}_{c,t}=g_{\theta,t}(\mathbf{X}_{m,c,t}),
\end{equation}
where the output layer and loss are selected according to the task type.

\paragraph{Evaluation metrics.}
CityRep uses nine task-appropriate metrics across the three prediction types.
For regression tasks, including road density, population, GDP, NTL, PM$_{2.5}$, and land surface temperature, we report $R^2$, mean absolute error (MAE), and root mean squared error (RMSE), with $R^2$ used as the primary metric.
For land-use classification, we report macro F1, macro recall, and macro precision, with macro F1 used as the primary metric because it gives equal weight to each class under imbalanced labels.
For age-distribution prediction, we report KL divergence, Chebyshev distance, and L1 distance between the predicted and target distributions, with KL divergence used as the primary metric.
Higher values are better for $R^2$, F1, recall, and precision, while lower values are better for MAE, RMSE, KL divergence, Chebyshev distance, and L1 distance.
Formal metric definitions are provided in Appendix~\ref{app:evaluation-metrics}.

\paragraph{Training and aggregation.}
All downstream predictors use the same MLP architecture and training protocol across models and tasks.
For each model--task--city setting, we run five spatial split seeds, $\{42,24,7,0,100\}$, and average the primary test metric over seeds to obtain a city-level score.
Table~\ref{tab:main_results_avg_cstd} reports, for each model and task, the mean of city-level scores across cities as \textit{Avg.}, together with the cross-city standard deviation as \textit{C Std.}
Since different tasks use different primary metrics and metric scales, CityRep reports raw task metrics in the main result columns and uses \textit{Mean City Rank} as a rank-based diagnostic summary for comparing models across tasks and cities.
Lower values indicate better overall rank.
Additional details on training, aggregation, and rank computation are provided in Appendix~\ref{app:training-protocol}.

\section{Experiments}
\label{sec:experiments}
We evaluate eleven representation models on eight tasks across eight cities. The experiments examine three questions: which representations are the best under a unified protocol, whether performance is stable across cities and tasks, and how much random splits overestimate generalization compared with spatial splits.

\subsection{Main Performance}
\label{sec:main_performance}

Table~\ref{tab:main_results_avg_cstd} shows that large-scale pretrained geospatial representations achieve the strongest overall transfer. AETHER, AlphaEarth, and TESSERA obtain the best mean city ranks, and they dominate most tasks. This suggests that broad spatial coverage and large-scale pretraining are highly valuable when a representation is expected to support heterogeneous urban prediction tasks. We further report linear-probe results in Appendix~\ref{app:linear_probe}, which show broadly comparable model rankings under a lower-capacity downstream evaluator.

However, the ranking is not determined by pretraining scale alone. Several specialized or simple representations remain competitive on specific tasks. CityFM achieves strong performance on LST, indicating that map entities can provide useful signals for built intensity and thermal conditions. CaLLiPer is also competitive on AGE. Even the PE baseline performs strongly on the spatially smoother environmental task PM$_{2.5}$, suggesting that location-only signals can be informative when the target exhibits broad spatial gradients. These results show that smaller, simpler, or more targeted representations can still be effective when their encoded signals align well with the downstream phenomenon.

Another observation is that input modality alone is insufficient to explain performance across tasks. More modalities do not necessarily translate into stronger performance. At the same time, models with similar modalities exhibit distinct task-specific strengths. Among raster or raster-enhanced embeddings, AlphaEarth performs best on LUC, POP, NTL, and LST, TESSERA leads RDE, and AETHER is strongest on GDP, PM$_{2.5}$, and AGE. POI- and entity-based models also differ substantially, with CaLLiPer outperforming Place2Vec and Space2Vec overall and CityFM remaining competitive on LST. These results indicate that benchmark performance is shaped not only by input modality, but also by model architecture, pretraining objective, spatial support, and alignment strategy.

{\renewcommand{\arraystretch}{1.22}
\begin{table*}[!tb]
\centering
\caption{
\textbf{Main benchmark results on CityRep.} For each task, \textit{Avg.} reports the mean primary metric across 8 cities and 5 random seeds under the spatial block split. \textit{C Std.} reports the cross-city standard deviation of city-level performance. Type indicates the main data sources used by each representation: \textit{L} = location, \textit{P} = POI, \textit{R} = remote sensing, \textit{S} = street-view imagery, and \textit{O} = other urban/map data. \textit{Mean City Rank} is computed from the unrounded city-level rankings across all tasks and cities (lower is better). Best, second-best, and third-best results in the Avg. columns are in \textbf{bold}, \underline{underlined}, and \uwave{wavy-underlined}, respectively. *For AGE, lower KL divergence indicates better performance, and results are reported only for the four cities with the most reliable age--sex source coverage.
}
\label{tab:main_results_avg_cstd}
\vspace{0.4em}
\scriptsize
\setlength{\tabcolsep}{1.8pt}
\resizebox{\textwidth}{!}{
\begin{tabular}{ll*{17}{c}}
\toprule
& & \multicolumn{4}{c}{$\spadesuit$~Morphology}
& \multicolumn{4}{c}{$\heartsuit$~Demographics}
& \multicolumn{4}{c}{$\diamondsuit$~Economy}
& \multicolumn{4}{c}{$\clubsuit$~Environment}
& \multicolumn{1}{c}{Overall} \\
\cmidrule(lr){3-6} \cmidrule(lr){7-10} \cmidrule(lr){11-14} \cmidrule(lr){15-18} \cmidrule(lr){19-19}
Model & Type
& \multicolumn{2}{c}{LUC}
& \multicolumn{2}{c}{RDE}
& \multicolumn{2}{c}{POP}
& \multicolumn{2}{c}{AGE$^{*}$}
& \multicolumn{2}{c}{GDP}
& \multicolumn{2}{c}{NTL}
& \multicolumn{2}{c}{PM$_{2.5}$}
& \multicolumn{2}{c}{LST}
& Rank \\
\cmidrule(lr){3-4} \cmidrule(lr){5-6}
\cmidrule(lr){7-8} \cmidrule(lr){9-10}
\cmidrule(lr){11-12} \cmidrule(lr){13-14}
\cmidrule(lr){15-16} \cmidrule(lr){17-18}
\cmidrule(lr){19-19}
&
& \shortstack{Avg.\\(F1 $\uparrow$)} & \shortstack{\textit{C}\\\textit{Std.}}
& \shortstack{Avg.\\($R^2 \uparrow$)} & \shortstack{\textit{C}\\\textit{Std.}}
& \shortstack{Avg.\\($R^2 \uparrow$)} & \shortstack{\textit{C}\\\textit{Std.}}
& \shortstack{Avg.\\(KL $\downarrow$)} & \shortstack{\textit{C}\\\textit{Std.}}
& \shortstack{Avg.\\($R^2 \uparrow$)} & \shortstack{\textit{C}\\\textit{Std.}}
& \shortstack{Avg.\\($R^2 \uparrow$)} & \shortstack{\textit{C}\\\textit{Std.}}
& \shortstack{Avg.\\($R^2 \uparrow$)} & \shortstack{\textit{C}\\\textit{Std.}}
& \shortstack{Avg.\\($R^2 \uparrow$)} & \shortstack{\textit{C}\\\textit{Std.}}
& \shortstack{Mean\\City Rank} \\
\midrule
PE & \textit{L}
& 0.149 & \textit{0.035} & 0.144 & \textit{0.080} & 0.176 & \textit{0.088} & 0.040 & \textit{0.022} & 0.132 & \textit{0.120} & 0.237 & \textit{0.269} & \underline{0.492} & \textit{0.281} & 0.378 & \textit{0.198} & 6.828 \\
Place2Vec & \textit{P}
& 0.168 & \textit{0.025} & 0.217 & \textit{0.093} & 0.288 & \textit{0.081} & 0.043 & \textit{0.022} & 0.191 & \textit{0.134} & 0.303 & \textit{0.159} & 0.063 & \textit{0.090} & 0.138 & \textit{0.097} & 7.313 \\
Space2Vec & \textit{LP}
& 0.166 & \textit{0.046} & 0.118 & \textit{0.060} & 0.218 & \textit{0.116} & 0.041 & \textit{0.022} & 0.205 & \textit{0.155} & 0.305 & \textit{0.145} & 0.236 & \textit{0.222} & 0.320 & \textit{0.186} & 7.438 \\
CaLLiPer & \textit{LP}
& 0.189 & \textit{0.059} & 0.161 & \textit{0.061} & 0.298 & \textit{0.131} & \underline{0.038} & \textit{0.022} & 0.263 & \textit{0.142} & 0.364 & \textit{0.172} & 0.328 & \textit{0.224} & 0.440 & \textit{0.178} & 5.359 \\
CityFM & \textit{PO}
& 0.165 & \textit{0.043} & 0.187 & \textit{0.083} & 0.228 & \textit{0.080} & 0.041 & \textit{0.024} & 0.199 & \textit{0.083} & 0.369 & \textit{0.206} & 0.348 & \textit{0.254} & \uwave{0.496} & \textit{0.166} & 6.250 \\
Urban2Vec & \textit{PRS}
& 0.143 & \textit{0.016} & 0.227 & \textit{0.100} & 0.278 & \textit{0.062} & 0.044 & \textit{0.022} & 0.316 & \textit{0.132} & 0.319 & \textit{0.163} & 0.068 & \textit{0.105} & 0.137 & \textit{0.082} & 7.469 \\
MuseCL & \textit{PRS}
& 0.153 & \textit{0.017} & 0.244 & \textit{0.099} & 0.306 & \textit{0.078} & 0.046 & \textit{0.022} & 0.338 & \textit{0.165} & 0.336 & \textit{0.153} & 0.039 & \textit{0.096} & 0.159 & \textit{0.117} & 7.422 \\
SatCLIP & \textit{LR}
& 0.107 & \textit{0.031} & 0.072 & \textit{0.051} & 0.067 & \textit{0.077} & 0.043 & \textit{0.022} & -0.003 & \textit{0.084} & 0.076 & \textit{0.204} & 0.032 & \textit{0.121} & 0.184 & \textit{0.086} & 9.984 \\
TESSERA & \textit{R}
& \uwave{0.322} & \textit{0.065} & \textbf{0.631} & \textit{0.085} & \uwave{0.675} & \textit{0.110} & 0.040 & \textit{0.019} & \uwave{0.581} & \textit{0.187} & \underline{0.595} & \textit{0.166} & \uwave{0.396} & \textit{0.158} & 0.465 & \textit{0.158} & \uwave{3.141} \\
AlphaEarth & \textit{R}
& \textbf{0.346} & \textit{0.064} & \underline{0.601} & \textit{0.088} & \textbf{0.695} & \textit{0.118} & \uwave{0.039} & \textit{0.017} & \underline{0.612} & \textit{0.158} & \textbf{0.650} & \textit{0.132} & 0.390 & \textit{0.179} & \textbf{0.509} & \textit{0.143} & \underline{2.625} \\
AETHER & \textit{PR}
& \underline{0.343} & \textit{0.070} & \uwave{0.566} & \textit{0.095} & \underline{0.694} & \textit{0.101} & \textbf{0.037} & \textit{0.019} & \textbf{0.750} & \textit{0.125} & \uwave{0.537} & \textit{0.120} & \textbf{0.568} & \textit{0.154} & \underline{0.500} & \textit{0.157} & \textbf{2.172} \\

\midrule
Mean over models & --
& 0.205 & \textit{0.043}
& 0.288 & \textit{0.081}
& 0.357 & \textit{0.095}
& 0.041 & \textit{0.021}
& 0.326 & \textit{0.135}
& 0.372 & \textit{0.172}
& 0.269 & \textit{0.171}
& 0.339 & \textit{0.143}
& -- \\
\bottomrule
\end{tabular}
}
\end{table*}
}

\subsection{Performance across Cities and Tasks}

The final row of Table~\ref{tab:main_results_avg_cstd} provides a complementary task-level view by averaging each column over models. It reveals that different urban phenomena are captured by current representations to different degrees. The cross-city standard deviations also differ substantially across tasks. NTL, PM$_{2.5}$, GDP, and LST exhibit larger average C-Std. values, indicating stronger city-dependent variation in model transfer, whereas RDE and POP are comparatively more stable. 

\label{sec:city_task_performance}
\begin{figure}[!tb]
    \centering
    \includegraphics[width=1\linewidth]{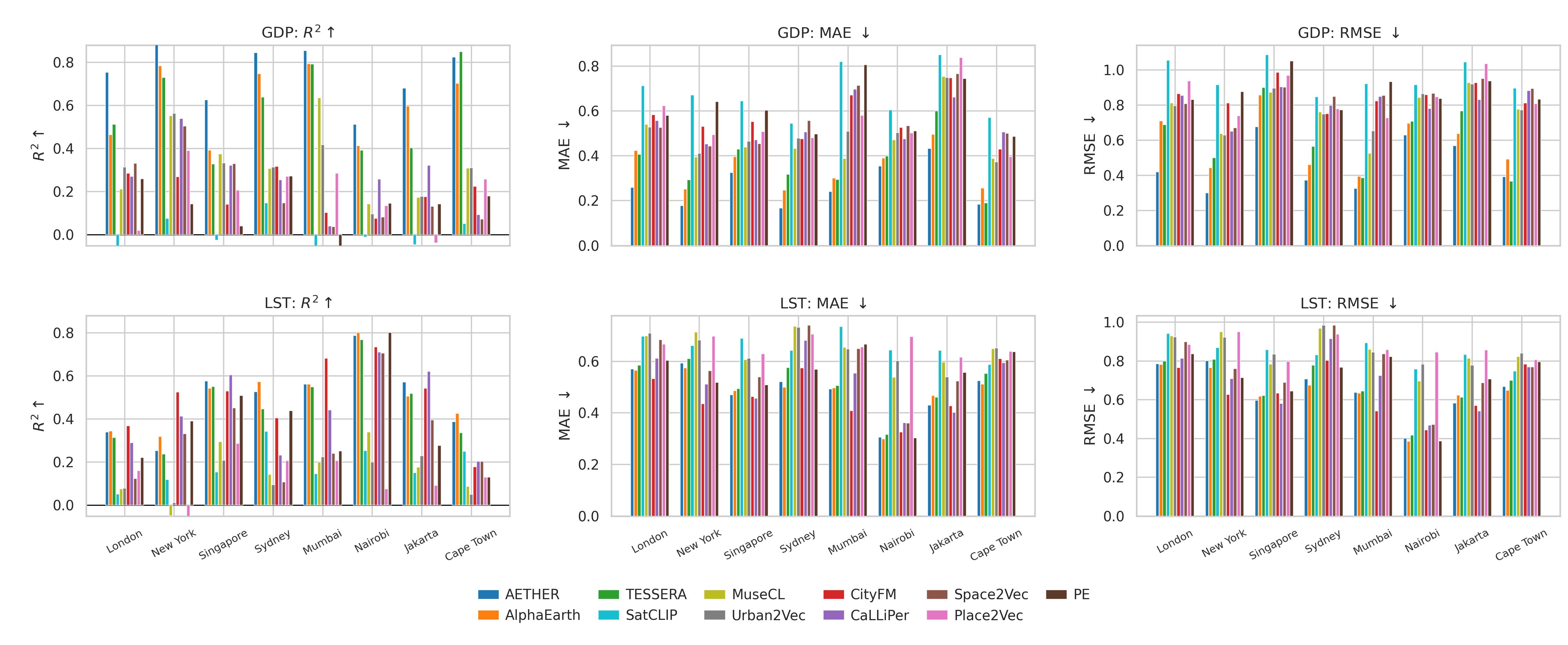}
    \caption{\textbf{GDP and LST prediction performance across cities.} The figure reports $R^2$, MAE, and RMSE for each model and city.}
    \label{fig:gdp_lst_city_performance}
\end{figure}

Figures~\ref{fig:gdp_lst_city_performance} and~\ref{fig:city_difficulty} show that benchmark difficulty varies strongly across both cities and tasks. Figure~\ref{fig:gdp_lst_city_performance} provides two representative examples, GDP and LST, while the full city-task results with error bars for all tasks and metrics are provided in Appendix~\ref{app:city-task-results}. Model rankings and the relative performance gaps among models change across cities. Thus, results from a single city would give an incomplete picture of model quality.

Figure~\ref{fig:city_difficulty} further shows that city-level difficulty is highly task-dependent. The city difficulty profiles cross substantially across tasks, indicating that no city is consistently easy or difficult for all prediction problems. For example, a city that is relatively easy for socioeconomic or built-environment prediction can be difficult for environmental prediction. This suggests that city-level generalization depends on the interaction between the target phenomenon, local urban structure, data quality, and the representation being evaluated. We examine whether these patterns persist beyond the eight  cities in an exploratory 26-city extension in Appendix~\ref{app:extended-26-cities}. We further analyze possible factors behind city--task difficulty variation in Appendix~\ref{app:city-difficulty-discussion}. Therefore, evaluating urban representations on a single city or a single task can obscure important differences in model robustness and transferability.

\subsection{The Effect of Spatial Splits}
\label{sec:spatial_split_effect}

\begin{figure}[!tb] 
    \centering

    \begin{minipage}[!tb]{0.48\linewidth}
        \centering
        \includegraphics[width=\linewidth]{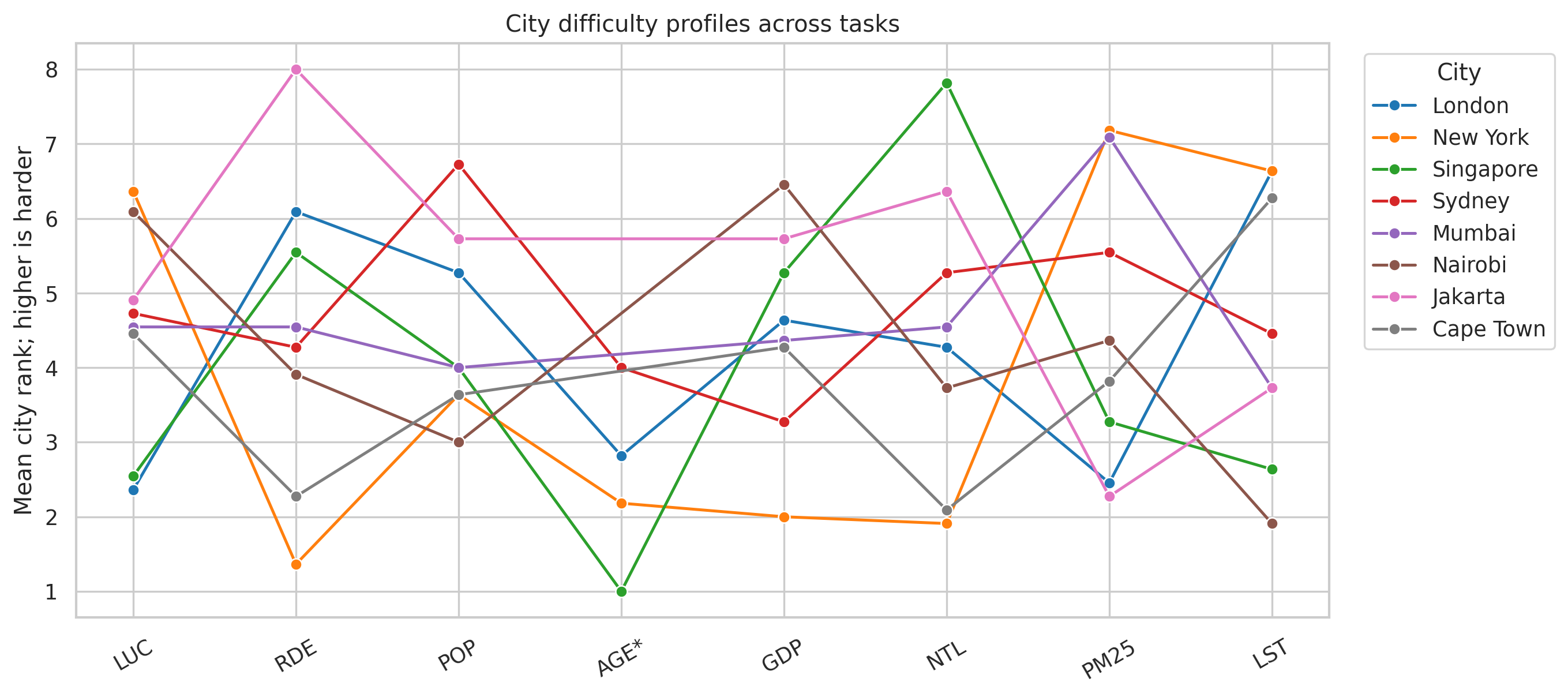}
        \caption{\textbf{City-level difficulty profiles across tasks.} Lower mean city rank indicates that a city is relatively easier to predict for a given task.}
        \label{fig:city_difficulty}
    \end{minipage}
    \hfill
    \begin{minipage}[!tb]{0.48\linewidth}
        \centering
        \includegraphics[width=\linewidth]{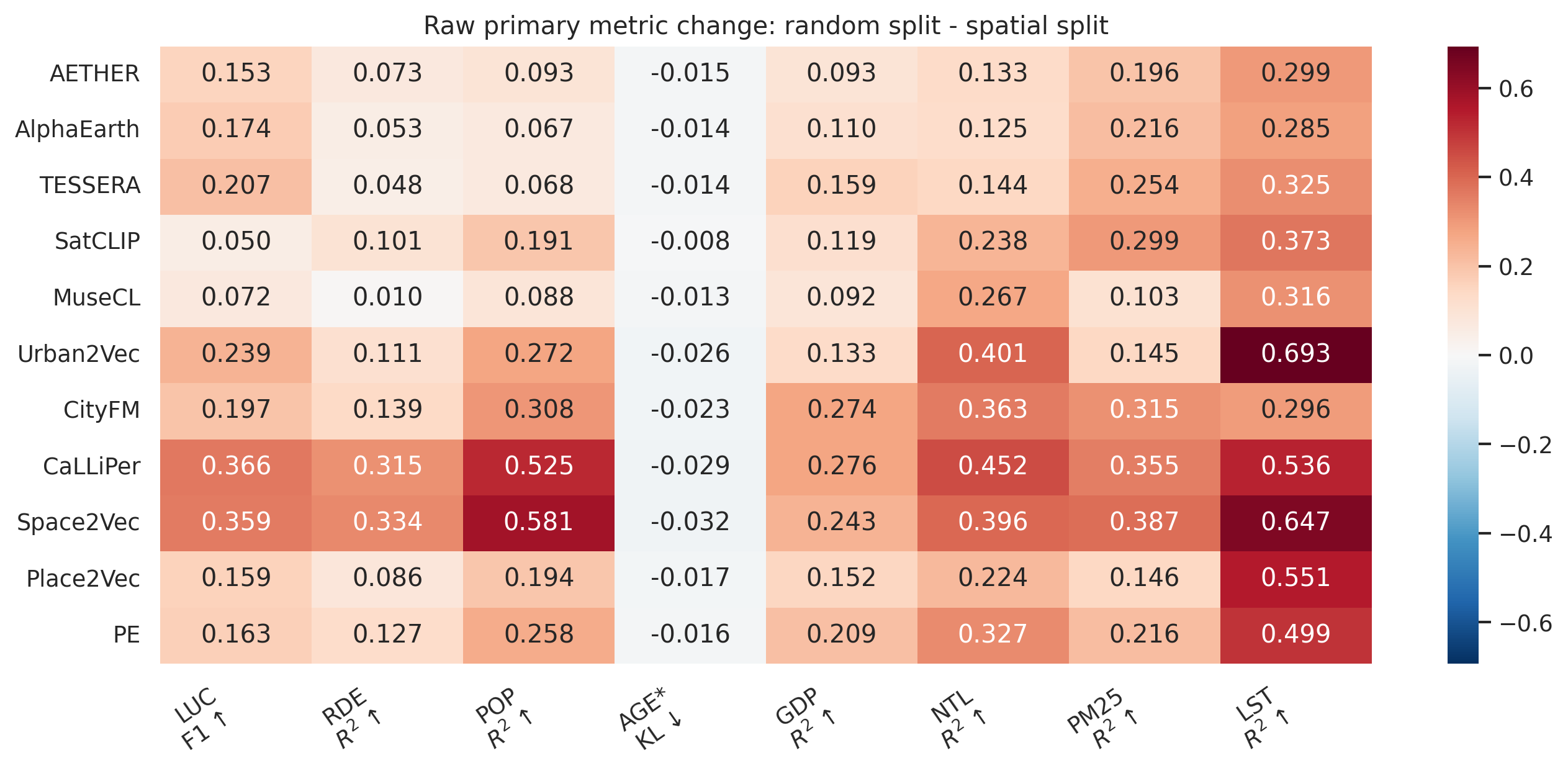}
        \caption{\textbf{Raw primary-metric changes under random splits compared with spatial splits.} Each cell reports random split minus spatial split for the primary evaluation metric. Positive values indicate higher raw metric values.}
        \label{fig:split_metric_delta}
    \end{minipage}

\end{figure}

Figure~\ref{fig:split_metric_delta} compares random splits with the spatial block split used in the main benchmark. Random splits consistently produce higher apparent performance for most tasks and models. This confirms that random splitting is overly optimistic in urban prediction because nearby training and test samples can share strong spatial autocorrelation.

The inflation is largest for densely sampled tasks such as population and LST, where random splits are more likely to place test samples close to training samples. It is smaller for coarser tasks such as GDP. Coordinate-based models such as CaLLiPer and Space2Vec show large gains under random splitting on dense targets, suggesting that smooth coordinate encoders can exploit local interpolation when spatial separation is not enforced. These patterns demonstrate that the effect of split protocol depends on both task resolution and model inductive bias.

This result supports the need for spatially structured evaluation. Random splits measure interpolation within observed urban areas, but they do not reliably measure transfer to unseen areas. CityRep therefore uses spatial block splits as the main protocol and treats random splits as a diagnostic for spatial leakage.

\section{Conclusion}
\label{sec:conclusion}

We presented CityRep, a unified benchmark for evaluating urban representations across heterogeneous spatial supports, varying data modalities, diverse cities, and comprehensive downstream tasks. Our results highlight three key findings: multi-city and multi-task evaluation is necessary because model performance varies substantially across urban contexts and target domains; spatially structured splits are essential because random splits can overestimate generalization through spatial leakage; and future urban representation learning could move toward larger-scale pretrained models that better integrate complementary modalities such as remote sensing, POIs, street-view, map entities, and locations. By making these comparisons systematic and reproducible, CityRep provides a foundation for measuring progress toward general-purpose urban representations and for developing urban foundation models.





\clearpage
{
\small
\bibliographystyle{unsrt}
\bibliography{main}
}
\clearpage

\appendix

\startcontents[appendices]

\section*{Appendix Contents}

\printcontents[appendices]{}{1}{\setcounter{tocdepth}{2}}

\clearpage

\section{Discussion}

\label{app:discussion}

\subsection{Limitations}
\label{app:limitations}

CityRep is designed to make urban representation evaluation broader and more comparable, but several limitations and scope boundaries remain.
First, the benchmark is constrained by data quality in both downstream task construction and baseline reproduction.
The downstream labels are derived from public geospatial products and city-specific land-use sources, whose uncertainty, update cycles, spatial resolutions, and semantic definitions vary across regions.
CityRep should therefore be interpreted as an evaluation of representations under the released data products, rather than as a definitive measurement of the underlying urban phenomena.
At the same time, reproducing existing representation models depends on the availability and quality of their required pretraining inputs.
Some original models rely on proprietary, city-specific, or unavailable modalities, so public reproductions may not capture every advantage of the original models under their original data environments.
Second, spatial splitting remains a methodological choice rather than a settled standard.
CityRep uses block-based spatial splits to reduce leakage and evaluate generalization to held-out urban areas, but other partitioning strategies are possible, such as administrative regions, spatial clustering, distance buffers, or task-specific regionalization.
The most appropriate split may depend on task resolution, spatial autocorrelation, city morphology, and the intended generalization target.
Further research is needed to understand how different spatial validation designs affect conclusions in urban representation learning.
Third, CityRep does not fully investigate how multimodal data scarcity shapes global urban embeddings.
Urban representations may depend on remote sensing, POIs, street-view imagery, road networks, mobility data, and administrative records, but these modalities are unevenly available across cities and regions.
Such an imbalance may affect what representations learn and may introduce geographic or socioeconomic biases into models intended for broad deployment.
In addition, the scope of this study is to evaluate representation at the intra-city level; the zero-shot cross-city generalization of future general-purpose urban models remains to be investigated further.
A systematic study of how missing or uneven multimodal data influences global urban representation learning remains an important direction for future work.

\subsection{Broader Impacts}
\label{app:broader-impacts}

CityRep is intended as an evaluation tool for understanding what urban representations encode and how reliably they transfer across tasks and cities. A standardized benchmark can reduce duplicated preprocessing effort, make model claims more comparable, and encourage evaluation protocols that better reflect spatial generalization. It may also help researchers identify where existing urban representations fail, especially in cities with limited data availability or urban forms underrepresented in current foundation models.

At the same time, urban prediction systems can influence planning, resource allocation, environmental assessment, and infrastructure decisions. Benchmark scores should therefore not be treated as evidence that a model is ready for direct policy deployment. Downstream use requires task-specific validation, uncertainty analysis, stakeholder review, and careful consideration of social context. The benchmark uses aggregated spatial labels, public geospatial layers, and derived task units; nevertheless, urban data can still encode socioeconomic inequalities and uneven data coverage. We therefore release CityRep as a research benchmark rather than a decision-making system, and we encourage users to report failures, biases, and city-specific limitations alongside aggregate scores.

\section{Details of Downstream Task Data}
\label{app:task-data}

\subsection{Raw Data}
\label{app:raw-data}

The released task data are processed into a common city--task registry, but the source layers retain different spatial supports, semantics, and access terms. Land use is a point-level classification task derived from city-specific zoning or land-use sources and harmonized into a shared taxonomy~\citep{nyc_dcp_pluto_25v4,ura_master_plan_2019_landuse_2025,nsw_dphi_epi_land_zoning,city_of_cape_town_zoning_2019,dki_jakarta_pergub_31_2022_rdtr,mcgm_mumbai_elu_2012,worldbank_nairobi_landuse_2010}. Because land-use redistribution terms vary by city, CityRep publicly releases harmonized land-use labels only where permitted by the corresponding source terms. Road density is computed from OpenStreetMap road geometries by aggregating drivable road length to the task grid~\citep{osm_road_network_extract_cityrep_2026}. Population and age distribution are derived from WorldPop products, with population represented as a scalar count and age represented as a distribution over age bins~\citep{worldpop_population_counts_r2024b,worldpop_agesex_r2024b}. GDP uses a gridded economic output product~\citep{kummu_gridded_gdp_2025}, nighttime lights use the VIIRS Nighttime Lights Annual V2.2 average masked radiance composite~\citep{eog_viirs_vnl_v22,elvidge2021viirs_vnl}, PM$_{2.5}$ uses global annual concentration grids~\citep{van_donkelaar_pm25_v5gl04_2024}, and LST uses MODIS/Terra daytime land surface temperature aggregated to an annual mean~\citep{wan_mod11a2_v061_2021}.

\begin{figure}[t]
    \centering
    \includegraphics[width=0.98\linewidth]{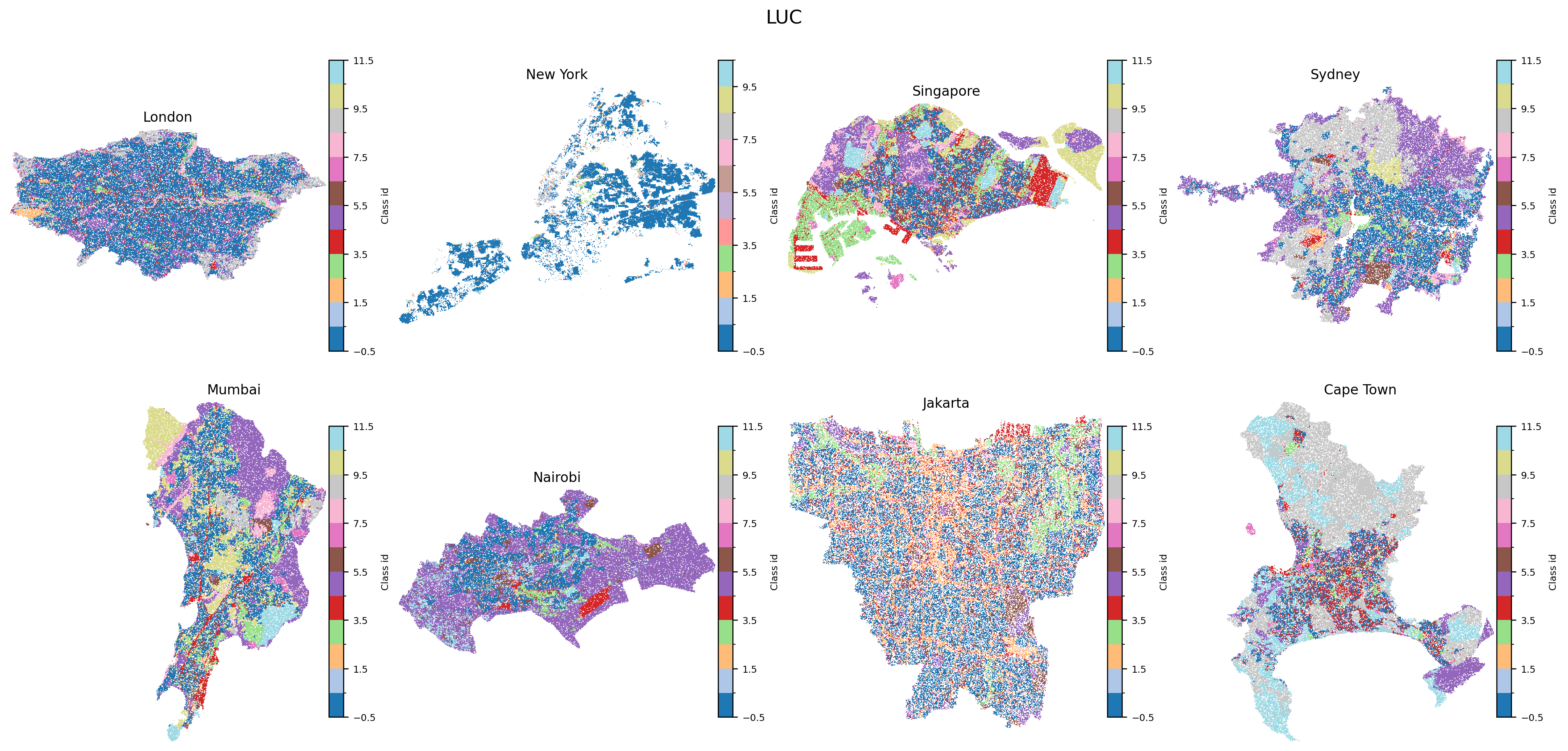}
    \caption{Land-use (LUC) labels across the eight benchmark cities.}
    \label{fig:app-raw-landuse}
\end{figure}

\begin{figure}[t]
    \centering
    \includegraphics[width=0.98\linewidth]{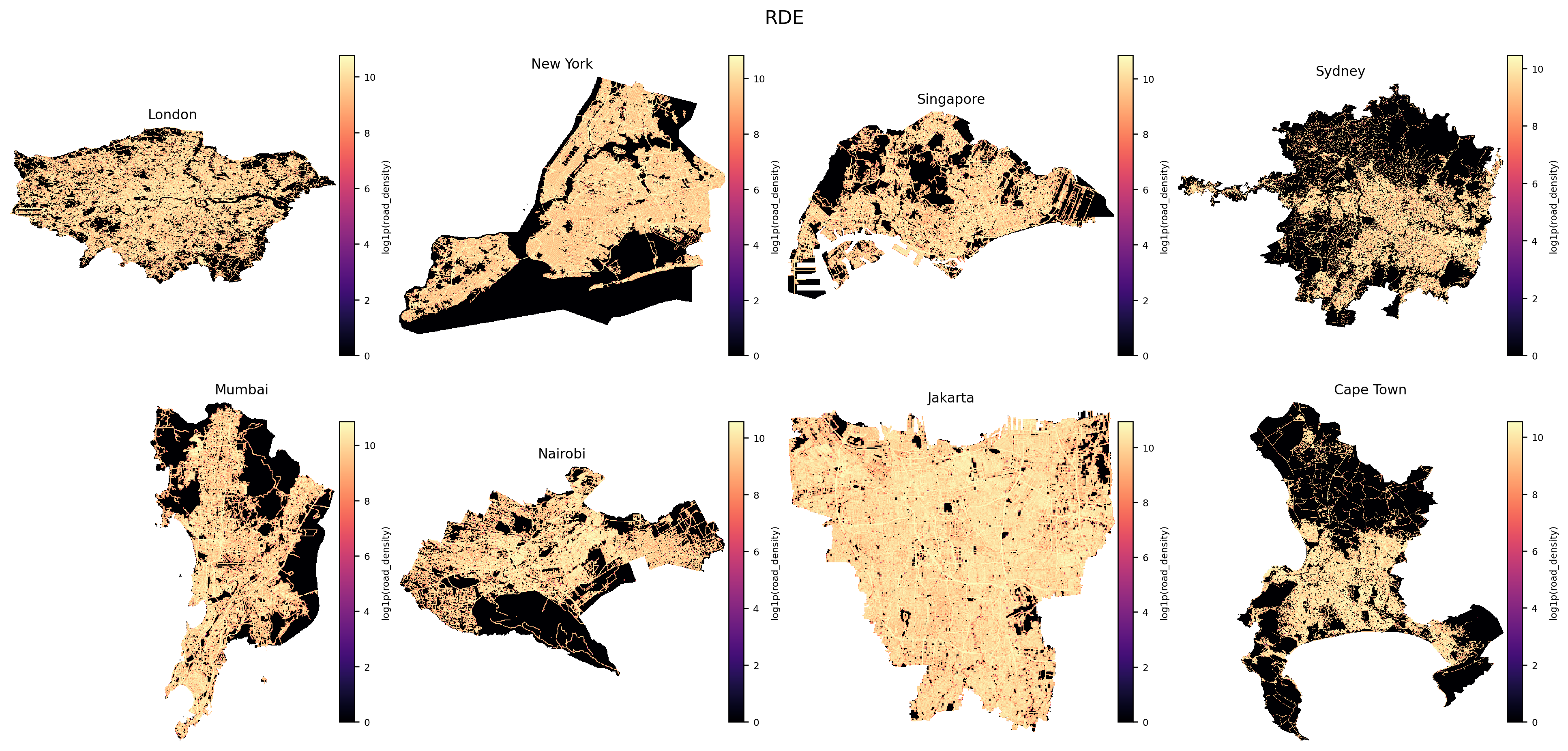}
    \caption{Road-density (RDE) labels across the eight benchmark cities.}
    \label{fig:app-raw-road-density}
\end{figure}

\begin{figure}[t]
    \centering
    \includegraphics[width=0.98\linewidth]{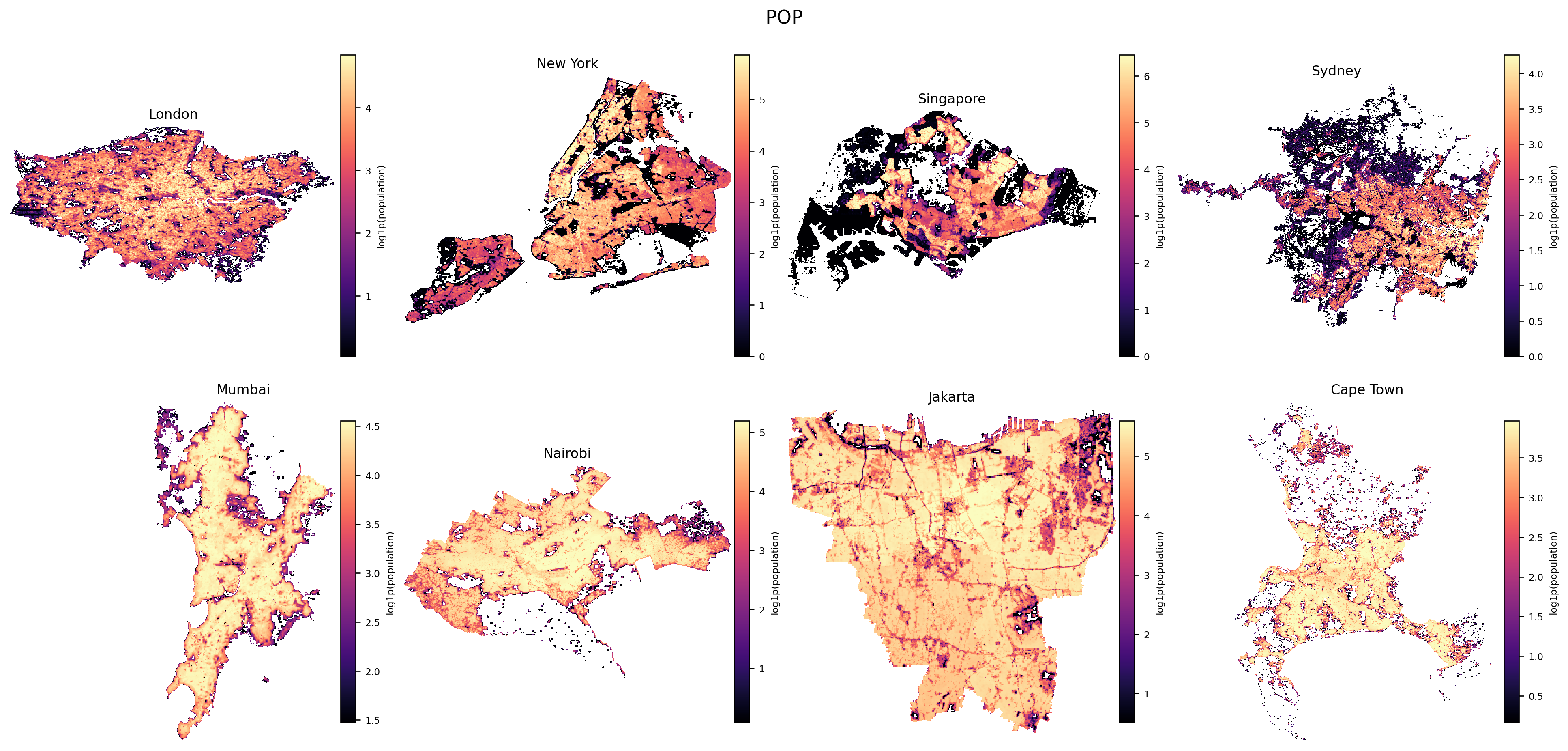}
    \caption{Population (POP) labels across the eight benchmark cities.}
    \label{fig:app-raw-population}
\end{figure}

\begin{figure}[t]
    \centering
    \includegraphics[width=0.98\linewidth]{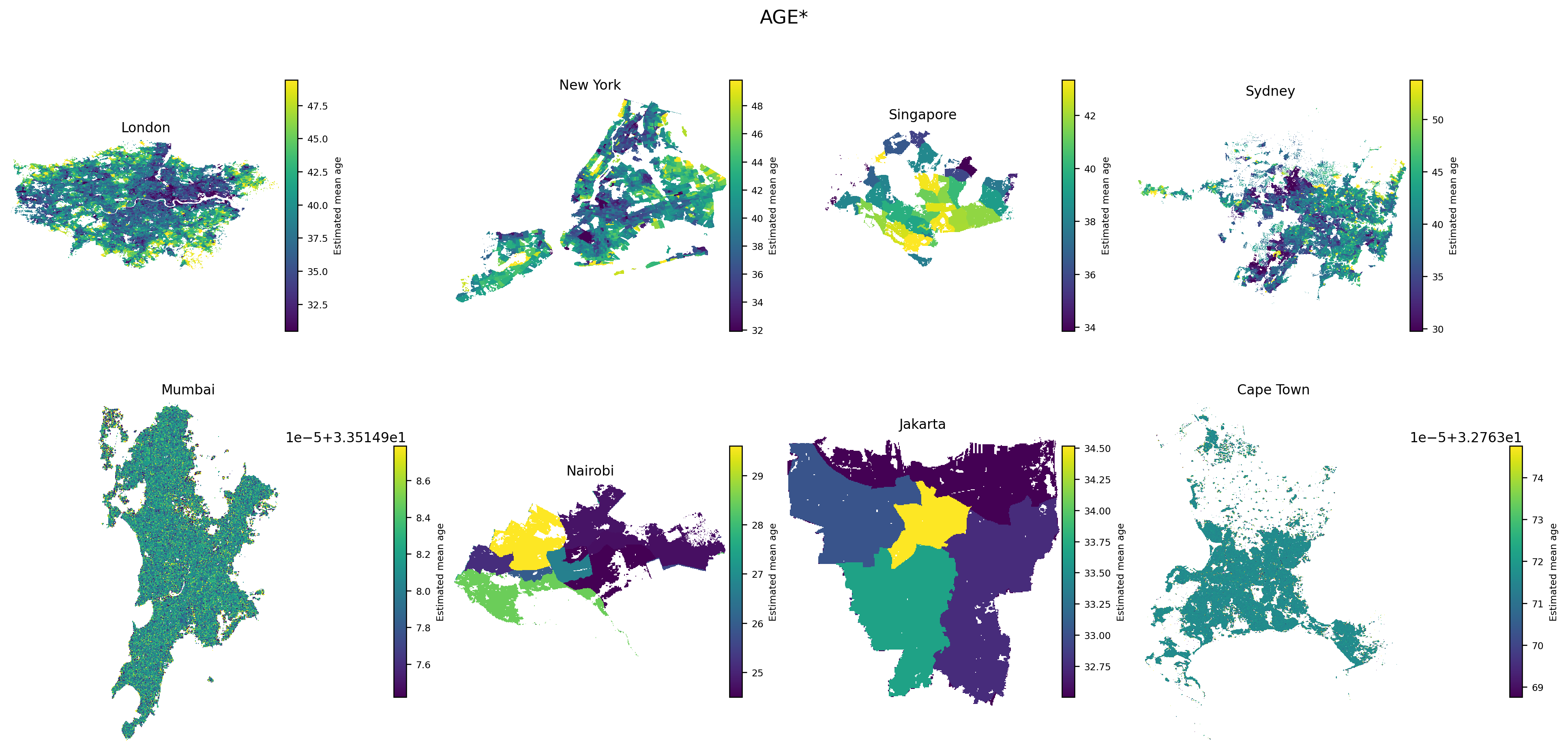}
    \caption{Age-distribution (AGE) labels across the eight benchmark cities.}
    \label{fig:app-raw-age}
\end{figure}

\begin{figure}[t]
    \centering
    \includegraphics[width=0.98\linewidth]{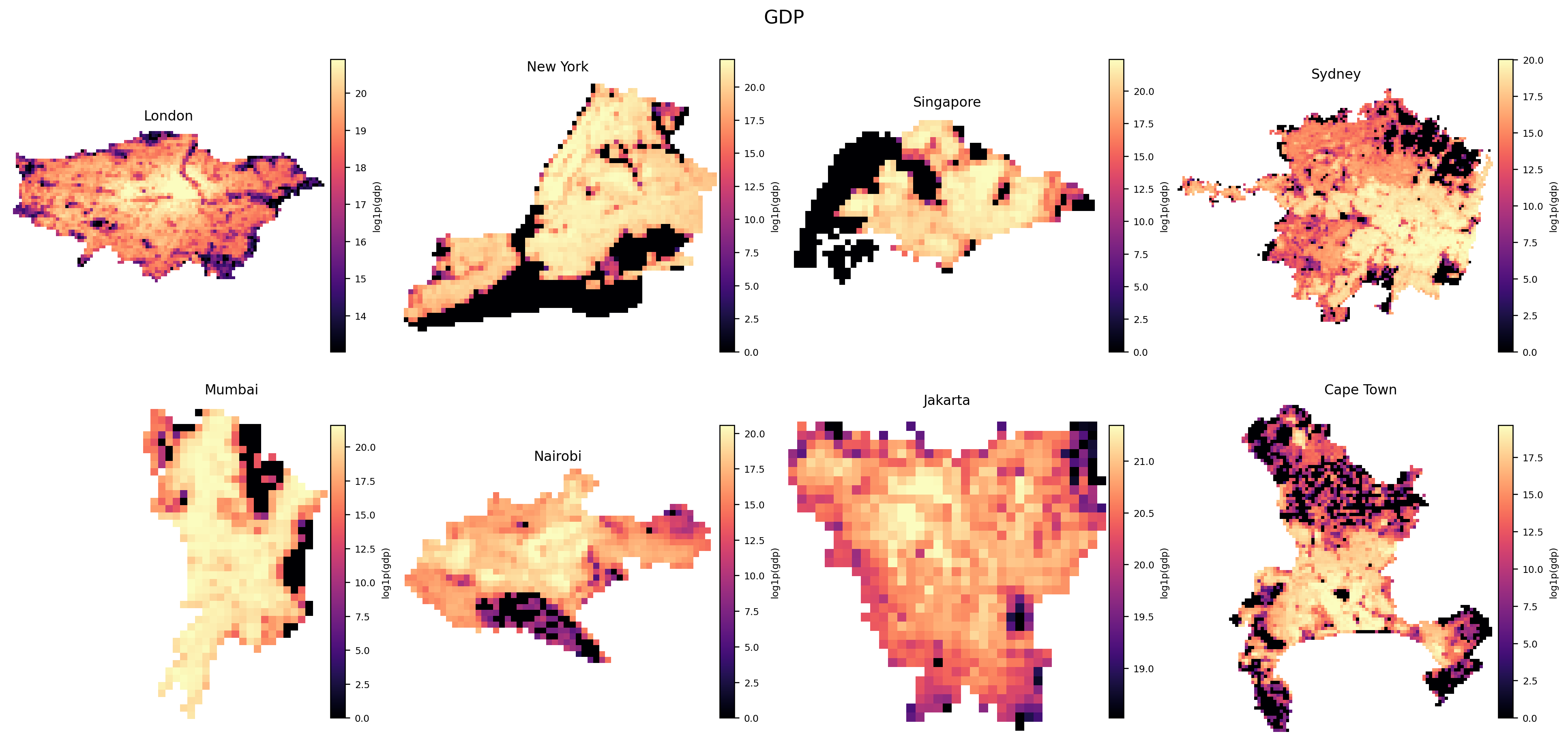}
    \caption{Gross Domestic Product (GDP) labels across the eight benchmark cities.}
    \label{fig:app-raw-gdp}
\end{figure}

\begin{figure}[t]
    \centering
    \includegraphics[width=0.98\linewidth]{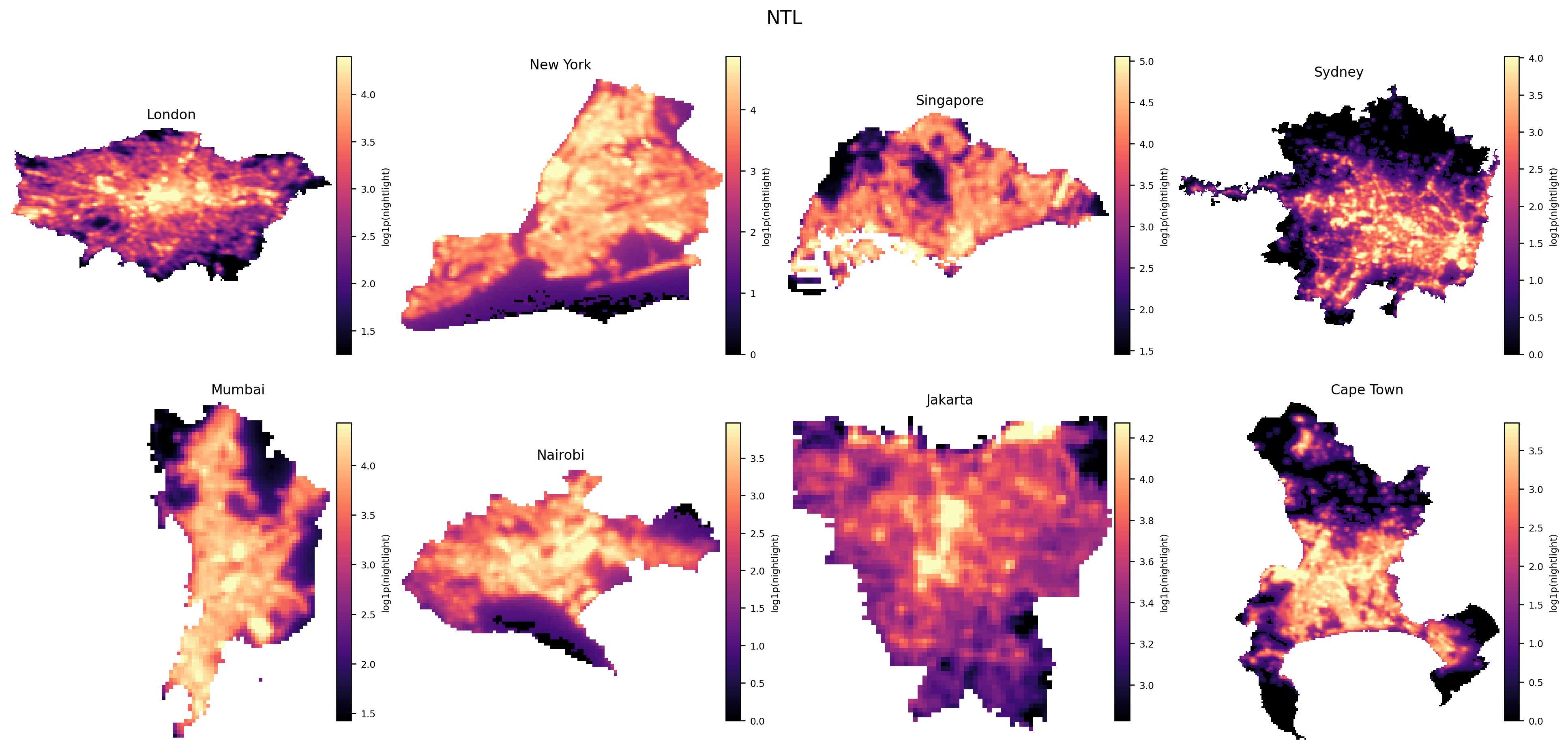}
    \caption{Nighttime lights (NTL) labels across the eight benchmark cities.}
    \label{fig:app-raw-ntl}
\end{figure}

\begin{figure}[t]
    \centering
    \includegraphics[width=0.98\linewidth]{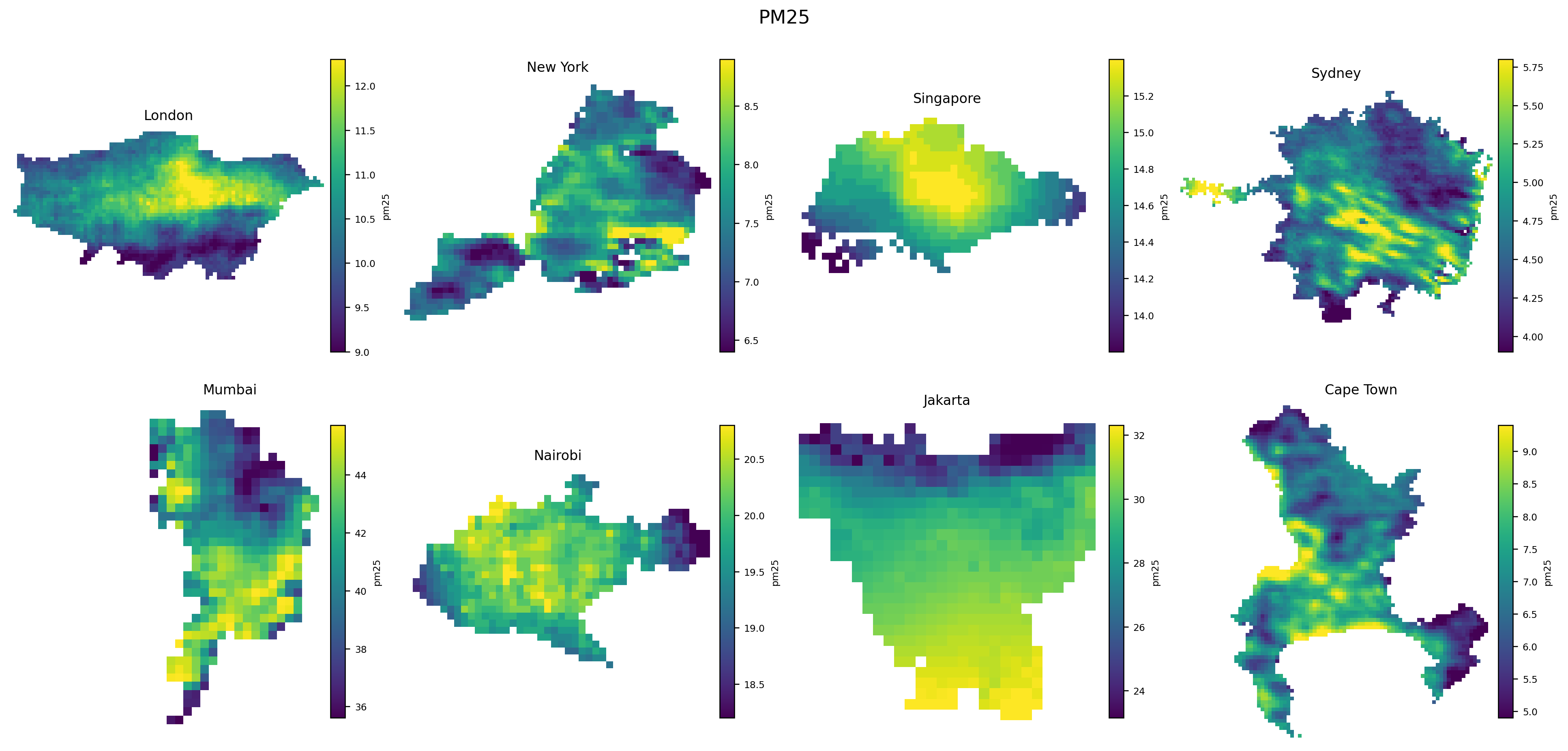}
    \caption{PM$_{2.5}$ labels across the eight benchmark cities.}
    \label{fig:app-raw-pm25}
\end{figure}

\begin{figure}[t]
    \centering
    \includegraphics[width=0.98\linewidth]{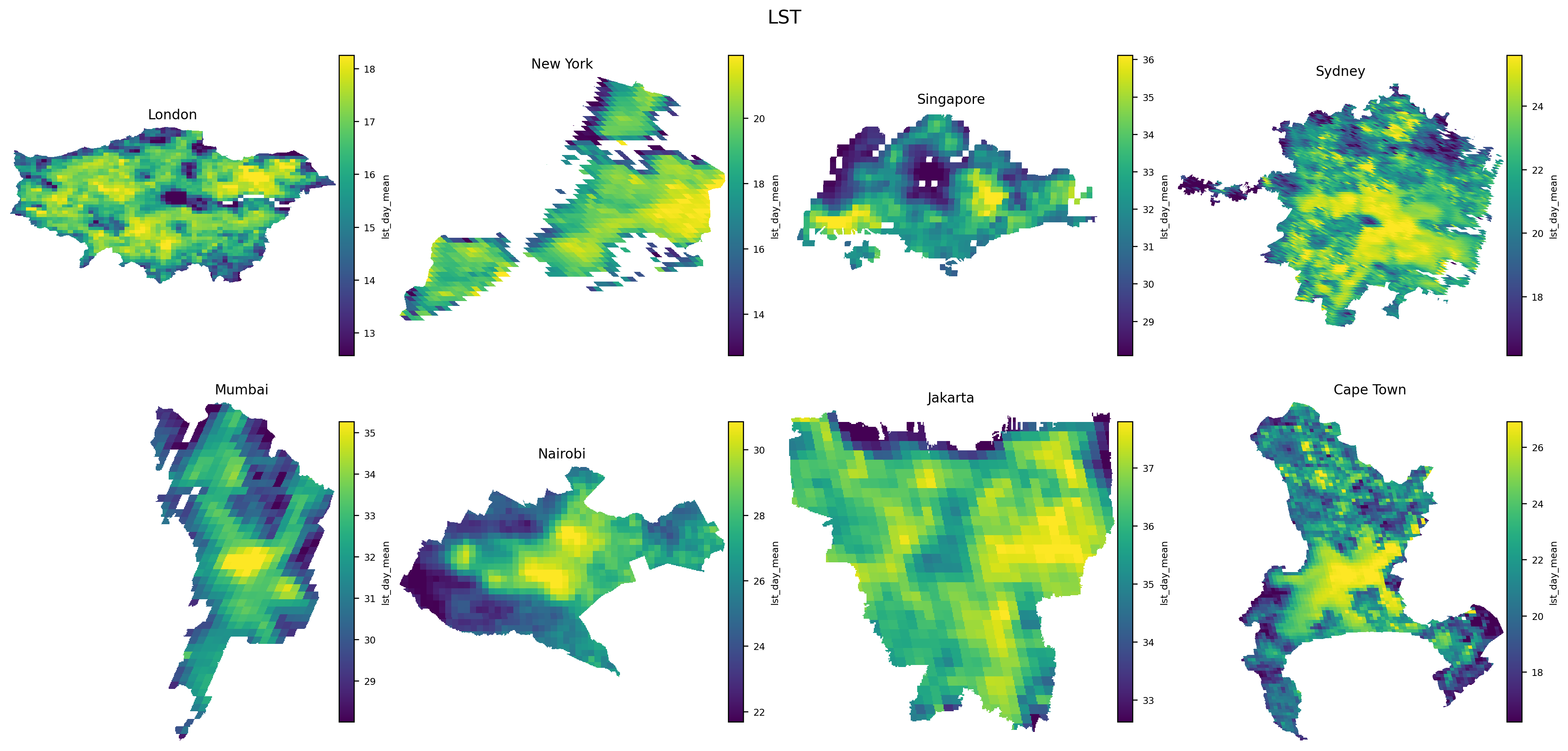}
    \caption{Land-surface-temperature labels across the eight benchmark cities.}
    \label{fig:app-raw-lst}
\end{figure}

Figures~\ref{fig:app-raw-landuse}--\ref{fig:app-raw-lst} visualize the processed downstream labels for all eight cities. The visualizations are included to show both the diversity of task supports and the differences in city extent. Some tasks are dense and fine grained, such as road density, population, and LST. Others are coarser, such as GDP, nighttime lights, and PM$_{2.5}$. CityRep preserves these task-specific supports instead of forcing all labels onto a single global grid, because the alignment module is responsible for mapping each representation to the target task unit.

Table~\ref{tab:app_raw_data_sources} summarizes the raw data sources used to construct the eight downstream tasks. Because the benchmark preserves task-specific spatial supports, the ``resolution/support'' column reports the native or effective support of the released task labels rather than forcing all sources into a single grid.

\begin{table}[t]
\centering
\caption{Raw data sources used to construct CityRep downstream tasks. Task abbreviations follow the main text: LUC = land-use classification, RDE = road-density estimation, POP = population prediction, AGE = age-distribution prediction, GDP = gross-domestic-product prediction, NTL = nighttime-lights prediction, PM$_{2.5}$ = PM$_{2.5}$ prediction, and LST = land-surface-temperature prediction. Native support describes the source product before CityRep processing; released support describes the spatial support used by the benchmark task files. Years in the raw-source column denote the actual source or CityRep task version used by the released benchmark files.}
\label{tab:app_raw_data_sources}
\scriptsize
\resizebox{\textwidth}{!}{
\begin{tabular}{p{0.08\linewidth}p{0.32\linewidth}p{0.18\linewidth}p{0.18\linewidth}p{0.18\linewidth}}
\toprule
Task & Raw source & Native support & Released support & License / terms \\
\midrule

\rowcolor{gray!15}\multicolumn{5}{l}{$\spadesuit$~\textit{Morphology}} \\

LUC
& City-specific official or public zoning / land-use layers ~\citep{nyc_dcp_pluto_25v4,ura_master_plan_2019_landuse_2025,nsw_dphi_epi_land_zoning,city_of_cape_town_zoning_2019,dki_jakarta_pergub_31_2022_rdtr,mcgm_mumbai_elu_2012,worldbank_nairobi_landuse_2010}
& Vector polygons or parcel-level records
& Harmonized sampled point labels, subject to source terms
& Varies by city; CityRep releases labels only where redistribution is permitted by source terms. \\

RDE
& OpenStreetMap drivable highway ways extracted for the CityRep 2026 road-density task version~\citep{osm_road_network_extract_cityrep_2026}
& Vector road geometries
& Road length density on the task grid
& ODbL 1.0; OpenStreetMap attribution required. \\

\rowcolor{gray!15}\multicolumn{5}{l}{$\heartsuit$~\textit{Demographics}} \\

POP
& WorldPop constrained population counts, R2024B, 2024~\citep{worldpop_population_counts_r2024b}
& 3 arc-second raster, approximately 100m at the equator
& Same task raster support
& CC BY 4.0. \\

AGE
& WorldPop age--sex structured population counts, R2024B, 2024~\citep{worldpop_agesex_r2024b}
& 3 arc-second rasters by age and sex group
& Age-bin distribution on the task raster support
& CC BY 4.0. \\

\rowcolor{gray!15}\multicolumn{5}{l}{$\diamondsuit$~\textit{Economy}} \\

GDP
& Kummu et al. gridded GDP total, 2024 band~\citep{kummu_gridded_gdp_2025}
& 30 arc-second gridded GDP product
& City-clipped task grid
& CC BY 4.0 for the public Kummu GDP product. \\

NTL
& VIIRS Nighttime Lights Annual V2.2 average masked radiance composite, 2024~\citep{eog_viirs_vnl_v22,elvidge2021viirs_vnl}
& Annual cloud-free radiance composite at approximately 15 arc-second resolution
& City-clipped nighttime-lights task grid
& Public domain / open EOG product; cite product page and source paper. \\

\rowcolor{gray!15}\multicolumn{5}{l}{$\clubsuit$~\textit{Environment}} \\

PM$_{2.5}$
& SEDAC/CIESIN Global Annual PM$_{2.5}$ Grids, V5.GL.04, 2022~\citep{van_donkelaar_pm25_v5gl04_2024}
& 0.01-degree WGS84 raster
& City-clipped task grid
& CIESIN open data policy; cite DOI and follow SEDAC/NASA use terms. \\

LST
& MODIS/Terra MOD11A2.061 8-day daytime land surface temperature annual mean, 2024~\citep{wan_mod11a2_v061_2021}
& 8-day, 1km MODIS sinusoidal grid
& Annual daytime mean on the CityRep task support
& NASA/LP DAAC data are openly shared without restriction; cite DOI. \\
\bottomrule
\end{tabular}
}
\end{table}

\clearpage
\subsection{Land Use Label Mapping}
\label{app:land-use-mapping}

We harmonize city-specific land-use labels into twelve benchmark classes:
Residential, Mixed Use, Commercial, Industrial, Transportation, Green /
Recreation, Institutional / Civic, Utilities, Water, Agriculture / Rural,
Vacant / Reserve, and Other. These classes are informed by established
land-use classification schemes, including the Land-Based Classification
Standards (LBCS) developed by the American Planning Association and partner
agencies \citep{apa_lbcs}, and the USGS Anderson land-use/land-cover
classification system \citep{anderson1976landuse}. Both schemes distinguish
broad urban functions such as residential, commercial, industrial,
transportation, utilities, institutional/public uses, recreation, agriculture,
and water. We adapt these high-level categories to a compact benchmark taxonomy
so that labels from heterogeneous municipal planning systems can be compared
across cities.

The mapping is city-specific because local planning systems use different
category names and different levels of detail. For example, a source category
may explicitly identify utilities in one city but group the same function under
public facilities in another. We therefore maintain a mapping table for each
city and export both a readable mapping table and audit files with source
categories, standardized labels, and counts.

We used an LLM-assisted procedure to produce the initial harmonization from
source labels to the twelve benchmark classes. The resulting mappings were then
manually audited by the authors. Ambiguous or inconsistent categories were
corrected before final task reconstruction. Categories that were duplicated or
semantically equivalent in the source mapping were also deduplicated. The
purpose of this review is not to impose a universal planning ontology, but to
ensure that the benchmark classes are consistent enough for cross-city
evaluation.

Table~\ref{tab:landuse-source-mapping-nosydney} reports the readable version of the city-specific label mapping used in the benchmark. For compactness, the Sydney mapping is omitted from this appendix table; the complete mapping, including Sydney, is provided in the released code.

%

\newcolumntype{L}[1]{>{\RaggedRight\arraybackslash}p{#1}}

\begingroup
\footnotesize
\setlength{\tabcolsep}{2.2pt}
\renewcommand{\arraystretch}{1.00}
\setlength{\LTleft}{0pt}
\setlength{\LTright}{0pt}
\setlength{\LTpre}{3pt}
\setlength{\LTpost}{3pt}
\emergencystretch=2em

\begin{longtable}{@{}L{0.090\linewidth}L{0.235\linewidth}L{0.150\linewidth}@{\hspace{0.014\linewidth}}L{0.090\linewidth}L{0.235\linewidth}L{0.150\linewidth}@{}}
\caption{City-specific land-use source labels mapping, excluding Sydney.}\label{tab:landuse-source-mapping-nosydney}\\
\toprule
\textbf{City} & \textbf{Source label} & \textbf{Mapped class} & \textbf{City} & \textbf{Source label} & \textbf{Mapped class} \\
\midrule
\endfirsthead

\caption[]{City-specific land-use source labels mapped to the common taxonomy, excluding Sydney (continued).}\\
\toprule
\textbf{City} & \textbf{Source label} & \textbf{Mapped class} & \textbf{City} & \textbf{Source label} & \textbf{Mapped class} \\
\midrule
\endhead

\midrule
\multicolumn{6}{r}{\footnotesize Continued on next page}\\
\endfoot

\bottomrule
\endlastfoot

London & High density residential with retail and commercial sites & Residential & London & Low density residential with amenities (suburbs and small villages / hamlets) & Residential \\
London & Medium density residential with high streets and amenities & Residential & London & Urban centres - mainly commercial/retail with residential pockets & Mixed Use \\
London & Business parks & Commercial & London & Large complex buildings various use (travel/recreation/ retail) & Commercial \\
London & Retail parks & Commercial & London & Industrial areas & Industrial \\
London & Principle Transport & Transportation & London & Coastal dunes & Green / Recreation \\
London & Coniferous and undifferentiated woodland & Green / Recreation & London & Deciduous woodland & Green / Recreation \\
London & Open or heath and moor land & Green / Recreation & London & Recreational land & Green / Recreation \\
London & Wetlands & Green / Recreation & London & Coastal water & Water \\
London & Inland Water & Water & London & Agriculture - mainly crops & Agriculture / Rural \\
London & Agriculture - mixed use & Agriculture / Rural & London & Farms & Agriculture / Rural \\
London & Glasshouses & Agriculture / Rural & London & Orchards & Agriculture / Rural \\
London & Mining and spoil areas & Other & New York & Multi-Family Elevator Buildings & Residential \\
New York & Multi-Family Walk-Up Buildings & Residential & New York & One \& Two Family Buildings & Residential \\
New York & Mixed Residential \& Commercial Buildings & Mixed Use & New York & Commercial \& Office Buildings & Commercial \\
New York & Industrial \& Manufacturing & Industrial & New York & Parking Facilities & Transportation \\
New York & Transportation \& Utility & Transportation & New York & Open Space \& Outdoor Recreation & Green / Recreation \\
New York & Public Facilities \& Institutions & Institutional / Civic & New York & Vacant Land & Vacant / Reserve \\
Singapore & RESIDENTIAL & Residential & Singapore & COMMERCIAL \& RESIDENTIAL & Mixed Use \\
Singapore & COMMERCIAL / INSTITUTION & Mixed Use & Singapore & RESIDENTIAL / INSTITUTION & Mixed Use \\
Singapore & RESIDENTIAL WITH COMMERCIAL AT 1ST STOREY & Mixed Use & Singapore & WHITE & Mixed Use \\
Singapore & COMMERCIAL & Commercial & Singapore & HOTEL & Commercial \\
Singapore & BUSINESS 1 & Industrial & Singapore & BUSINESS 1 - WHITE & Industrial \\
Singapore & BUSINESS 2 & Industrial & Singapore & BUSINESS 2 - WHITE & Industrial \\
Singapore & BUSINESS PARK & Industrial & Singapore & BUSINESS PARK - WHITE & Industrial \\
Singapore & LIGHT RAPID TRANSIT & Transportation & Singapore & MASS RAPID TRANSIT & Transportation \\
Singapore & PORT / AIRPORT & Transportation & Singapore & ROAD & Transportation \\
Singapore & TRANSPORT FACILITIES & Transportation & Singapore & BEACH AREA & Green / Recreation \\
Singapore & CEMETERY & Green / Recreation & Singapore & OPEN SPACE & Green / Recreation \\
Singapore & PARK & Green / Recreation & Singapore & SPORTS \& RECREATION & Green / Recreation \\
Singapore & CIVIC \& COMMUNITY INSTITUTION & Institutional / Civic & Singapore & EDUCATIONAL INSTITUTION & Institutional / Civic \\
Singapore & HEALTH \& MEDICAL CARE & Institutional / Civic & Singapore & PLACE OF WORSHIP & Institutional / Civic \\
Singapore & UTILITY & Utilities & Singapore & WATERBODY & Water \\
Singapore & AGRICULTURE & Agriculture / Rural & Singapore & RESERVE SITE & Vacant / Reserve \\
Singapore & SPECIAL USE & Other & Mumbai & Residential & Residential \\
Mumbai & Slum / Cluster & Residential & Mumbai & Urban Villages & Residential \\
Mumbai & Commercial Activities & Commercial & Mumbai & Informal Market & Commercial \\
Mumbai & Municipal Market & Commercial & Mumbai & Other Offices & Commercial \\
Mumbai & Industrial Use & Industrial & Mumbai & Transport & Transportation \\
Mumbai & Cemetery & Green / Recreation & Mumbai & Natural Areas & Green / Recreation \\
Mumbai & Open Spaces & Green / Recreation & Mumbai & Swimming Pool & Green / Recreation \\
Mumbai & Educational Amenities & Institutional / Civic & Mumbai & Government Office & Institutional / Civic \\
Mumbai & Law and Order & Institutional / Civic & Mumbai & Medical Amenities & Institutional / Civic \\
Mumbai & Municipal Chowkies & Institutional / Civic & Mumbai & Municipal Office & Institutional / Civic \\
Mumbai & Social Amenities & Institutional / Civic & Mumbai & Communication & Utilities \\
Mumbai & Public Utility and Facity & Utilities & Mumbai & Town Duty / Octroi Office & Utilities \\
Mumbai & Water & Water & Mumbai & Primary Activity & Agriculture / Rural \\
Mumbai & Area Under SPA & Vacant / Reserve & Mumbai & Under Construction & Vacant / Reserve \\
Mumbai & Vacant & Vacant / Reserve & Mumbai & Unclassified & Other \\
Nairobi & res\_slum & Residential & Nairobi & residential & Residential \\
Nairobi & mixed CI & Mixed Use & Nairobi & mixed RC & Mixed Use \\
Nairobi & commercial & Commercial & Nairobi & industrial & Industrial \\
Nairobi & transportation & Transportation & Nairobi & open space & Green / Recreation \\
Nairobi & open\_space & Green / Recreation & Nairobi & recreational & Green / Recreation \\
Nairobi & institutional & Institutional / Civic & Nairobi & water & Water \\
Nairobi & no\_structures & Vacant / Reserve & Nairobi & unknown & Other \\
Jakarta & Zona Perumahan & Residential & Jakarta & Zona Pariwisata & Commercial \\
Jakarta & Zona Perdagangan dan Jasa & Commercial & Jakarta & Zona Perkantoran & Commercial \\
Jakarta & Zona Kawasan Peruntukan Industri & Industrial & Jakarta & Zona Badan Jalan & Transportation \\
Jakarta & Zona Transportasi & Transportation & Jakarta & Zona Ekosistem Mangrove & Green / Recreation \\
Jakarta & Zona Hutan Lindung & Green / Recreation & Jakarta & Zona Hutan produksi & Green / Recreation \\
Jakarta & Zona Konservasi & Green / Recreation & Jakarta & Zona Perlindungan Setempat & Green / Recreation \\
Jakarta & Zona Ruang Terbuka Hijau & Green / Recreation & Jakarta & Zona Pelayanan Umum & Institutional / Civic \\
Jakarta & Zona Pertahanan dan Keamanan & Institutional / Civic & Jakarta & Zona Pembangkit Tenaga Listrik & Utilities \\
Jakarta & Zona Badan Air & Water & Jakarta & Zona Perikanan & Agriculture / Rural \\
Jakarta & Zona Pertanian & Agriculture / Rural & Jakarta & 0 & Other \\
Cape Town & General Residential 1 : Group Housing & Residential & Cape Town & General Residential 2 & Residential \\
Cape Town & General Residential 3 & Residential & Cape Town & General Residential 4 & Residential \\
Cape Town & General Residential 5 & Residential & Cape Town & General Residential 6 & Residential \\
Cape Town & Residential 1 : Conventional Housing & Residential & Cape Town & Residential 2 : Incremental Housing & Residential \\
Cape Town & Mixed Use 1 & Mixed Use & Cape Town & Mixed Use 2 & Mixed Use \\
Cape Town & Mixed Use 3 & Mixed Use & Cape Town & General Business 1 & Commercial \\
Cape Town & General Business 2 & Commercial & Cape Town & General Business 3 & Commercial \\
Cape Town & General Business 4 & Commercial & Cape Town & General Business 5 & Commercial \\
Cape Town & General Business 6 & Commercial & Cape Town & General Business 7 & Commercial \\
Cape Town & Local Business 1 : Intermediate Business & Commercial & Cape Town & Local Business 2 : Local Business & Commercial \\
Cape Town & General Industrial 1 & Industrial & Cape Town & General Industrial 2 & Industrial \\
Cape Town & Risk Industry & Industrial & Cape Town & Transport 1 : Transport Use & Transportation \\
Cape Town & Transport 2 : Public Road and Public Parking & Transportation & Cape Town & Open Space 1 : Environmental Conservation & Green / Recreation \\
Cape Town & Open Space 2 : Public Open Space & Green / Recreation & Cape Town & Open Space 3: Special Open Space & Green / Recreation \\
Cape Town & Community 1 : Local & Institutional / Civic & Cape Town & Community 2 : Regional & Institutional / Civic \\
Cape Town & Utility & Utilities & Cape Town & Agricultural & Agriculture / Rural \\
Cape Town & Rural & Agriculture / Rural & Cape Town & Limited Use Zone & Other \\
Cape Town & None & Other &  &  &  \\
\end{longtable}

\endgroup

\section{Additional Experiment Setting Details}
\label{app:experiment-setting}

\subsection{Spatial Split}
\label{app:spatial-split}

The main benchmark uses spatial block splitting because urban labels are strongly spatially autocorrelated. If nearby samples from the same neighborhood are placed into both training and test sets, a downstream predictor can achieve high performance through local interpolation rather than through representation transfer. This is especially problematic for dense raster tasks such as population, road density, and land surface temperature, where adjacent cells often share similar values.

\begin{figure}[t]
    \centering
    \includegraphics[width=\linewidth]{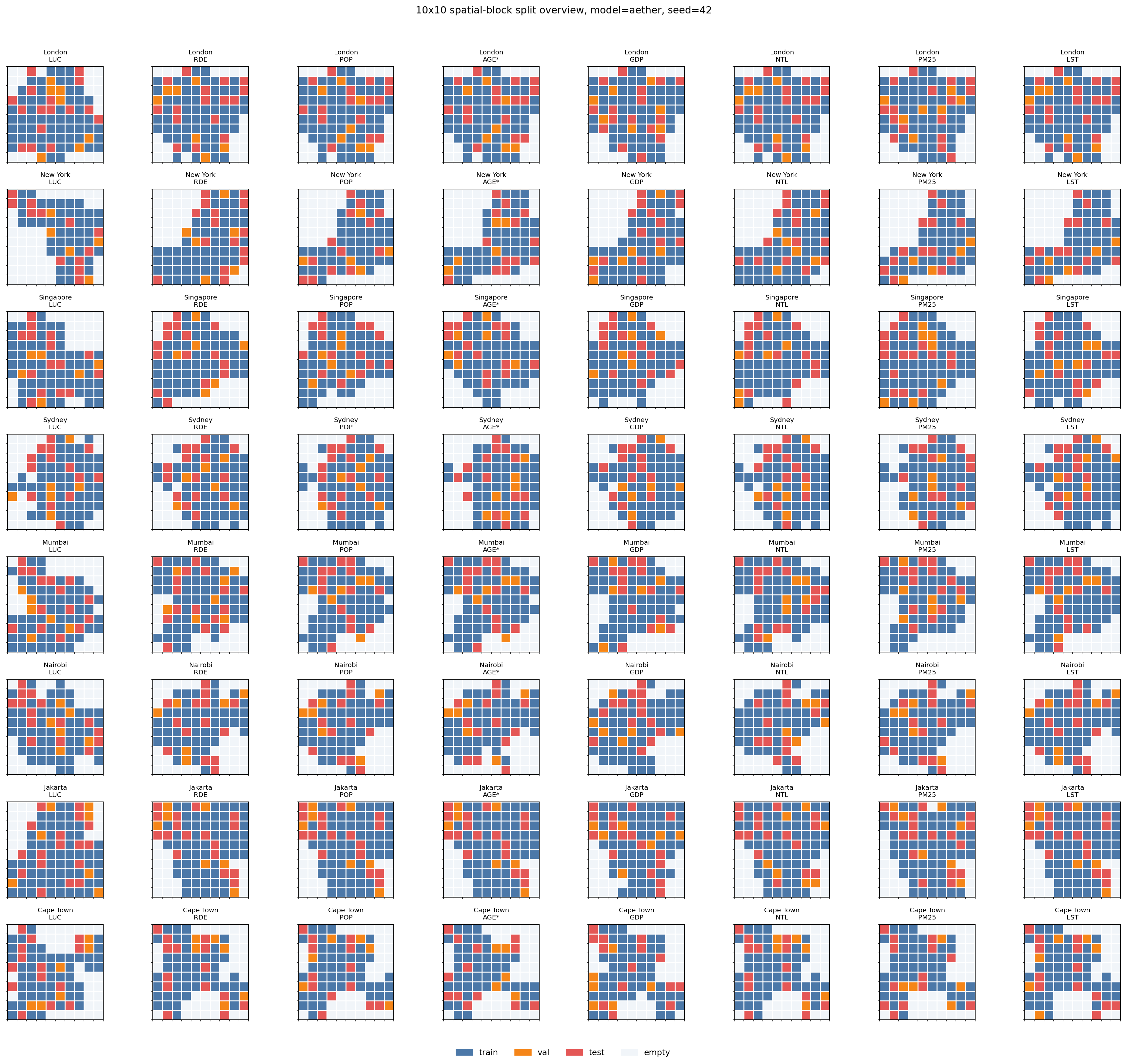}
    \caption{Example 10-by-10 spatial block split assignments for AETHER under seed 42. Blocks, rather than individual samples, are assigned to train, validation, and test splits.}
    \label{fig:app-spatial-block-split}
\end{figure}

\begin{figure}[t]
    \centering
    \includegraphics[width=\linewidth]{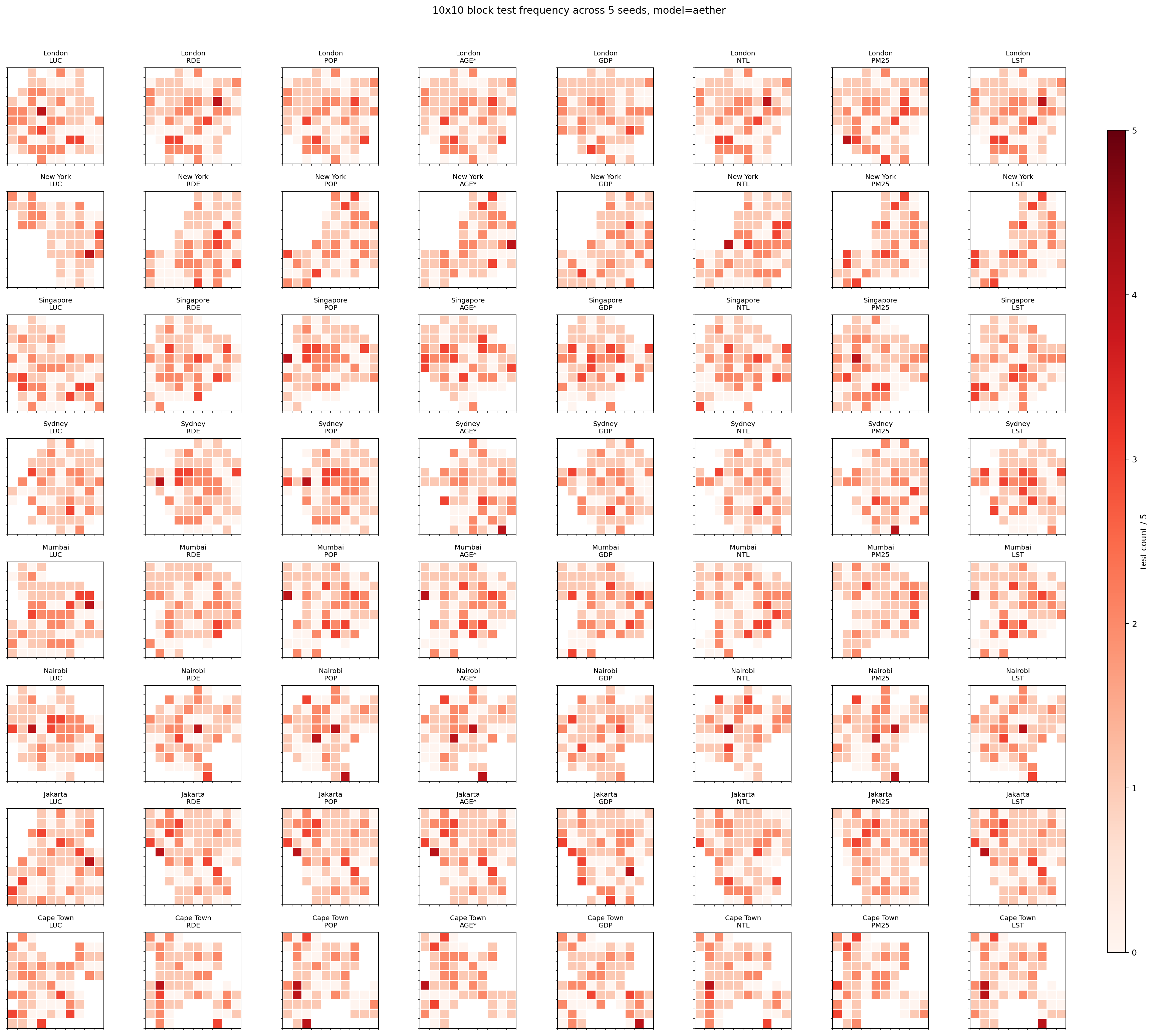}
    \caption{Test-block frequency across the five spatial split seeds. The figure shows how often each spatial block is assigned to the test partition.}
    \label{fig:app-spatial-test-frequency}
\end{figure}

For each city--task dataset, we construct a $10 \times 10$ block grid over the valid task extent. Blocks are assigned, rather than individual task units, to train, validation, and test partitions. The test partition contains 20\% of the blocks, and 10\% of the remaining training blocks are used for validation. We repeat the procedure with five seeds, $\{42,24,7,0,100\}$, and all models evaluated on the same city--task pair use the same split assignments. Figure~\ref{fig:app-spatial-block-split} illustrates the resulting block layout for the main protocol, and Figure~\ref{fig:app-spatial-test-frequency} shows how often each block is used as test across seeds.

\begin{figure}[t]
    \centering
    \includegraphics[width=\linewidth]{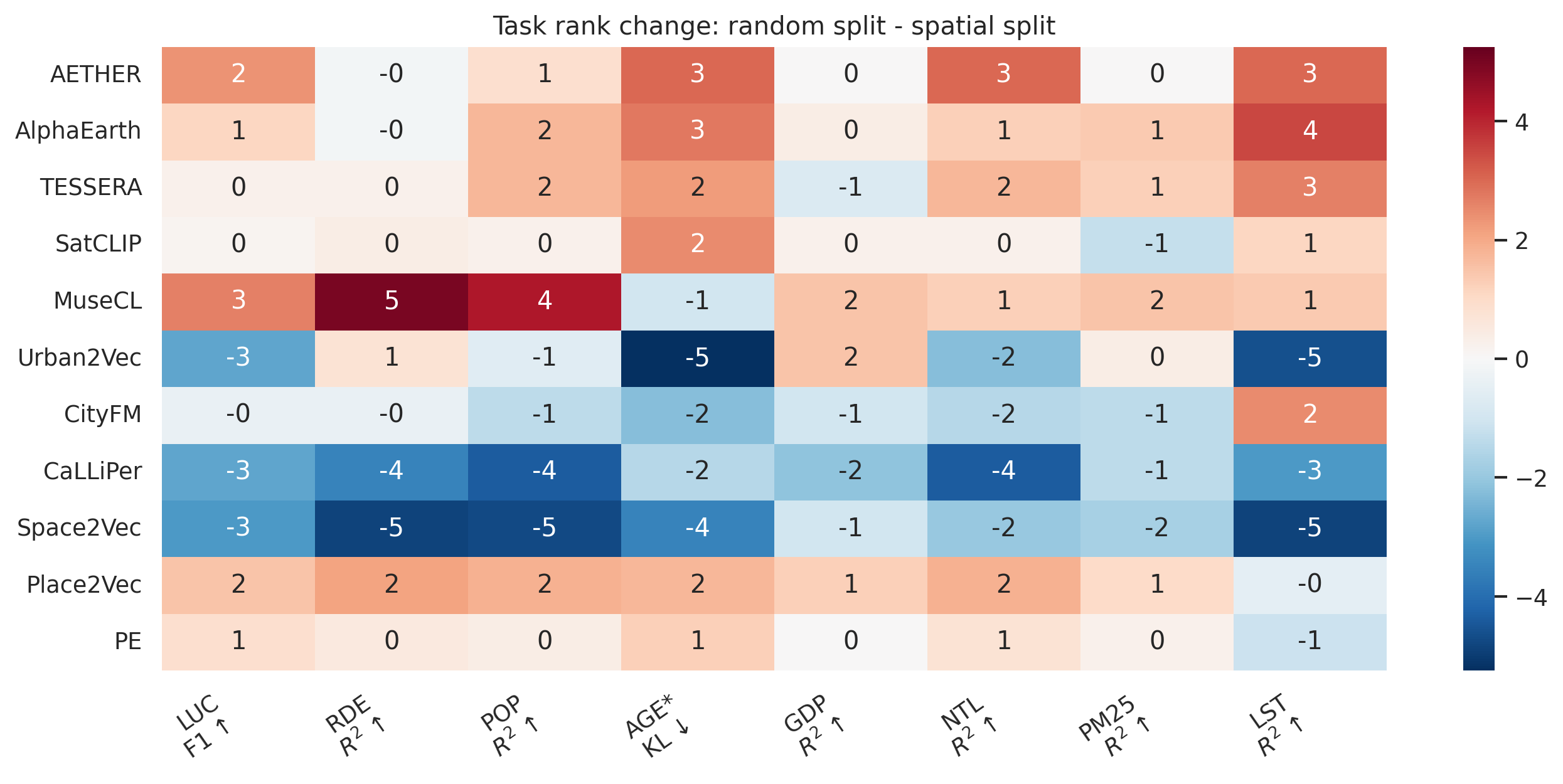}
    \caption{Task-level rank changes between random and spatial splits by model and task. Negative values indicate better rank under random splitting.}
    \label{fig:app-split-task-rank-delta-heatmap}
\end{figure}

\begin{figure}[t]
    \centering
    \includegraphics[width=\linewidth]{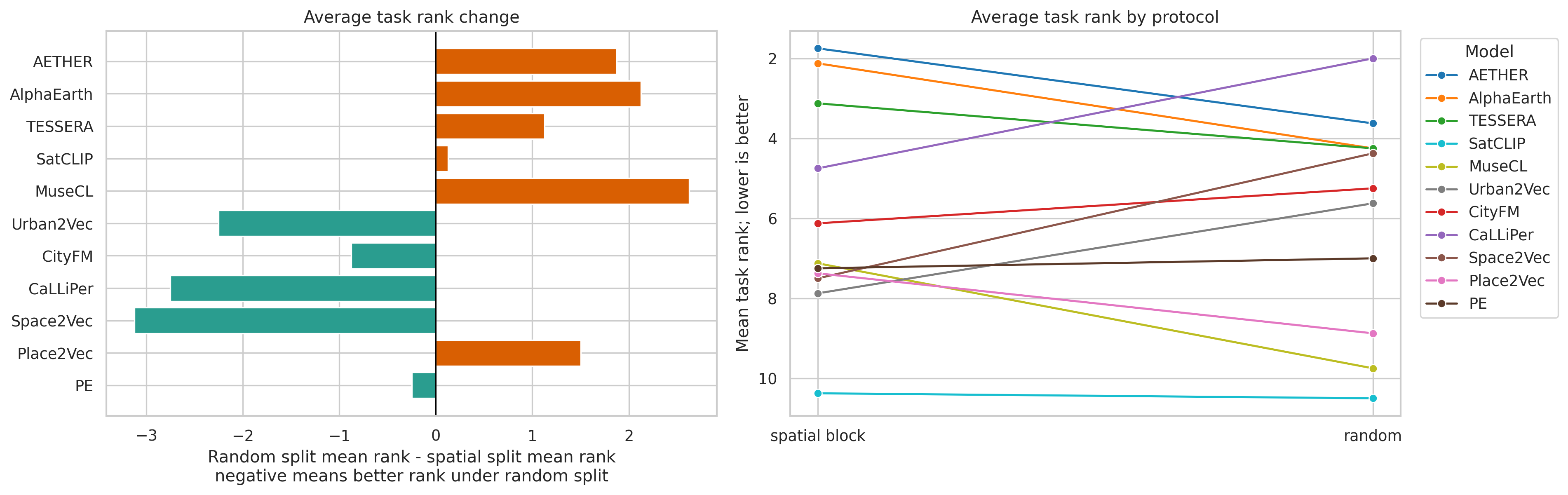}
    \caption{Overall mean-rank change between random and spatial splits across tasks. Negative values indicate better rank under random splitting.}
    \label{fig:app-split-overall-rank-change-bar}
\end{figure}

We also run a random split with the same ratios and seeds as a diagnostic comparison. Figures~\ref{fig:app-split-task-rank-delta-heatmap} and~\ref{fig:app-split-overall-rank-change-bar} show that random splitting often increases apparent performance and can alter rankings. This motivates our choice to use spatial splits as the main benchmark protocol and to treat random splits as a measure of spatial leakage sensitivity.

\clearpage

\clearpage

\subsection{Details of Baselines Reproduction}
\begingroup
\scriptsize
\setlength{\tabcolsep}{4pt}
\renewcommand{\arraystretch}{1.18}

\begin{xltabular}{\textwidth}{
  >{\raggedright\arraybackslash}p{0.12\textwidth}
  >{\raggedright\arraybackslash}p{0.17\textwidth}
  >{\raggedright\arraybackslash}X
  >{\raggedright\arraybackslash}p{0.20\textwidth}
  >{\raggedright\arraybackslash}p{0.19\textwidth}
}
\caption{Reproduction and interface details for the eleven CityRep baselines. The table distinguishes released feature products, official checkpoints, and CityRep-compatible reproductions. ``Dim.'' is the feature dimension used by the downstream evaluator. ``Support'' is the spatial interface registered in CityRep before alignment to each task. All models are evaluated with the same downstream predictor, spatial split protocol, and city--task files.}
\label{tab:app_baseline_reproduction} \\

\toprule
Model & Reproduction status & CityRep input data & Dim. / support & Checkpoint, feature, or code source \\
\midrule
\endfirsthead

\toprule
Model & Reproduction status & CityRep input data & Dim. / support & Checkpoint, feature, or code source \\
\midrule
\endhead

\midrule
\multicolumn{5}{r}{\textit{Continued on next page}} \\
\endfoot

\bottomrule
\endlastfoot

\rowcolor{gray!15}
\multicolumn{5}{l}{\textit{Remote-sensing raster representations}} \\
\midrule

AlphaEarth
& Released feature product
& Google AlphaEarth / Satellite Embedding annual raster product, 2024.
& 64-d raster. Native raster sampled at land-use points; task-grid rasters use mean aggregation or coordinate sampling.
& Official feature product from Google Earth Engine / Source Cooperative AlphaEarth Foundations COGs; CityRep wrapper code is used for cropping and task-grid export. \\
\multicolumn{5}{p{0.96\textwidth}}{\textit{Implementation note.}
AlphaEarth is evaluated as an off-the-shelf released feature product. Results reflect the public remote-sensing embedding together with CityRep's raster alignment rule.} \\
\midrule

TESSERA
& Official package / feature export
& TESSERA / GeoTessera 2024 remote-sensing embeddings from Sentinel-1/Sentinel-2 temporal-spectral inputs.
& 128-d raster. Reference-grid raster for point sampling; task-grid rasters for raster tasks.
& Official GeoTessera code/package: \url{https://github.com/ucam-eo/geotessera}; official pretrained embedding export. \\
\multicolumn{5}{p{0.96\textwidth}}{\textit{Implementation note.}
TESSERA is evaluated as frozen remote-sensing features exported with the official package/API. CityRep standardizes the task-grid aggregation, point sampling, and downstream evaluator.} \\
\midrule

AETHER
& CityRep-trained checkpoint
& AlphaEarth 64-d raster embeddings plus Foursquare POI semantic text; POI text encoded with \texttt{qwen-3} and projected into the shared space.
& 128-d raster. Native AETHER rasters and task-specific mean rasters registered for all 8 cities and 8 tasks.
& Official AETHER code: \url{https://github.com/inwind0212/AETHER}; CityRep AETHER code and checkpoints released with the benchmark artifacts;  \\
\multicolumn{5}{p{0.96\textwidth}}{\textit{Implementation note.}
AETHER is the CityRep multimodal representation baseline. It combines remote-sensing and POI semantic signals under the same alignment and downstream evaluation protocol as the other baselines.} \\
\midrule

\rowcolor{gray!15}
\multicolumn{5}{l}{\textit{Map, POI, street-view, and region representations}} \\
\midrule

CityFM
& Mixed official and reproduced
& OSM-derived road, polygon, tag, and surface features prepared in CityFM format.
& 2560-d H3 region table, resolution 8. Task samples are mapped to H3 cells.
& Official CityFM code: \url{https://github.com/PasqualeTurin/CityFM}; official weights used where available, with CityRep-compatible training/export for other cities. \\
\multicolumn{5}{p{0.96\textwidth}}{\textit{Implementation note.}
We use official CityFM weights where available and produce CityFM-compatible embeddings for the remaining benchmark cities with the same registered input interface. The released export uses components that can be prepared consistently across all CityRep cities.} \\
\midrule

Place2Vec
& CityRep-compatible reproduction
& Foursquare Open Places, second-level POI categories.
& 128-d H3 region table, resolution 8. POI category embeddings are pooled into H3 cells.
& Reproduce code according to the paper; CityRep wrapper prepares Foursquare POIs and H3 exports. \\
\multicolumn{5}{p{0.96\textwidth}}{\textit{Implementation note.}
CityRep uses Foursquare Open Places as the common POI source across cities. POI counts provide the available activity signal for constructing a reproducible Place2Vec-style region representation.} \\
\midrule

Urban2Vec
& CityRep-compatible reproduction
& Mapillary street-view imagery plus Foursquare POI categories aggregated to H3 regions.
& 128-d H3 region table, resolution 8. Variant registered as street-view plus POI.
& Urban2Vec open-source code: \url{https://github.com/wangzhecheng/urban2vec_}; CityRep wrapper adapts inputs to Mapillary, Foursquare, and H3 regions. \\
\multicolumn{5}{p{0.96\textwidth}}{\textit{Implementation note.}
Urban2Vec is reproduced with the public street-view and POI inputs available in CityRep. The registered embeddings use the union of available modality support so that regions with one available modality can still be evaluated.} \\
\midrule

MuseCL
& CityRep-compatible variant
& Mapillary street-view branch, AlphaEarth-based remote branch, and Foursquare POI text branch.
& 128-d H3 region table, resolution 8. Region embeddings are looked up by H3 cell.
& Official MuseCL code: \url{https://github.com/XixianYong/MuseCL}; CityRep wrapper provides cross-city asset preparation and H3-region export. \\
\multicolumn{5}{p{0.96\textwidth}}{\textit{Implementation note.}
MuseCL is implemented as a CityRep-compatible multimodal region encoder using modalities that can be registered consistently across the benchmark. The mobility-related activity signal is represented with available POI-type information.} \\
\midrule

\rowcolor{gray!15}
\multicolumn{5}{l}{\textit{Coordinate and POI-supervised location encoders}} \\
\midrule
PE
& Deterministic location baseline
& Longitude--latitude coordinates at each CityRep task unit.
& 192-d entity/coordinate table. Fixed Sphere2Vec-\texttt{sphereC} position encoding with 64 frequencies; task samples are matched by \texttt{sample\_id} after evaluating at point coordinates or raster-cell centers.
& CityRep implementation following the Sphere2Vec-\texttt{sphereC} positional basis: \url{https://github.com/gengchenmai/sphere2vec}; no learned checkpoint. \\
\multicolumn{5}{p{0.96\textwidth}}{\textit{Implementation note.}
PE is a non-learned location-only baseline. It is included to measure how much downstream performance can be explained by deterministic coordinate encoding alone under the same spatial split and downstream evaluator.} \\
\midrule

CaLLiPer
& CityRep-compatible reproduction
& Foursquare POI coordinate--text pairs; POI descriptions include names and categories.
& 128-d coordinate encoder. Land-use uses point-table inference; raster tasks query the encoder at task pixel centers.
& Official CaLLiPer code: \url{https://github.com/xlwang233/CaLLiPer}; CityRep wrapper retrains/exports coordinate embeddings on benchmark POI inputs. \\
\multicolumn{5}{p{0.96\textwidth}}{\textit{Implementation note.}
CaLLiPer is trained with the common Foursquare input corpus used by CityRep so that the same city coverage and task alignment protocol can be applied to all benchmark cities.} \\
\midrule

Space2Vec
& CityRep-compatible reproduction
& Foursquare POI coordinates and second-level category labels.
& 128-d coordinate encoder. Land-use uses point-table inference; raster tasks query the encoder at task pixel centers.
& Official Space2Vec code: \url{https://github.com/gengchenmai/space2vec}; CityRep wrapper prepares POI classification data and task-grid exports. \\
\multicolumn{5}{p{0.96\textwidth}}{\textit{Implementation note.}
Space2Vec is trained under a uniform Foursquare-based POI classification setup for all cities. This provides a consistent coordinate-encoder baseline with the same task sampling and raster-center inference interface used by other coordinate models.} \\
\midrule

SatCLIP
& Official checkpoint
& Longitude--latitude coordinates only at CityRep task locations.
& 256-d coordinate encoder. Land-use and raster tasks are queried directly at sample coordinates or raster-cell centers.
& Official SatCLIP code: \url{https://github.com/microsoft/satclip}; official checkpoint \texttt{microsoft/SatCLIP-ViT16-L40}. \\
\multicolumn{5}{p{0.96\textwidth}}{\textit{Implementation note.}
SatCLIP is evaluated as an off-the-shelf coordinate checkpoint. It uses only location coordinates at CityRep task units and does not use CityRep POI, map, street-view, or task-label data.} \\

\end{xltabular}
\endgroup

\clearpage

\subsection{Aggregation for Sparse Entity Representations}
\label{app:h3-first-aggregation}

CityRep uses an H3-first aggregation strategy for sparse entity-based representations such as POI and map-entity embeddings. Instead of directly aggregating sparse entities to each downstream task unit, we first aggregate entities to H3 resolution-8 cells and then align the resulting cell embeddings to task units. This intermediate support reduces missing or unstable features when task units are small or when entities are unevenly distributed.

Table~\ref{tab:app_h3_first_alignment} compares this strategy with direct task-unit aggregation for CityFM and Place2Vec. The H3-first strategy improves most task metrics for both models, supporting our choice to use an intermediate spatial support before downstream alignment.

Table~\ref{tab:app_h3_first_coverage} further compares the average feature coverage of H3-first aggregation and direct task-unit aggregation on dense raster tasks. Coverage is defined as the fraction of task units that receive a valid non-missing representation after alignment. H3-first aggregation substantially improves coverage for both CityFM and Place2Vec, especially on LST and POP, explaining why the intermediate H3 support leads to more stable downstream evaluation.

\newcommand{\gain}[1]{\textcolor{green!50!black}{\scriptsize~(#1)}}
\newcommand{\loss}[1]{\textcolor{red!70!black}{\scriptsize~(#1)}}

\begin{table}[t]
\centering
\caption{
Average feature coverage of H3-first aggregation and direct task-unit aggregation for sparse entity-based representations.
Coverage is the fraction of downstream task units with a valid aligned representation.
Numbers in parentheses on H3-first rows report $\Delta=$ H3-first minus Task-unit direct.
Green indicates higher coverage, and red indicates lower coverage.
}
\label{tab:app_h3_first_coverage}
\scriptsize
\resizebox{\textwidth}{!}{
\begin{tabular}{llcccccccc}
\toprule
Model & Variant
& LUC & RDE & POP & AGE & GDP & NTL & PM$_{2.5}$ & LST \\
\midrule
CityFM
& H3-first
& 0.817 \gain{+0.488}
& 0.775 \gain{+0.479}
& 0.912 \gain{+0.554}
& 0.958 \gain{+0.545}
& 0.781 \loss{-0.003}
& 0.776 \gain{+0.134}
& 0.797 \loss{-0.039}
& 0.814 \gain{+0.505} \\
& Task-unit direct
& 0.329
& 0.296
& 0.358
& 0.413
& 0.784
& 0.642
& 0.836
& 0.309 \\
\midrule
Place2Vec
& H3-first
& 0.840 \gain{+0.482}
& 0.806 \gain{+0.474}
& 0.927 \gain{+0.513}
& 0.958 \gain{+0.480}
& 0.810 \loss{-0.003}
& 0.806 \gain{+0.118}
& 0.827 \loss{-0.029}
& 0.836 \gain{+0.474} \\
& Task-unit direct
& 0.358
& 0.332
& 0.414
& 0.478
& 0.813
& 0.688
& 0.856
& 0.362 \\
\bottomrule
\end{tabular}
}
\end{table}

\begin{table}[t]
\centering
\caption{
Effect of H3-first aggregation compared with direct task-unit aggregation for sparse entity-based representations.
Numbers in parentheses on H3-first rows report $\Delta=$ H3-first minus Task-unit direct.
Green indicates improvement according to the metric direction, and red indicates degradation.
}
\label{tab:app_h3_first_alignment}
\scriptsize
\resizebox{\textwidth}{!}{
\begin{tabular}{llcccccccc}
\toprule
Model & Variant
& LUC & RDE & POP & AGE & GDP & NTL & PM$_{2.5}$ & LST \\
&
& F1$\uparrow$
& $R^2 \uparrow$
& $R^2 \uparrow$
& KL$\downarrow$
& $R^2 \uparrow$
& $R^2 \uparrow$
& $R^2 \uparrow$
& $R^2 \uparrow$ \\
\midrule
CityFM
& H3-first
& 0.165 \gain{+0.041}
& 0.187 \gain{+0.057}
& 0.228 \gain{+0.093}
& 0.021 \gain{-0.003}
& 0.199 \gain{+0.001}
& 0.369 \gain{+0.011}
& 0.348 \loss{-0.038}
& 0.496 \gain{+0.385} \\
& Task-unit direct
& 0.125
& 0.130
& 0.135
& 0.024
& 0.198
& 0.358
& 0.386
& 0.111 \\
\midrule
Place2Vec
& H3-first
& 0.168 \gain{+0.034}
& 0.217 \gain{+0.037}
& 0.288 \gain{+0.135}
& 0.023 \gain{-0.003}
& 0.191 \gain{+0.006}
& 0.303 \gain{+0.041}
& 0.063 \loss{-0.022}
& 0.138 \gain{+0.159} \\
& Task-unit direct
& 0.134
& 0.180
& 0.154
& 0.026
& 0.185
& 0.262
& 0.085
& -0.022 \\
\bottomrule
\end{tabular}
}
\end{table}

\subsection{Evaluation Metrics}
\label{app:evaluation-metrics}
CityRep evaluates downstream prediction performance using task-appropriate metrics for regression, classification, and distribution prediction. This appendix provides the formal definitions of all reported metrics.

\paragraph{Regression metrics.}
For regression tasks, including road density, population, GDP, NTL, PM$_{2.5}$, and land-surface temperature, we report coefficient of determination ($R^2$), mean absolute error (MAE), and root mean squared error (RMSE).

The coefficient of determination is defined as
\begin{equation}
R^2
=
1-
\frac{
\sum_{i=1}^{n}(y_i-\hat{y}_i)^2
}{
\sum_{i=1}^{n}(y_i-\bar{y})^2
},
\end{equation}
where $y_i$ is the ground-truth target, $\hat{y}_i$ is the prediction, and $\bar{y}$ is the mean target value.

Mean absolute error is defined as
\begin{equation}
\mathrm{MAE}
=
\frac{1}{n}
\sum_{i=1}^{n}
|y_i-\hat{y}_i|.
\end{equation}

Root mean squared error is defined as
\begin{equation}
\mathrm{RMSE}
=
\sqrt{
\frac{1}{n}
\sum_{i=1}^{n}
(y_i-\hat{y}_i)^2
}.
\end{equation}

Among these metrics, $R^2$ is used as the primary metric because it measures explained variance and is less sensitive to target scale differences across tasks and cities.

\paragraph{Classification metrics.}
For land-use classification, we report macro F1, macro precision, and macro recall. Let $C$ denote the number of classes. For class $c$, precision and recall are defined as
\begin{equation}
P_c
=
\frac{\mathrm{TP}_c}
{\mathrm{TP}_c+\mathrm{FP}_c},
\qquad
R_c
=
\frac{\mathrm{TP}_c}
{\mathrm{TP}_c+\mathrm{FN}_c},
\end{equation}
where $\mathrm{TP}_c$, $\mathrm{FP}_c$, and $\mathrm{FN}_c$ denote true positives, false positives, and false negatives for class $c$.

The class-wise F1 score is
\begin{equation}
F1_c
=
\frac{2P_cR_c}{P_c+R_c}.
\end{equation}

Macro precision, macro recall, and macro F1 are computed by averaging across classes:
\begin{equation}
\mathrm{MacroPrecision}
=
\frac{1}{C}
\sum_{c=1}^{C} P_c,
\end{equation}

\begin{equation}
\mathrm{MacroRecall}
=
\frac{1}{C}
\sum_{c=1}^{C} R_c,
\end{equation}

\begin{equation}
\mathrm{MacroF1}
=
\frac{1}{C}
\sum_{c=1}^{C} F1_c.
\end{equation}

Macro F1 is used as the primary metric because the land-use labels are imbalanced across cities and categories, and macro averaging gives equal weight to each class.

\paragraph{Distribution-prediction metrics.}
For age-distribution prediction, the target for each task unit is a probability distribution over age groups. Let $\mathbf{p}=(p_1,\dots,p_K)$ denote the target distribution and $\mathbf{q}=(q_1,\dots,q_K)$ the predicted distribution.

We report KL divergence:
\begin{equation}
D_{\mathrm{KL}}(\mathbf{p}\|\mathbf{q})
=
\sum_{k=1}^{K}
p_k
\log
\frac{p_k}{q_k+\epsilon},
\end{equation}
where $\epsilon$ is a small numerical constant added for stability.

We also report Chebyshev distance:
\begin{equation}
D_{\mathrm{Chebyshev}}
=
\max_k |p_k-q_k|,
\end{equation}
and L1 distance:
\begin{equation}
D_{L1}
=
\sum_{k=1}^{K}
|p_k-q_k|.
\end{equation}

KL divergence is used as the primary metric because it measures distributional mismatch while accounting for relative probability mass differences across age groups.

\paragraph{Metric direction.}
Higher values indicate better performance for $R^2$, macro precision, macro recall, and macro F1. Lower values indicate better performance for MAE, RMSE, KL divergence, Chebyshev distance, and L1 distance.

\subsection{Training protocol.}
\label{app:training-protocol}
All downstream predictors are trained with the same hyperparameters across models and tasks: a multilayer perceptron with hidden dimension 1024, batch size 512, learning rate $10^{-3}$, a maximum of 100 epochs, and validation early stopping with patience 10.
Regression targets are standardized within each city--task dataset during training.
These choices keep the downstream model sufficiently expressive to use the embeddings, while limiting task-specific tuning.

All experiments were conducted using PyTorch on a workstation equipped with two NVIDIA RTX PRO 6000 Blackwell GPUs, each with 96GB memory, and an AMD Ryzen Threadripper PRO 9975WX CPU with 32 cores and 64 threads.
Spatial preprocessing and alignment were performed on the CPU using standard geospatial raster and vector processing libraries, while downstream predictors were trained on GPUs.
Most downstream evaluation runs used a single GPU and required less than several hours per city--task pair, depending on task resolution and embedding size.

\paragraph{Aggregation across seeds and cities.}
Let $s_{m,t,c,k}$ denote the primary test score for model $m$, task $t$, city $c$, and seed $k$.
In our benchmark instantiation, we run five split seeds, $\{42,24,7,0,100\}$, and first average the primary metric over seeds to obtain a city-level score:
\begin{equation}
s_{m,t,c}=\frac{1}{5}\sum_{k=1}^{5}s_{m,t,c,k}.
\end{equation}
We then report, for each model and task, the mean of these city-level scores across cities as
\begin{equation}
\mathrm{Avg}_{m,t}
=
\frac{1}{|\mathcal{C}_t|}
\sum_{c\in\mathcal{C}_t} s_{m,t,c},
\end{equation}
together with the cross-city standard deviation
\begin{equation}
\mathrm{C\mbox{-}Std.}_{m,t}
=
\sqrt{
\frac{1}{|\mathcal{C}_t|-1}
\sum_{c\in\mathcal{C}_t}
\left(s_{m,t,c}-\mathrm{Avg}_{m,t}\right)^2
}.
\end{equation}
Here $\mathcal{C}_t$ denotes the set of cities included for task $t$.
The average score measures overall task utility, while C-Std. measures how much performance varies across urban contexts.
Since different tasks use different primary metrics and metric scales, CityRep uses raw task metrics for the main result columns and uses rank-based summaries only as diagnostic views for comparing models across tasks.

\paragraph{Overall rank.}
The overall rank in Table~\ref{tab:main_results_avg_cstd} is computed from city-level rankings rather than from normalized metric values.
For each task--city pair $(t,c)$, we rank all models according to the seed-averaged city-level score $s_{m,t,c}$, with rank 1 assigned to the best model.
Higher values are better for $R^2$ and macro F1, while lower values are better for KL divergence in the age-distribution task.
Ties receive the best shared rank.
This gives a city-level rank $r_{m,t,c}$.

We then average city-level ranks within each task:
\begin{equation}
R_{m,t}=\frac{1}{|\mathcal{C}_t|}\sum_{c\in\mathcal{C}_t} r_{m,t,c},
\end{equation}
where $\mathcal{C}_t$ is the set of evaluated cities for task $t$.
For all tasks except age distribution, $\mathcal{C}_t$ contains the eight benchmark cities.
For age distribution, $\mathcal{C}_t$ contains only London, New York, Singapore, and Sydney because the remaining four cities are excluded from the main AGE aggregation due to source-data quality concerns.

Finally, the overall rank is the equal-task average:
\begin{equation}
R_m=\frac{1}{|\mathcal{T}|}\sum_{t\in\mathcal{T}} R_{m,t}.
\end{equation}
Lower values indicate better overall performance.
This aggregation avoids comparing raw metric scales across tasks and prevents tasks with more evaluated cities or denser samples from dominating the overall ranking.

\section{Comprehensive Results}
\label{app:results}

\begin{figure}[t]
    \centering
    \includegraphics[width=0.98\linewidth]{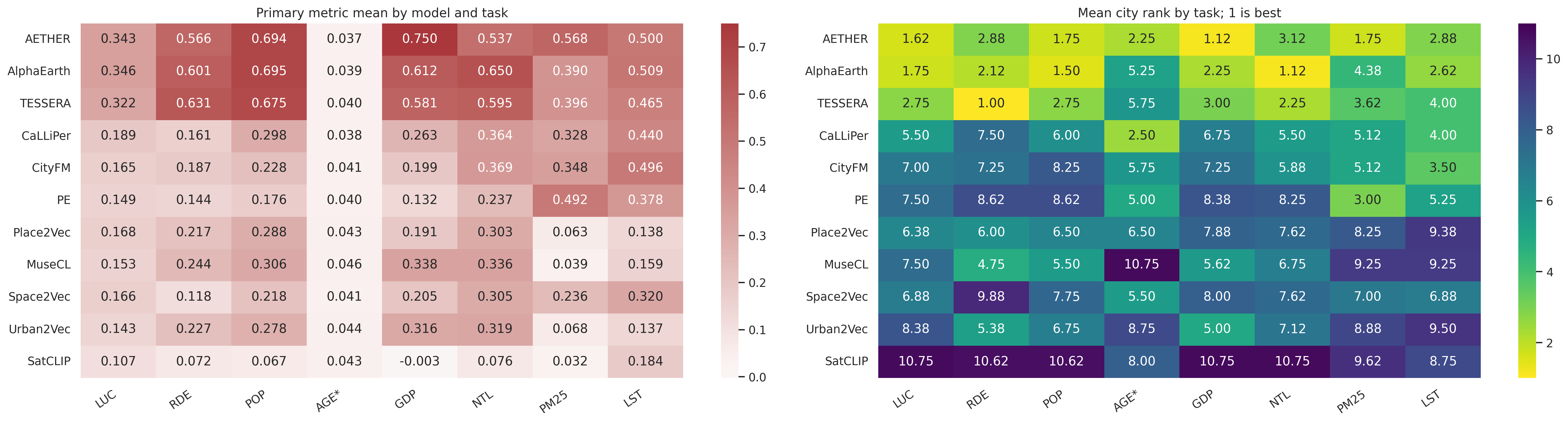}
    \caption{Spatial-split primary metric values and rank diagnostics for all evaluated models.}
    \label{fig:app-main-primary-rank-heatmaps}
\end{figure}

\begin{figure}[t]
    \centering
    \includegraphics[width=0.98\linewidth]{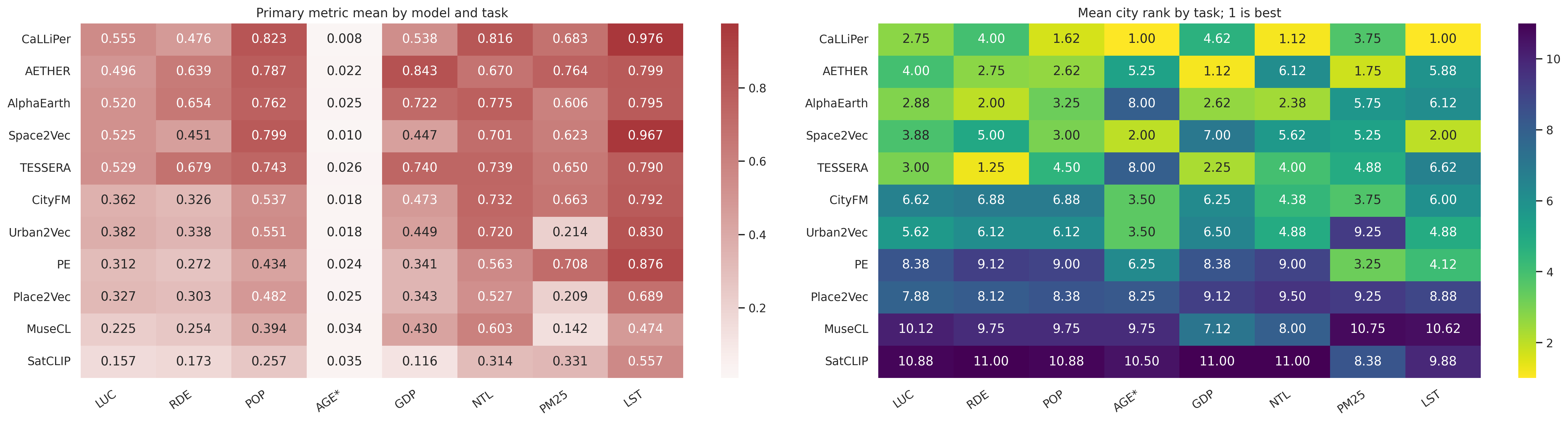}
    \caption{Random-split primary metric values and rank diagnostics for all evaluated models.}
    \label{fig:app-random-primary-rank-heatmaps}
\end{figure}

Figure~\ref{fig:app-main-primary-rank-heatmaps} provides spatial-split diagnostics corresponding to the main benchmark results, and Figure~\ref{fig:app-random-primary-rank-heatmaps} provides the same view under random splitting. These figures are diagnostic rather than replacements for the main table: raw primary metrics remain task-specific, while rank views are used to compare model ordering across heterogeneous metrics. We omit separate mean-city-rank figures here because that heatmap is already included in the right panel of each primary-metric diagnostic figure.

\subsection{Results Across Tasks and Cities}
\label{app:city-task-results}

\begin{figure}[t]
    \centering
    \includegraphics[width=0.98\linewidth]{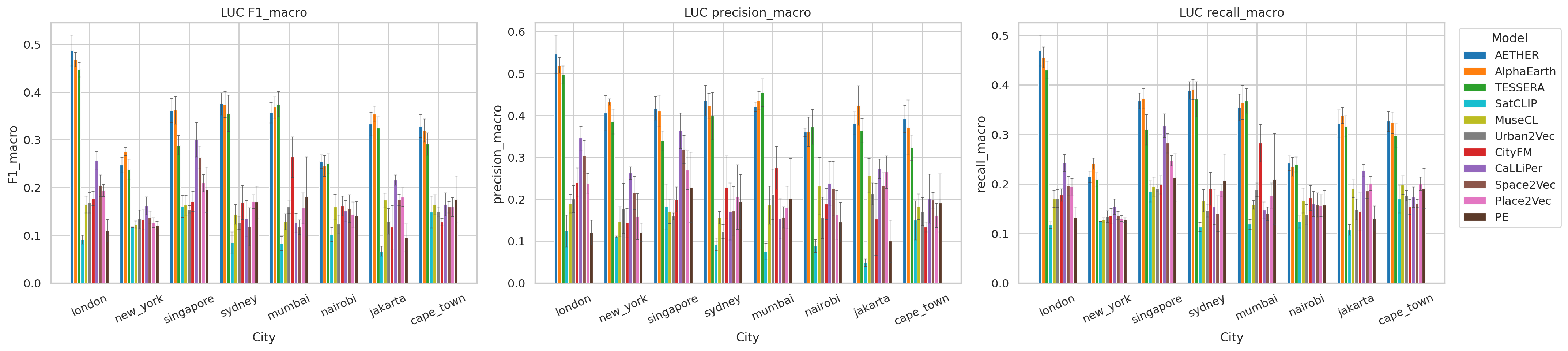}
    \caption{Land-use classification: full metric bars under spatial block splits.}
    \label{fig:app-luc-full-metrics-spatial}
\end{figure}

\begin{figure}[t]
    \centering
    \includegraphics[width=0.98\linewidth]{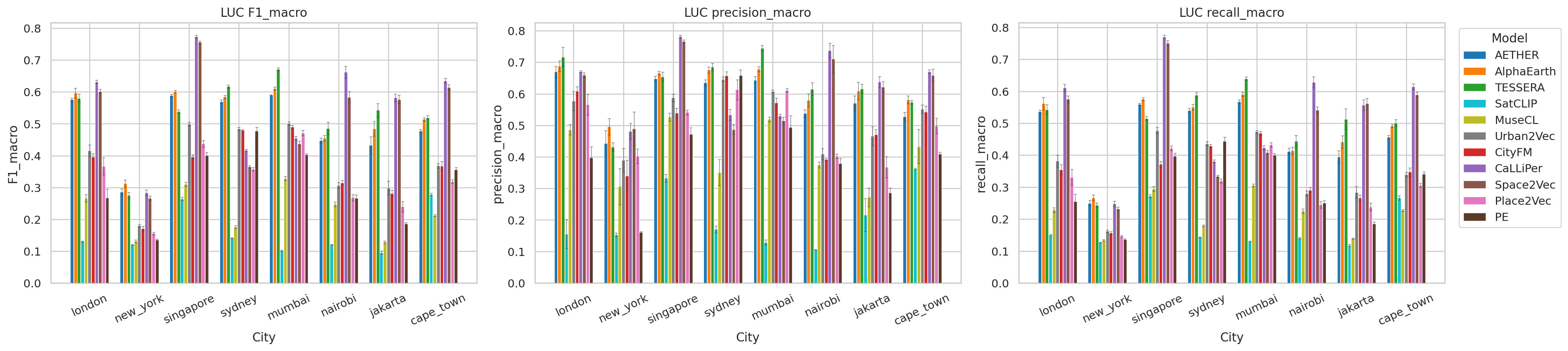}
    \caption{Land-use classification: full metric bars under random splits.}
    \label{fig:app-luc-full-metrics-random}
\end{figure}

\begin{figure}[t]
    \centering
    \includegraphics[width=0.98\linewidth]{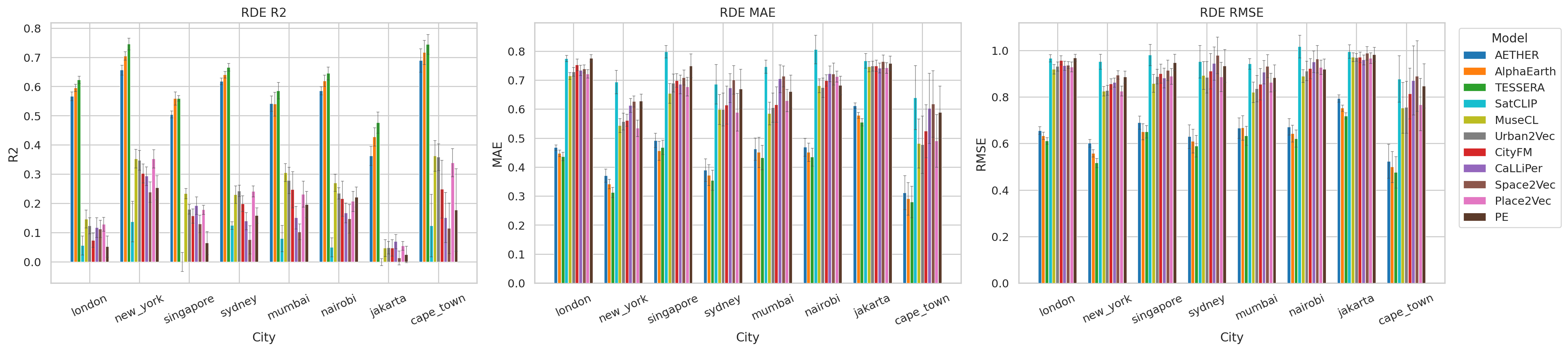}
    \caption{Road-density regression: full metric bars under spatial block splits.}
    \label{fig:app-rde-full-metrics-spatial}
\end{figure}

\begin{figure}[t]
    \centering
    \includegraphics[width=0.98\linewidth]{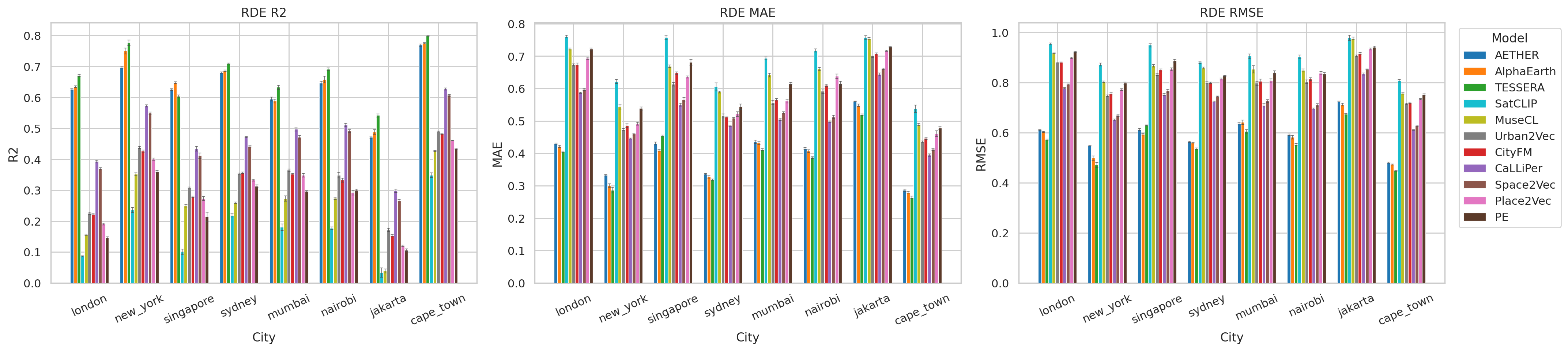}
    \caption{Road-density regression: full metric bars under random splits.}
    \label{fig:app-rde-full-metrics-random}
\end{figure}

\begin{figure}[t]
    \centering
    \includegraphics[width=0.98\linewidth]{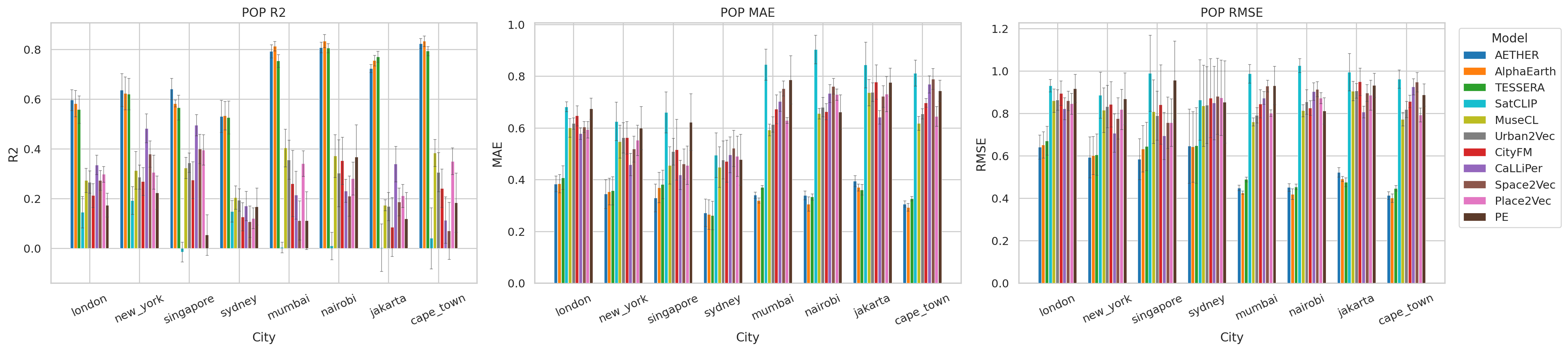}
    \caption{Population regression: full metric bars under spatial block splits.}
    \label{fig:app-pop-full-metrics-spatial}
\end{figure}

\begin{figure}[t]
    \centering
    \includegraphics[width=0.98\linewidth]{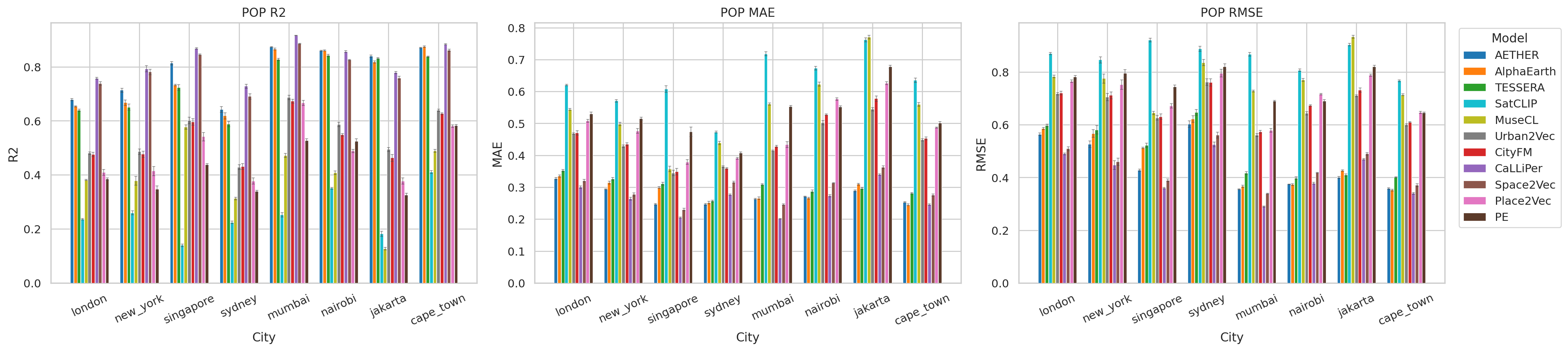}
    \caption{Population regression: full metric bars under random splits.}
    \label{fig:app-pop-full-metrics-random}
\end{figure}

\begin{figure}[t]
    \centering
    \includegraphics[width=0.98\linewidth]{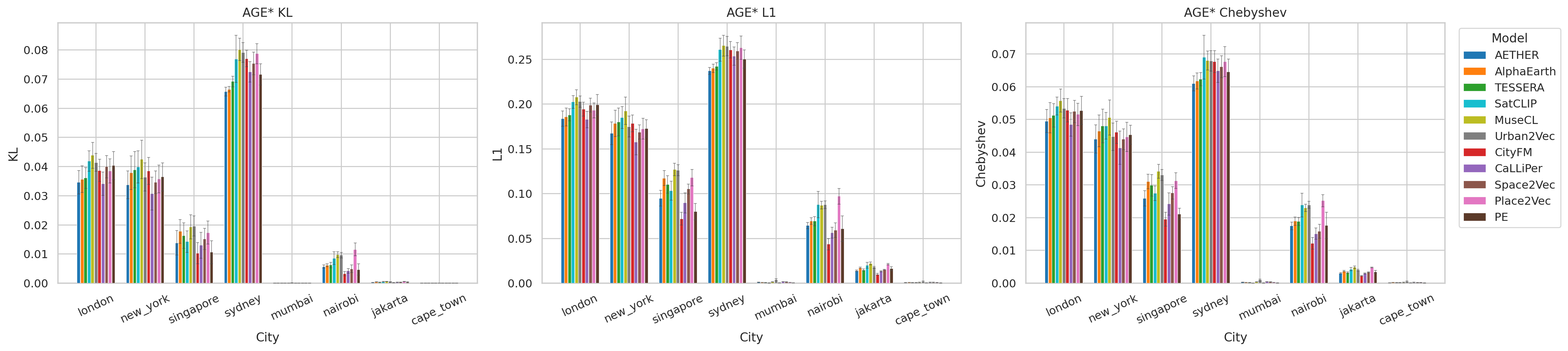}
    \caption{Age-distribution prediction: full metric bars under spatial block splits.}
    \label{fig:app-age-full-metrics-spatial}
\end{figure}

\begin{figure}[t]
    \centering
    \includegraphics[width=0.98\linewidth]{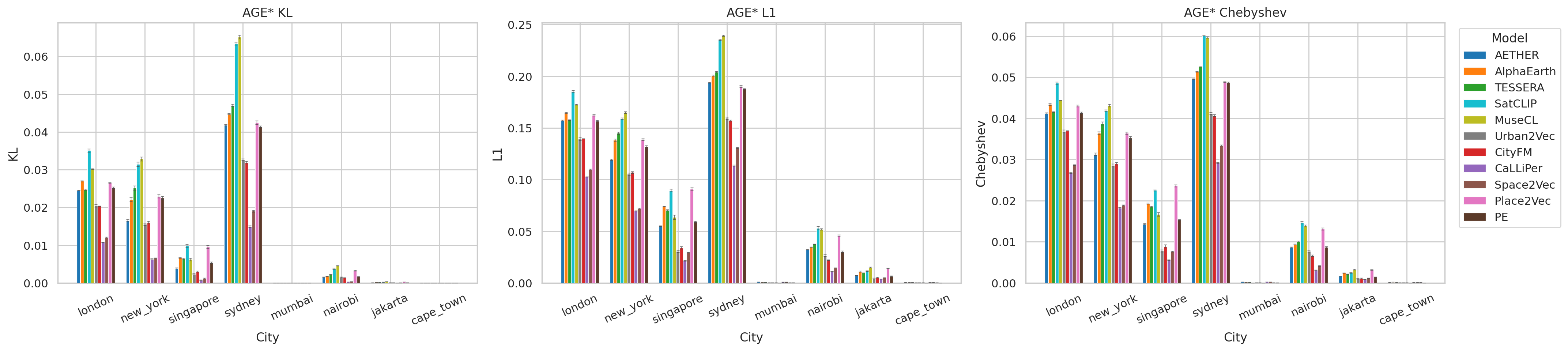}
    \caption{Age-distribution prediction: full metric bars under random splits.}
    \label{fig:app-age-full-metrics-random}
\end{figure}

\begin{figure}[t]
    \centering
    \includegraphics[width=0.98\linewidth]{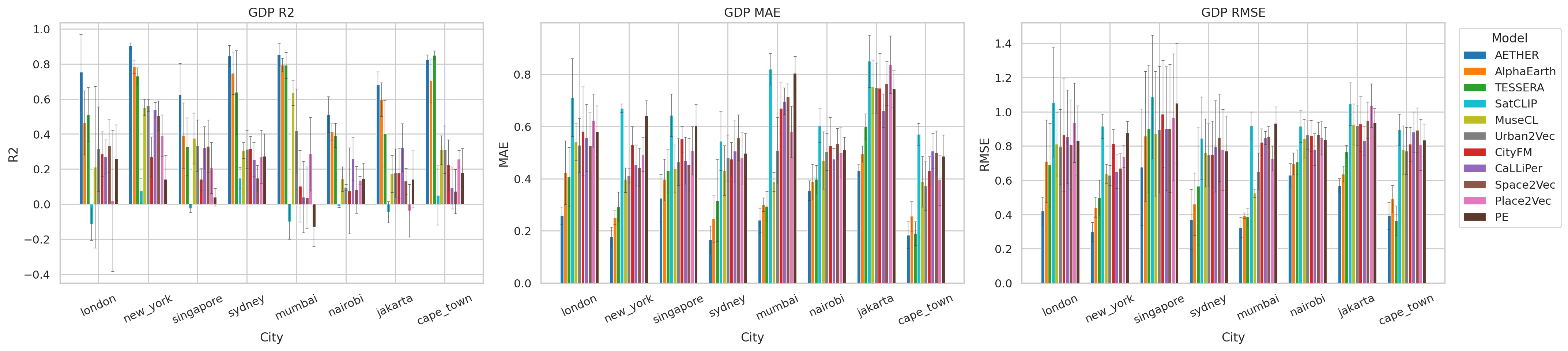}
    \caption{GDP regression: full metric bars under spatial block splits.}
    \label{fig:app-gdp-full-metrics-spatial}
\end{figure}

\begin{figure}[t]
    \centering
    \includegraphics[width=0.98\linewidth]{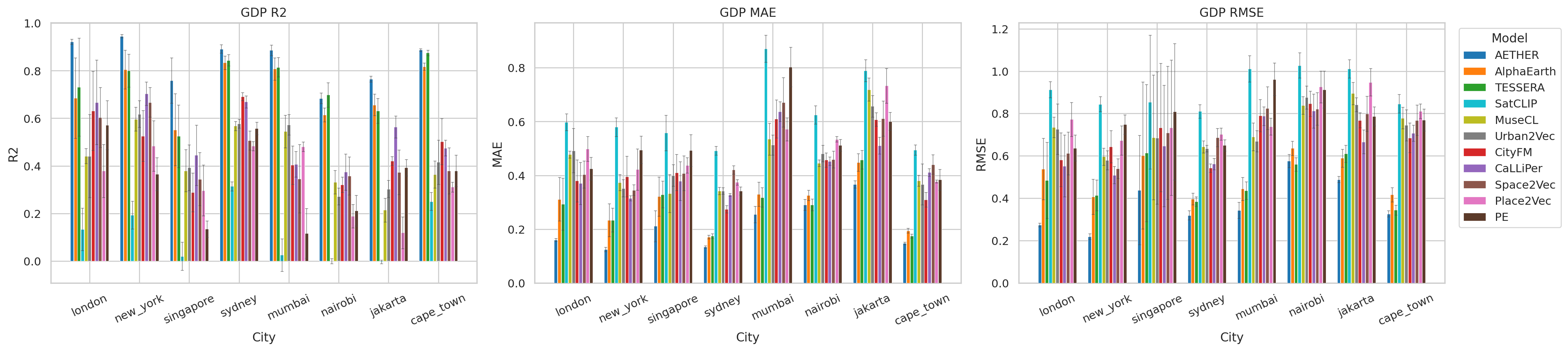}
    \caption{GDP regression: full metric bars under random splits.}
    \label{fig:app-gdp-full-metrics-random}
\end{figure}

\begin{figure}[t]
    \centering
    \includegraphics[width=0.98\linewidth]{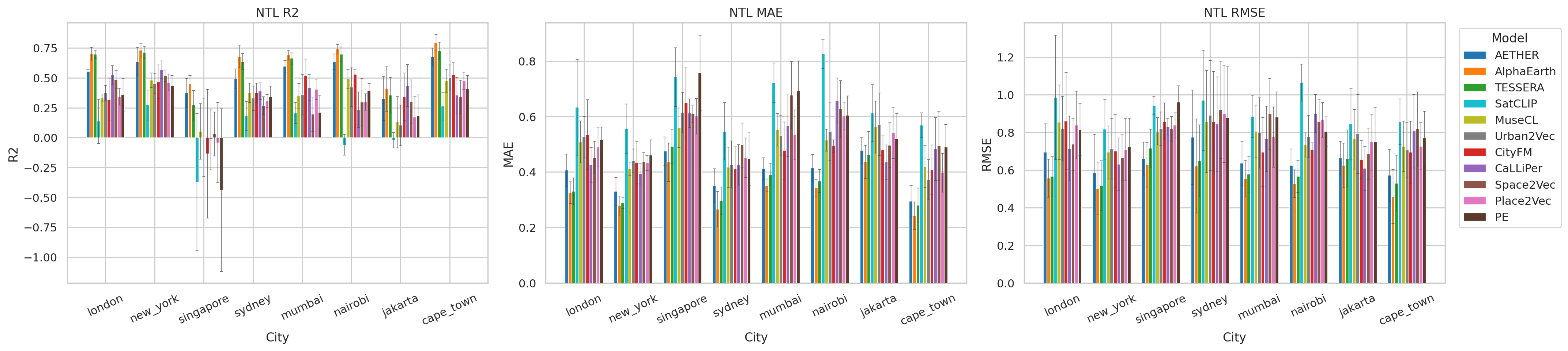}
    \caption{NTL regression: full metric bars under spatial block splits.}
    \label{fig:app-ntl-full-metrics-spatial}
\end{figure}

\begin{figure}[t]
    \centering
    \includegraphics[width=0.98\linewidth]{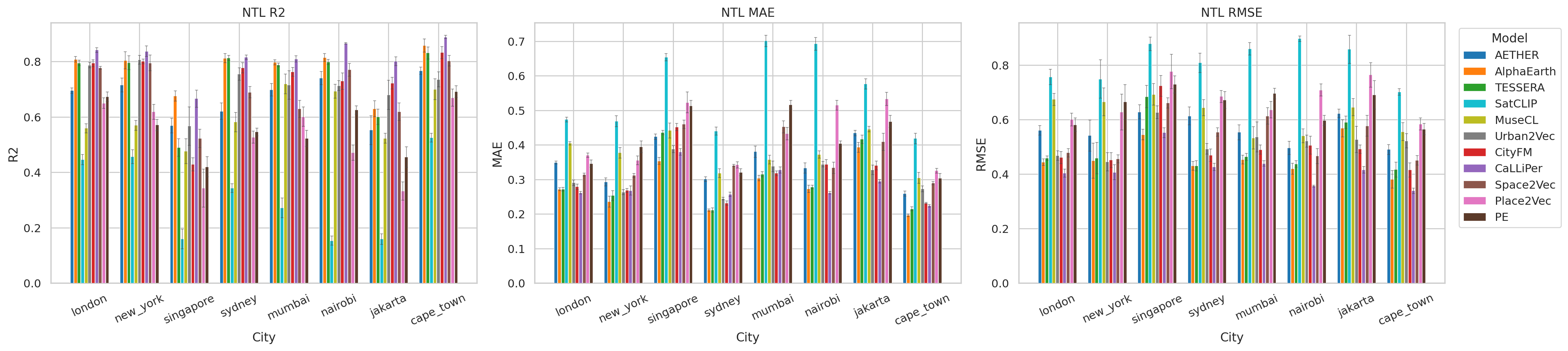}
    \caption{NTL regression: full metric bars under random splits.}
    \label{fig:app-ntl-full-metrics-random}
\end{figure}

\begin{figure}[t]
    \centering
    \includegraphics[width=0.98\linewidth]{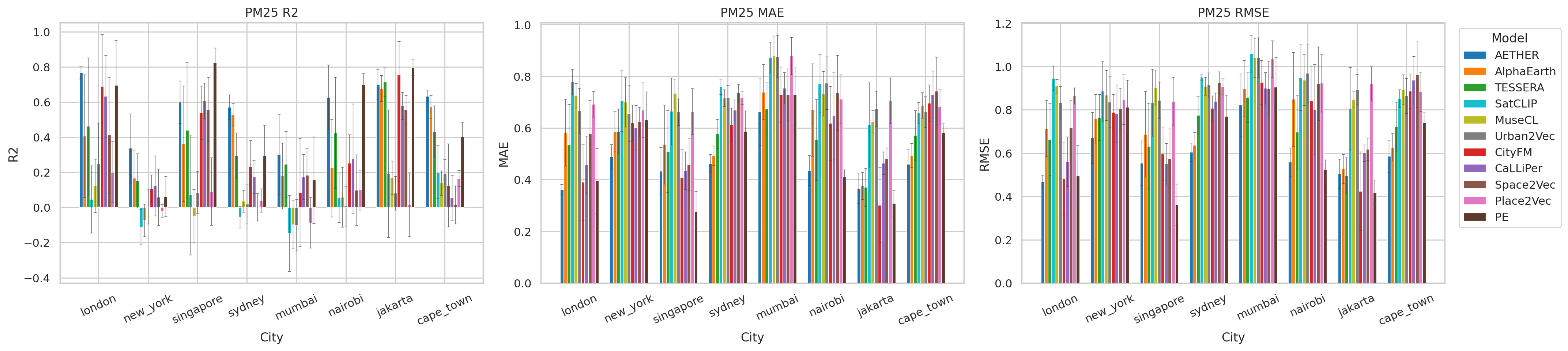}
    \caption{PM$_{2.5}$ regression: full metric bars under spatial block splits.}
    \label{fig:app-pm25-full-metrics-spatial}
\end{figure}

\begin{figure}[t]
    \centering
    \includegraphics[width=0.98\linewidth]{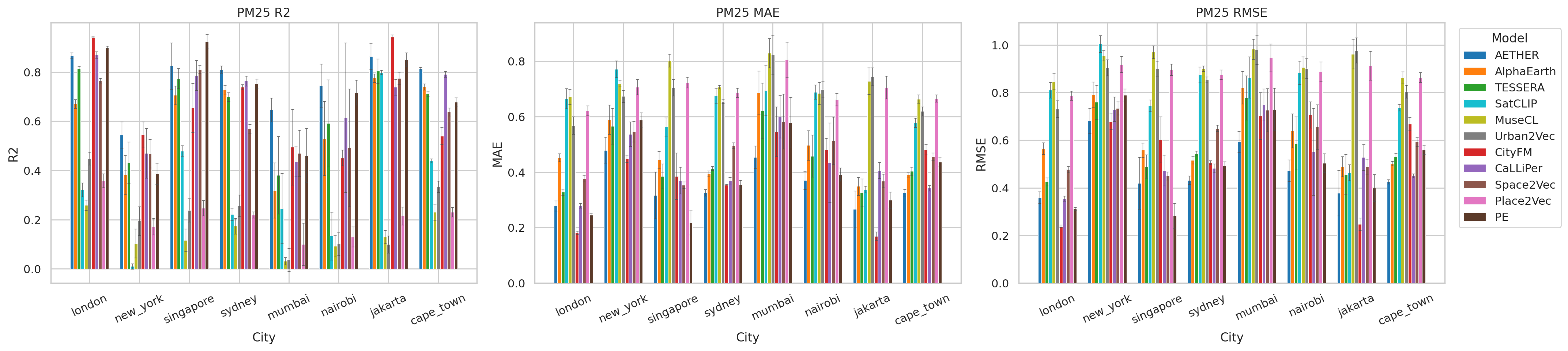}
    \caption{PM$_{2.5}$ regression: full metric bars under random splits.}
    \label{fig:app-pm25-full-metrics-random}
\end{figure}

\begin{figure}[t]
    \centering
    \includegraphics[width=0.98\linewidth]{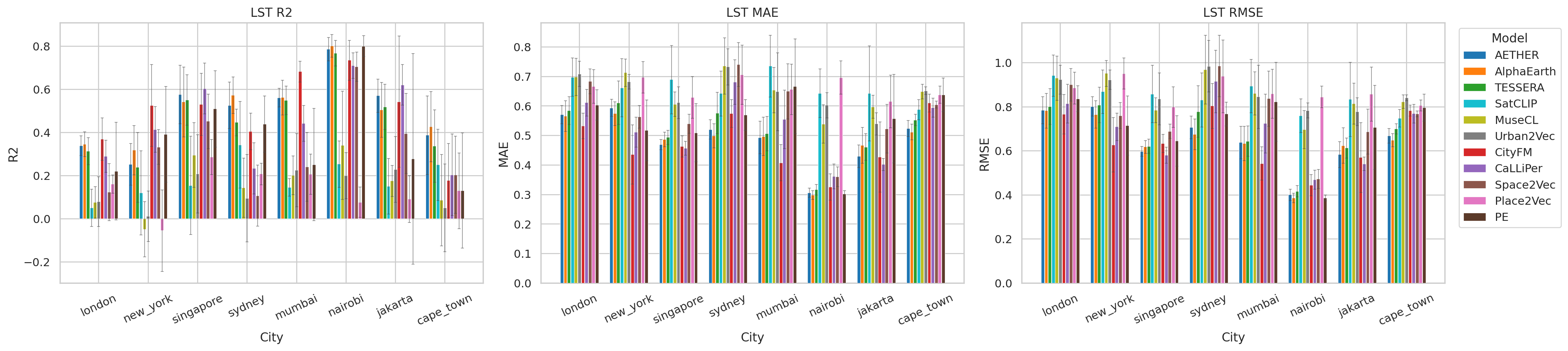}
    \caption{Land-surface-temperature regression: full metric bars under spatial block splits.}
    \label{fig:app-lst-full-metrics-spatial}
\end{figure}

\begin{figure}[t]
    \centering
    \includegraphics[width=0.98\linewidth]{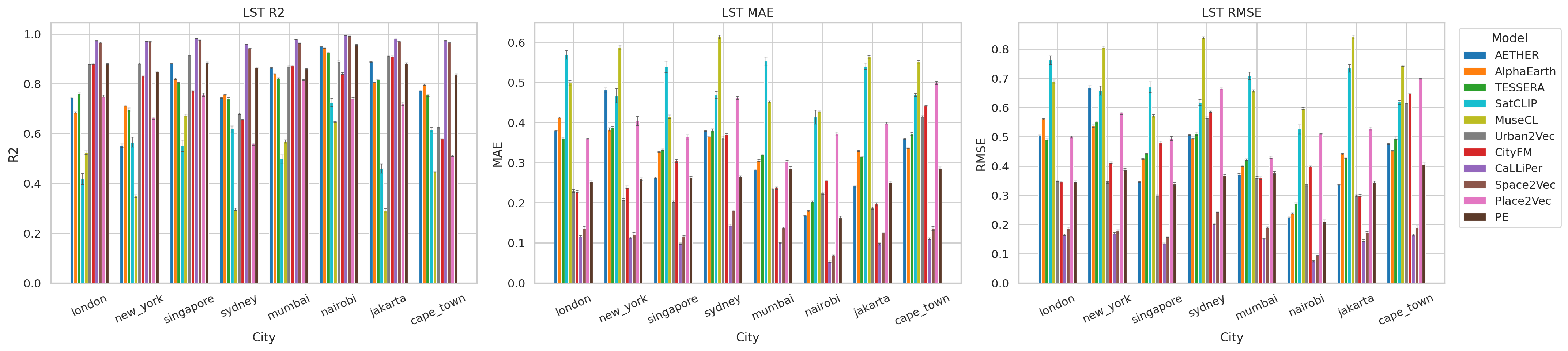}
    \caption{Land-surface-temperature regression: full metric bars under random splits.}
    \label{fig:app-lst-full-metrics-random}
\end{figure}

This section provides the full task-level metric plots for both spatial and random splits. Figures~\ref{fig:app-luc-full-metrics-spatial} and~\ref{fig:app-luc-full-metrics-random} show the land-use results; Figures~\ref{fig:app-rde-full-metrics-spatial} and~\ref{fig:app-rde-full-metrics-random} show the road-density results; Figures~\ref{fig:app-pop-full-metrics-spatial} and~\ref{fig:app-pop-full-metrics-random} show the population results; Figures~\ref{fig:app-age-full-metrics-spatial} and~\ref{fig:app-age-full-metrics-random} show the age-distribution results; Figures~\ref{fig:app-gdp-full-metrics-spatial} and~\ref{fig:app-gdp-full-metrics-random} show the GDP results; Figures~\ref{fig:app-ntl-full-metrics-spatial} and~\ref{fig:app-ntl-full-metrics-random} show the NTL results; Figures~\ref{fig:app-pm25-full-metrics-spatial} and~\ref{fig:app-pm25-full-metrics-random} show the PM$_{2.5}$ results; and Figures~\ref{fig:app-lst-full-metrics-spatial} and~\ref{fig:app-lst-full-metrics-random} show the land-surface-temperature results. Unlike the main table, which reports one primary metric per task, these figures include secondary metrics as well: macro precision and macro recall for land use, MAE and RMSE for regression tasks, and L1 distance and Chebyshev distance for age-distribution prediction. Bars report the mean across five seeds, and error bars report the standard deviation across those seeds. These views are useful for checking whether a model's ranking is driven by one metric alone or whether it is stable across multiple error summaries.

The paired layout also shows how the split protocol changes both apparent performance and uncertainty. Spatial splits generally yield lower apparent performance on spatially smooth targets, while random splits can make local interpolation easier. In addition, spatial splits often produce larger error bars than random splits, especially for coarse raster tasks such as GDP, NTL, and PM$_{2.5}$. This does not necessarily indicate only optimization instability. Under the spatial protocol, different seeds hold out different geographic blocks, so the standard deviation combines model stochasticity with variation in the difficulty and target variance of the held-out regions. This effect is particularly visible for $R^2$, whose denominator depends on the variance of the test targets; if a held-out block has a narrow target range, a moderate prediction error can lead to a very low or negative $R^2$. MAE and RMSE are therefore reported alongside $R^2$ to help distinguish unstable explained-variance estimates from changes in absolute prediction error. The difference is not uniform across tasks or models, which is why we report full task-level plots rather than only a single aggregate rank. Spatial and random figures are shown separately to keep labels and city-level bars readable.

\clearpage

\subsection{Results with Linear Downstream Task Head}
\label{app:linear_probe}
\paragraph{Linear-probe evaluation.}
To examine whether the main benchmark conclusions depend on the capacity of the downstream MLP evaluator, we additionally evaluate all representations with a linear probe under the same spatial block splits, city--task datasets, metrics, and aggregation protocol. Table~\ref{tab:linear_probe_results_avg_cstd} reports the resulting average performance and cross-city standard deviation. Overall, the ranking is broadly consistent with the main benchmark: AETHER obtains the best mean city rank, followed by AlphaEarth and TESSERA. The three raster or raster-enhanced representations remain the strongest overall, suggesting that their advantage is not solely due to the nonlinear downstream head. At the task level, AETHER performs best on LUC, POP, AGE, GDP, and PM$_{2.5}$, AlphaEarth leads on NTL and LST, and TESSERA achieves the best result on RDE. As expected, the linear probe generally yields lower absolute scores than the MLP evaluator, especially on tasks that require more nonlinear decision boundaries or heterogeneous spatial signals. These results indicate that CityRep rankings are reasonably stable under a lower-capacity downstream evaluator, while also showing that some tasks benefit from nonlinear probing.

{\renewcommand{\arraystretch}{1.22}
\begin{table*}[t]
\centering
\caption{
\textbf{Linear-probe benchmark results on CityRep.} Same layout as the main benchmark table, but using a linear downstream probe. For each task, \textit{Avg.} reports the mean primary metric across 8 cities and 5 random seeds under the spatial block split. \textit{C Std.} reports the cross-city standard deviation of city-level performance. Type indicates the main data sources used by each representation: \textit{L} = location, \textit{P} = POI, \textit{R} = remote sensing, \textit{S} = street-view imagery, and \textit{O} = other urban/map data. \textit{Mean City Rank} is computed from unrounded city-level rankings across all tasks and cities (lower is better). Best, second-best, and third-best results in the Avg. columns are in \textbf{bold}, \underline{underlined}, and \uwave{wavy-underlined}, respectively. *For AGE, lower KL divergence indicates better performance, and results are reported only for the four cities with the most reliable age--sex source coverage.
}
\label{tab:linear_probe_results_avg_cstd}
\vspace{0.4em}
\scriptsize
\setlength{\tabcolsep}{1.8pt}
\resizebox{\textwidth}{!}{
\begin{tabular}{ll*{17}{c}}
\toprule
& & \multicolumn{4}{c}{$\spadesuit$~Morphology}
& \multicolumn{4}{c}{$\heartsuit$~Demographics}
& \multicolumn{4}{c}{$\diamondsuit$~Economy}
& \multicolumn{4}{c}{$\clubsuit$~Environment}
& \multicolumn{1}{c}{Overall} \\
\cmidrule(lr){3-6} \cmidrule(lr){7-10} \cmidrule(lr){11-14} \cmidrule(lr){15-18} \cmidrule(lr){19-19}
Model & Type
& \multicolumn{2}{c}{LUC}
& \multicolumn{2}{c}{RDE}
& \multicolumn{2}{c}{POP}
& \multicolumn{2}{c}{AGE$^{*}$}
& \multicolumn{2}{c}{GDP}
& \multicolumn{2}{c}{NTL}
& \multicolumn{2}{c}{PM$_{2.5}$}
& \multicolumn{2}{c}{LST}
& Rank \\
\cmidrule(lr){3-4} \cmidrule(lr){5-6}
\cmidrule(lr){7-8} \cmidrule(lr){9-10}
\cmidrule(lr){11-12} \cmidrule(lr){13-14}
\cmidrule(lr){15-16} \cmidrule(lr){17-18}
\cmidrule(lr){19-19}
&
& \shortstack{Avg.\\(F1 $\uparrow$)} & \shortstack{\textit{C}\\\textit{Std.}}
& \shortstack{Avg.\\($R^2 \uparrow$)} & \shortstack{\textit{C}\\\textit{Std.}}
& \shortstack{Avg.\\($R^2 \uparrow$)} & \shortstack{\textit{C}\\\textit{Std.}}
& \shortstack{Avg.\\(KL $\downarrow$)} & \shortstack{\textit{C}\\\textit{Std.}}
& \shortstack{Avg.\\($R^2 \uparrow$)} & \shortstack{\textit{C}\\\textit{Std.}}
& \shortstack{Avg.\\($R^2 \uparrow$)} & \shortstack{\textit{C}\\\textit{Std.}}
& \shortstack{Avg.\\($R^2 \uparrow$)} & \shortstack{\textit{C}\\\textit{Std.}}
& \shortstack{Avg.\\($R^2 \uparrow$)} & \shortstack{\textit{C}\\\textit{Std.}}
& \shortstack{Mean\\City Rank} \\
\midrule
PE & \textit{L}
& 0.125 & \textit{0.027}
& 0.111 & \textit{0.050}
& 0.130 & \textit{0.087}
& \uwave{0.042} & \textit{0.023}
& 0.183 & \textit{0.128}
& 0.220 & \textit{0.204}
& \underline{0.536} & \textit{0.243}
& 0.152 & \textit{0.124}
& 6.484 \\

Place2Vec & \textit{P}
& 0.137 & \textit{0.026}
& 0.159 & \textit{0.097}
& 0.217 & \textit{0.062}
& 0.046 & \textit{0.025}
& 0.200 & \textit{0.070}
& 0.266 & \textit{0.131}
& 0.084 & \textit{0.069}
& 0.077 & \textit{0.060}
& 7.422 \\

Space2Vec & \textit{LP}
& 0.128 & \textit{0.030}
& 0.107 & \textit{0.049}
& 0.180 & \textit{0.083}
& 0.048 & \textit{0.027}
& 0.202 & \textit{0.161}
& 0.282 & \textit{0.154}
& 0.271 & \textit{0.251}
& 0.135 & \textit{0.154}
& 7.578 \\

CaLLiPer & \textit{LP}
& 0.150 & \textit{0.048}
& 0.137 & \textit{0.055}
& 0.250 & \textit{0.110}
& 0.045 & \textit{0.027}
& 0.243 & \textit{0.156}
& 0.315 & \textit{0.134}
& 0.335 & \textit{0.266}
& 0.245 & \textit{0.138}
& 5.734 \\

CityFM & \textit{PO}
& 0.119 & \textit{0.015}
& 0.139 & \textit{0.079}
& 0.141 & \textit{0.084}
& 0.053 & \textit{0.032}
& 0.229 & \textit{0.034}
& 0.284 & \textit{0.173}
& 0.348 & \textit{0.304}
& 0.134 & \textit{0.183}
& 7.516 \\

Urban2Vec & \textit{PRS}
& 0.119 & \textit{0.014}
& 0.192 & \textit{0.090}
& 0.232 & \textit{0.085}
& 0.045 & \textit{0.022}
& 0.268 & \textit{0.147}
& 0.226 & \textit{0.149}
& 0.061 & \textit{0.105}
& 0.126 & \textit{0.099}
& 7.281 \\

MuseCL & \textit{PRS}
& 0.131 & \textit{0.016}
& 0.206 & \textit{0.094}
& 0.267 & \textit{0.091}
& 0.046 & \textit{0.022}
& 0.321 & \textit{0.119}
& 0.347 & \textit{0.113}
& 0.054 & \textit{0.065}
& 0.118 & \textit{0.100}
& 6.500 \\

SatCLIP & \textit{LR}
& 0.092 & \textit{0.022}
& 0.054 & \textit{0.039}
& 0.022 & \textit{0.050}
& 0.048 & \textit{0.021}
& -0.015 & \textit{0.096}
& 0.046 & \textit{0.148}
& 0.033 & \textit{0.087}
& 0.073 & \textit{0.139}
& 10.078 \\

TESSERA & \textit{R}
& \underline{0.258} & \textit{0.067}
& \textbf{0.538} & \textit{0.086}
& \uwave{0.592} & \textit{0.140}
& 0.043 & \textit{0.020}
& \uwave{0.596} & \textit{0.134}
& \uwave{0.595} & \textit{0.143}
& \uwave{0.445} & \textit{0.159}
& 0.373 & \textit{0.161}
& \uwave{2.688} \\

AlphaEarth & \textit{R}
& \uwave{0.254} & \textit{0.076}
& \underline{0.514} & \textit{0.087}
& \underline{0.615} & \textit{0.158}
& \underline{0.042} & \textit{0.019}
& \underline{0.598} & \textit{0.127}
& \textbf{0.632} & \textit{0.105}
& 0.436 & \textit{0.149}
& \textbf{0.442} & \textit{0.139}
& \underline{2.656} \\

AETHER & \textit{PR}
& \textbf{0.259} & \textit{0.079}
& \uwave{0.494} & \textit{0.104}
& \textbf{0.637} & \textit{0.108}
& \textbf{0.040} & \textit{0.020}
& \textbf{0.741} & \textit{0.101}
& \underline{0.488} & \textit{0.131}
& \textbf{0.562} & \textit{0.171}
& \underline{0.433} & \textit{0.136}
& \textbf{1.969} \\

\midrule
Mean over models & --
& 0.161 & \textit{0.038}
& 0.241 & \textit{0.075}
& 0.298 & \textit{0.096}
& 0.045 & \textit{0.023}
& 0.324 & \textit{0.116}
& 0.354 & \textit{0.144}
& 0.288 & \textit{0.170}
& 0.210 & \textit{0.130}
&  -- \\
\bottomrule
\end{tabular}
}
\end{table*}
}

\subsection{Results with 20x20 Spatial Split}
\label{app:split20_results}

\begin{figure}[t]
    \centering
    \includegraphics[width=\textwidth]{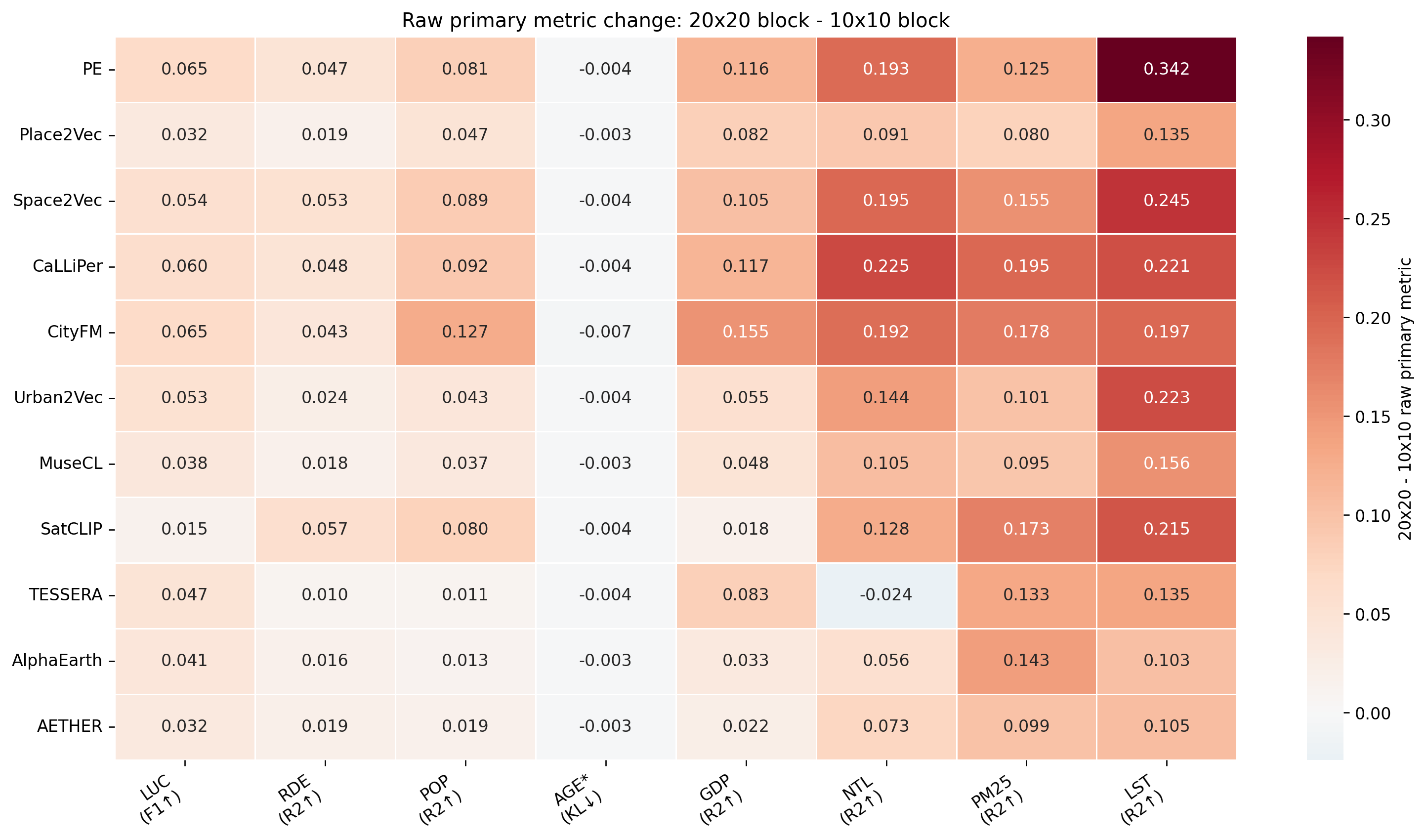}
    \caption{
    Raw primary metric change from the main 10$\times$10 spatial split to the 20$\times$20 spatial split. Values are computed as 20$\times$20 block result minus 10$\times$10 block result. Positive values indicate higher raw metric values; for AGE, the metric is KL divergence, so lower values are better.
    }
    \label{fig:split20_delta}
\end{figure}

To examine whether the benchmark conclusions are sensitive to the spatial block granularity, we additionally evaluate all baselines using a finer 20$\times$20 spatial split while keeping the same downstream models, task files, evaluation metrics, and five random seeds. In Figure~\ref{fig:split20_delta}, the 20$\times$20 split produces higher raw scores than the main 10$\times$10 split for most regression and classification tasks, especially for LST, NTL, and PM$_{2.5}$. This is expected because smaller spatial blocks reduce the distance between training and test regions and therefore make the spatial generalization setting less strict. 

The results are in Table~\ref{tab:main_results_split20}. Importantly, the relative ordering of model families remains broadly consistent. Remote-sensing raster representations remain the strongest group overall, while coordinate and region models improve more noticeably on several tasks under the finer split. These results support the main benchmark choice of the 10$\times$10 split as the stricter protocol, and show that the main findings are not an artifact of a single spatial partitioning resolution.

{\renewcommand{\arraystretch}{1.22}
\begin{table*}[!tb]
\centering
\caption{
\textbf{Benchmark results under the 20$\times$20 spatial block split.}
For each task, \textit{Avg.} reports the mean primary metric across cities and 5 random seeds. \textit{C Std.} reports the cross-city standard deviation of city-level performance. Type indicates the main data sources used by each representation: \textit{L} = location, \textit{P} = POI, \textit{R} = remote sensing, \textit{S} = street-view imagery, and \textit{O} = other urban/map data. \textit{Mean City Rank} is computed from unrounded city-level rankings across all tasks and cities (lower is better). Best, second-best, and third-best results in the Avg. columns are in \textbf{bold}, \underline{underlined}, and \uwave{wavy-underlined}, respectively. *For AGE, lower KL divergence indicates better performance, and results are reported only for the four cities with the most reliable age--sex source coverage.
}
\label{tab:main_results_split20}
\vspace{0.4em}
\scriptsize
\setlength{\tabcolsep}{1.8pt}
\resizebox{\textwidth}{!}{
\begin{tabular}{ll*{17}{c}}
\toprule
& & \multicolumn{4}{c}{$\spadesuit$~Morphology}
& \multicolumn{4}{c}{$\heartsuit$~Demographics}
& \multicolumn{4}{c}{$\diamondsuit$~Economy}
& \multicolumn{4}{c}{$\clubsuit$~Environment}
& \multicolumn{1}{c}{Overall} \\
\cmidrule(lr){3-6} \cmidrule(lr){7-10} \cmidrule(lr){11-14} \cmidrule(lr){15-18} \cmidrule(lr){19-19}
Model & Type
& \multicolumn{2}{c}{LUC}
& \multicolumn{2}{c}{RDE}
& \multicolumn{2}{c}{POP}
& \multicolumn{2}{c}{AGE$^{*}$}
& \multicolumn{2}{c}{GDP}
& \multicolumn{2}{c}{NTL}
& \multicolumn{2}{c}{PM$_{2.5}$}
& \multicolumn{2}{c}{LST}
& Rank \\
\cmidrule(lr){3-4} \cmidrule(lr){5-6}
\cmidrule(lr){7-8} \cmidrule(lr){9-10}
\cmidrule(lr){11-12} \cmidrule(lr){13-14}
\cmidrule(lr){15-16} \cmidrule(lr){17-18}
\cmidrule(lr){19-19}
&
& \shortstack{Avg.\\(F1 $\uparrow$)} & \shortstack{\textit{C}\\\textit{Std.}}
& \shortstack{Avg.\\($R^2 \uparrow$)} & \shortstack{\textit{C}\\\textit{Std.}}
& \shortstack{Avg.\\($R^2 \uparrow$)} & \shortstack{\textit{C}\\\textit{Std.}}
& \shortstack{Avg.\\(KL $\downarrow$)} & \shortstack{\textit{C}\\\textit{Std.}}
& \shortstack{Avg.\\($R^2 \uparrow$)} & \shortstack{\textit{C}\\\textit{Std.}}
& \shortstack{Avg.\\($R^2 \uparrow$)} & \shortstack{\textit{C}\\\textit{Std.}}
& \shortstack{Avg.\\($R^2 \uparrow$)} & \shortstack{\textit{C}\\\textit{Std.}}
& \shortstack{Avg.\\($R^2 \uparrow$)} & \shortstack{\textit{C}\\\textit{Std.}}
& \shortstack{Mean\\City Rank} \\
\midrule
PE & \textit{L} & 0.213 & \textit{0.070} & 0.191 & \textit{0.084} & 0.257 & \textit{0.091} & 0.036 & \textit{0.020} & 0.248 & \textit{0.154} & 0.430 & \textit{0.084} & \underline{0.617} & \textit{0.252} & \textbf{0.720} & \textit{0.098} & 6.297 \\

Place2Vec & \textit{P} & 0.200 & \textit{0.044} & 0.236 & \textit{0.098} & 0.336 & \textit{0.088} & 0.040 & \textit{0.020} & 0.273 & \textit{0.158} & 0.394 & \textit{0.105} & 0.143 & \textit{0.058} & 0.273 & \textit{0.105} & 8.359 \\

Space2Vec & \textit{LP} & 0.220 & \textit{0.054} & 0.171 & \textit{0.056} & 0.306 & \textit{0.093} & 0.037 & \textit{0.021} & 0.309 & \textit{0.174} & 0.500 & \textit{0.110} & 0.391 & \textit{0.207} & 0.565 & \textit{0.169} & 6.984 \\

CaLLiPer & \textit{LP} & 0.249 & \textit{0.075} & 0.209 & \textit{0.054} & 0.390 & \textit{0.102} & \textbf{0.033} & \textit{0.021} & 0.379 & \textit{0.188} & \uwave{0.589} & \textit{0.096} & 0.523 & \textit{0.210} & \uwave{0.661} & \textit{0.145} & 4.641 \\

CityFM & \textit{PO} & 0.230 & \textit{0.068} & 0.230 & \textit{0.080} & 0.355 & \textit{0.095} & \uwave{0.035} & \textit{0.021} & 0.354 & \textit{0.142} & 0.560 & \textit{0.129} & 0.526 & \textit{0.259} & \underline{0.693} & \textit{0.132} & 5.156 \\

Urban2Vec & \textit{PRS} & 0.196 & \textit{0.052} & 0.251 & \textit{0.096} & 0.321 & \textit{0.090} & 0.040 & \textit{0.021} & 0.371 & \textit{0.147} & 0.463 & \textit{0.078} & 0.169 & \textit{0.117} & 0.360 & \textit{0.161} & 7.719 \\

MuseCL & \textit{PRS} & 0.191 & \textit{0.038} & 0.262 & \textit{0.096} & 0.343 & \textit{0.092} & 0.044 & \textit{0.021} & 0.386 & \textit{0.159} & 0.441 & \textit{0.088} & 0.134 & \textit{0.081} & 0.315 & \textit{0.187} & 8.031 \\

SatCLIP & \textit{LR} & 0.122 & \textit{0.044} & 0.129 & \textit{0.079} & 0.147 & \textit{0.094} & 0.040 & \textit{0.020} & 0.016 & \textit{0.090} & 0.205 & \textit{0.125} & 0.205 & \textit{0.228} & 0.399 & \textit{0.106} & 10.031 \\

TESSERA & \textit{R} & \uwave{0.368} & \textit{0.075} & \textbf{0.640} & \textit{0.088} & \uwave{0.685} & \textit{0.105} & 0.037 & \textit{0.018} & \underline{0.664} & \textit{0.145} & 0.572 & \textit{0.136} & 0.529 & \textit{0.148} & 0.600 & \textit{0.132} & \uwave{3.422} \\

AlphaEarth & \textit{R} & \textbf{0.387} & \textit{0.068} & \underline{0.617} & \textit{0.085} & \underline{0.708} & \textit{0.112} & 0.036 & \textit{0.017} & \uwave{0.645} & \textit{0.132} & \textbf{0.706} & \textit{0.074} & \uwave{0.533} & \textit{0.137} & 0.612 & \textit{0.142} & \underline{2.859} \\

AETHER & \textit{PR} & \underline{0.375} & \textit{0.080} & \uwave{0.585} & \textit{0.091} & \textbf{0.713} & \textit{0.095} & \underline{0.034} & \textit{0.018} & \textbf{0.772} & \textit{0.136} & \underline{0.610} & \textit{0.084} & \textbf{0.667} & \textit{0.148} & 0.605 & \textit{0.167} & \textbf{2.500} \\

\midrule
Mean over models & -- & 0.250 & \textit{0.061} & 0.320 & \textit{0.082} & 0.415 & \textit{0.096} & 0.037 & \textit{0.020} & 0.402 & \textit{0.148} & 0.497 & \textit{0.101} & 0.403 & \textit{0.168} & 0.527 & \textit{0.140} & -- \\

\bottomrule
\end{tabular}
}
\end{table*}
}

\subsection{Exploratory Extension to 26 Cities}
\label{app:extended-26-cities}

\begin{figure}[!ht]
    \centering
    \includegraphics[width=0.95\linewidth]{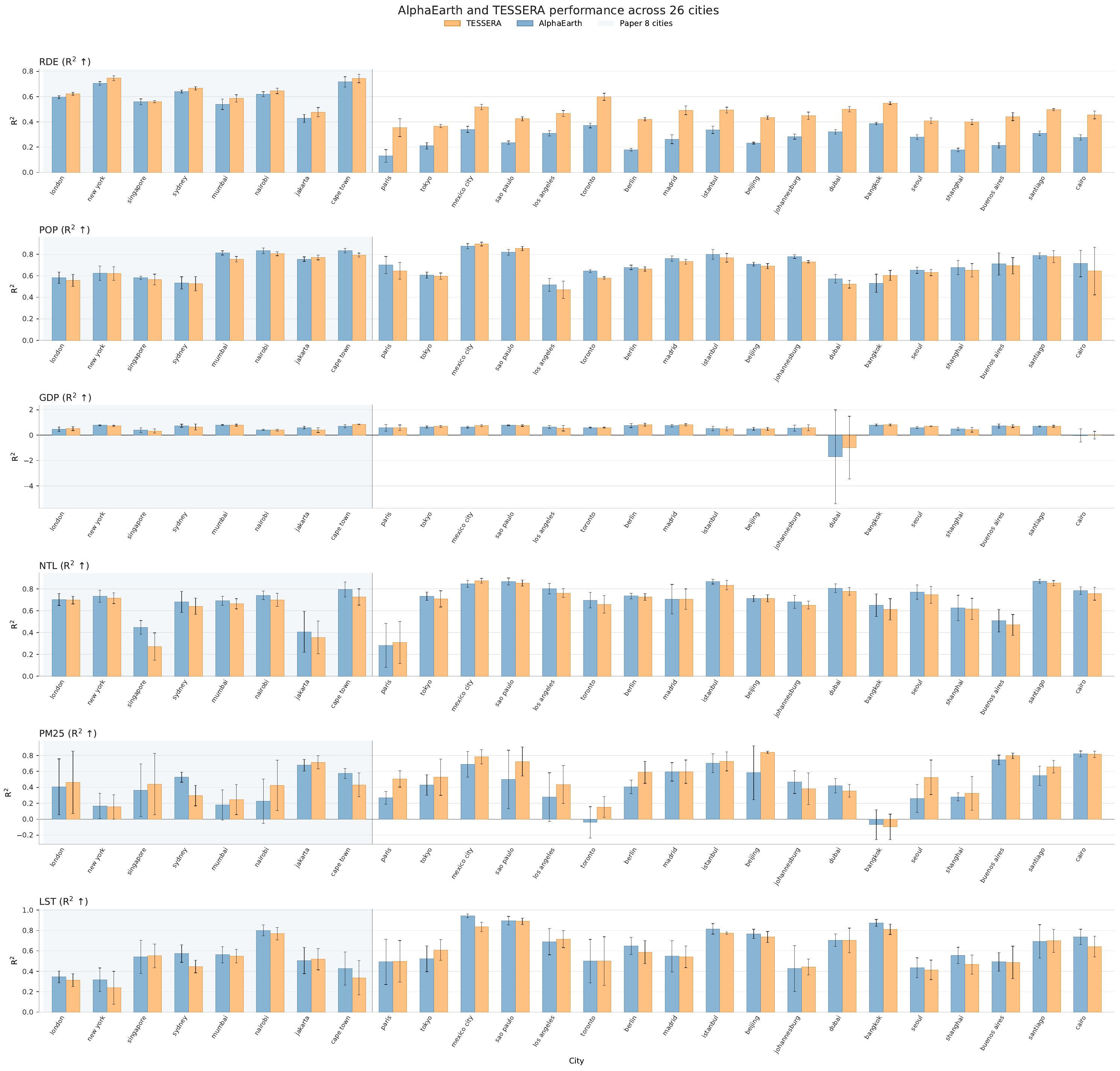}
    \caption{
    Exploratory 26-city extension using AlphaEarth and TESSERA. Bars report the mean primary metric across five spatial-split seeds for each city-task pair, and error bars report the standard deviation across seeds. The shaded region marks the eight cities used in the main benchmark setting. The extension uses the six globally available raster-based tasks: RDE, POP, GDP, NTL, PM$_{2.5}$, and LST. Higher is better for all reported metrics.
    }
    \label{fig:remote_sensing_26city_extension}
\end{figure}

To examine whether CityRep can scale beyond the eight benchmark cities, we construct an exploratory 26-city extension using the same task registry, alignment pipeline, spatial split protocol, and evaluation code. The extension includes the eight main benchmark cities and 18 additional cities for which the global raster-based task layers can be prepared. In this exploratory setting, we evaluate two raster representation models, AlphaEarth and TESSERA, on the six tasks available for all 26 cities: road density, population, GDP, NTL, PM$_{2.5}$, and land surface temperature.

Figure~\ref{fig:remote_sensing_26city_extension} compares AlphaEarth and TESSERA across all available city–task pairs. Overall, the two models exhibit broadly consistent cross-city trends: cities that are difficult for one model often also tend to be difficult for the other, suggesting that part of the performance variation reflects city- and task-specific data characteristics rather than purely model-specific behavior. At the same time, the two models show different strengths across individual tasks. TESSERA performs strongly on several morphology-related and environmental tasks, whereas AlphaEarth is competitive or stronger on some demographic and thermal tasks. This highlights the necessity of multi-task evaluation: results from any single task are insufficient for comprehensively understanding model behavior.

Although the multi-city evaluation reveals overall patterns of consistency, observations from an individual city can sometimes contradict the aggregate trend. For example, on the population task, TESSERA performs better in Mexico City and São Paulo, even though AlphaEarth outperforms TESSERA in most cities overall. We therefore treat this extension as evidence for the value of systematic multi-city evaluation. The 26-city extension is not used as the main leaderboard because not all baselines currently provide complete embeddings or reproducible checkpoints for the expanded city set, and tasks such as land use and age distribution require additional city-specific auditing. Nevertheless, the experiment demonstrates that CityRep’s registry-based design can support broader evaluation once new city–task files and model embeddings are registered.

\section{Analysis of Performance Heterogeneity Across Tasks and Cities} 
\label{app:city-difficulty-discussion}

\begin{figure}[t]
    \centering
    \includegraphics[width=\linewidth]{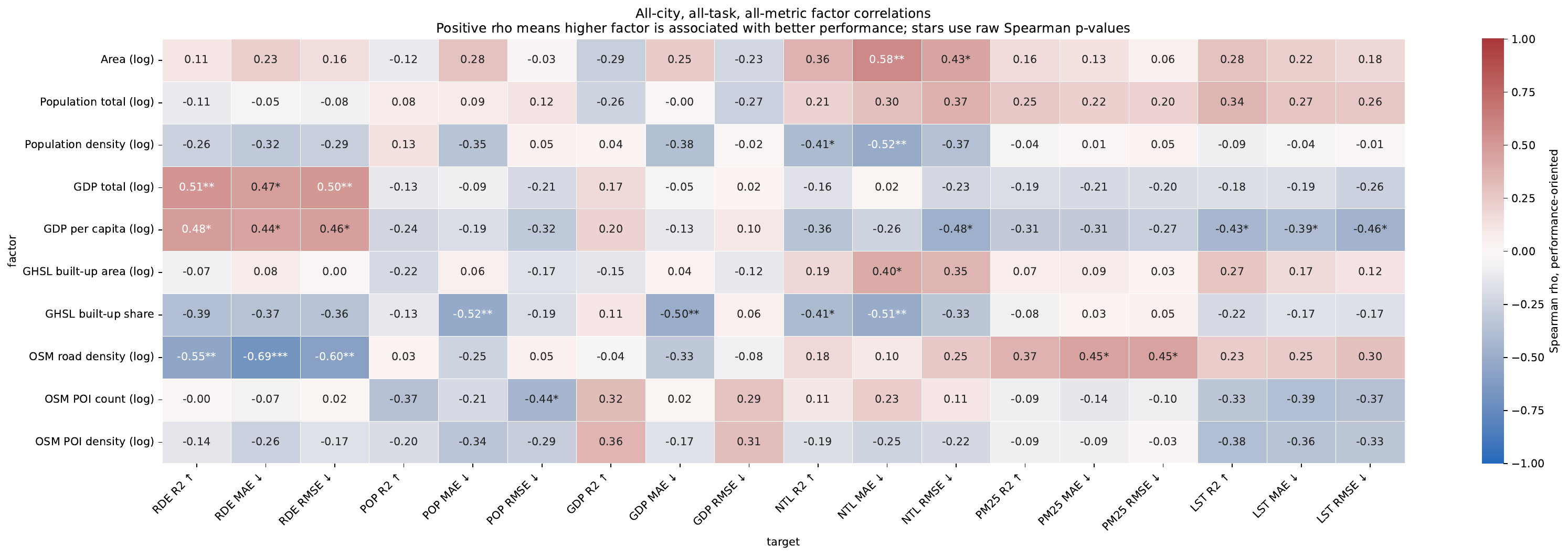}
    \caption{
    Exploratory correlation analysis between city-level factors and downstream performance across the 26-city remote-sensing extension. Each cell reports the Spearman correlation between a city factor and the two-model average performance of AlphaEarth and TESSERA for a task-metric pair. Metrics are oriented so that higher values always indicate better performance; KL, MAE, and RMSE are sign-flipped before averaging. Positive correlations therefore mean that larger factor values are associated with better performance. Stars denote raw, unadjusted Spearman p-values. The analysis is descriptive and intended to identify sources of performance heterogeneity rather than establish causal effects.
    }
    \label{fig:city_factor_correlation_heatmap}
\end{figure}

To better characterize this heterogeneity, we conduct an exploratory city-factor analysis in the 26-city remote-sensing extension. For each city, we construct a common set of interpretable descriptors: total area, population, population density, total GDP, GDP per capita, GHSL built-up area, GHSL built-up share, OSM road density, OSM POI count, and OSM POI density. We then combine AlphaEarth and TESSERA results to obtain a two-model average performance signal for each city-task-metric pair. Metrics are first converted to a common performance orientation, so that larger values always indicate better prediction: $R^2$ and F1 are kept unchanged, while KL divergence, MAE, and RMSE are sign-flipped. Because both models are evaluated on the same task and metric, we average the raw performance-oriented values directly rather than normalizing across cities. Finally, we compute Spearman correlations between each city factor and each task-metric performance signal. 

Figure~\ref{fig:city_factor_correlation_heatmap} reports these correlations. The heatmap shows that city difficulty across all tasks is not explained by a single factor such as city size, population, or economic development. Some associations are task-specific and occasionally counterintuitive. GDP-related factors are positively associated with road-density prediction performance, but they do not uniformly improve performance on socioeconomic or environmental targets. Population density and built-up share are often negatively associated with performance on several metrics, suggesting that compact and highly built-up environments can be more difficult for current representations. OSM road density is strongly negatively associated with road-density prediction performance, indicating that dense street systems remain challenging even when the target itself is derived from road structure. POI-based factors show weaker and less consistent correlations, which may reflect differences in OSM completeness across cities as well as the fact that POI distributions only partially capture the spatial signals needed by the downstream tasks.

A useful interpretation is that development-related factors involve two competing effects. On the one hand, more developed cities often have more complex urban structure, stronger functional specialization, denser infrastructure, and more heterogeneous economic activity. These properties can make tasks related to urban function and socioeconomic characteristics harder, because the mapping from observable geospatial patterns to targets such as GDP, nighttime lights, or other functional indicators may be less direct. On the other hand, more developed cities also tend to have richer and higher-quality supporting data, including better coverage of roads, POIs, and other spatial layers, which can improve both model fitting and evaluation reliability.

The opposite tension may hold for cities in less developed contexts. Their urban form may in some cases be simpler or more spatially regular, which could make certain prediction problems easier in principle. However, this potential advantage can be offset by weaker data coverage, lower source-data quality, and greater measurement uncertainty. We therefore do not interpret the correlations as showing that either developed or developing regions are uniformly easier. Instead, the observed performance reflects a trade-off between urban complexity and data availability, together with task-specific differences in what each target requires from the representation.

This analysis supports the design choice of CityRep. A single aggregate leaderboard score can hide large differences across cities, tasks, and evaluation settings. Robust evaluation of urban foundation models should therefore report not only average performance, but also spatial-split results, random-split results, city-level variation, and coarse geographic stratifications.


\end{document}